\documentclass{article}

\usepackage[final]{neurips_2021}

\usepackage[utf8]{inputenc} 
\usepackage[T1]{fontenc}    
\usepackage{hyperref}       
\usepackage{url}            
\usepackage{booktabs}       
\usepackage{amsfonts}       
\usepackage{nicefrac}       
\usepackage{microtype}      
\usepackage{xcolor}         

\usepackage{bbm}
\usepackage{amssymb}
\usepackage{amsmath}
\usepackage{graphics}
\usepackage{mathtools}
\usepackage{cleveref}
\usepackage{subcaption}
\usepackage{adjustbox}
\usepackage{listings}
\usepackage{multirow}

\newcommand\blfootnote[1]{%
  \begingroup
  \renewcommand\thefootnote{}\footnote{#1}%
  \addtocounter{footnote}{-1}%
  \endgroup
}

\DeclareMathOperator{\supp}{supp}
\DeclareMathOperator{\SB}{SB}
\DeclareMathOperator{\logit}{logit}
\DeclareMathOperator*{\argmin}{arg\,min}

\usepackage{stackengine}

\makeatletter
\newcommand{\distas}[1]{\mathbin{\overset{#1}{\kern\z@\sim}}}%

\DeclareMathOperator*{\Bernoulli}{Bernoulli}

\newcommand{\Cat}{\text{Cat}}

\newcommand{\wrt}{w.r.t.~}

\newcommand{\beqs}{\vspace{0mm}\begin{eqnarray}}
\newcommand{\eeqs}{\vspace{0mm}\end{eqnarray}}
\newcommand{\barr}{\begin{array}}
\newcommand{\earr}{\end{array}}

\newcommand{\E}{\mathbb{E}}
\newcommand{\1}{\mathbbm{1}}

\newcommand{\swap}[2]{#1 \leftrightharpoons #2}

\newcommand{\vpidist}{\pi \sim \prod_k \text{Dirichlet}(1_C)}

\newcommand{\vscj}{s^{\swap{c}{j}}}

\newcommand{\vsmj}{s^{\swap{m}{j}}}

\newcommand{\zs}[2]{z^{\swap{#1}{#2}}}

\newcommand{\gARS}{g_{\text{ARS}}}
\newcommand{\gARSM}{g_{\text{ARSM}}}

\title{Coupled Gradient Estimators for Discrete Latent Variables}
\author{
    Zhe Dong \\
    Google Research, Brain Team \\
    \texttt{zhedong@google.com}
    \And
    Andriy Mnih \\
    DeepMind \\
    \texttt{andriy@deepmind.com} \\
    \And
    George Tucker \\
    Google Research, Brain Team \\
    \texttt{gjt@google.com} \\ 
}

\begin{document}

\maketitle

\begin{abstract}
Training models with discrete latent variables is challenging due to the high variance of unbiased gradient estimators. While low-variance reparameterization gradients of a continuous relaxation can provide an effective solution, a continuous relaxation is not always available or tractable. \citet{dong2020disarm} and \citet{yin2020probabilistic} introduced a performant estimator that does not rely on continuous relaxations; however, it is limited to binary random variables. We introduce a novel derivation of their estimator based on importance sampling and statistical couplings, which we extend to the categorical setting. Motivated by the construction of a stick-breaking coupling, we introduce gradient estimators based on reparameterizing categorical variables as sequences of binary variables and Rao-Blackwellization. In systematic experiments, we show that our proposed categorical gradient estimators provide state-of-the-art performance, whereas even with additional Rao-Blackwellization, previous estimators~\citep{yin2019arsm} underperform a simpler REINFORCE with a leave-one-out-baseline estimator~\citep{kool2019buy}. 
\blfootnote{Code and additional information: \url{https://sites.google.com/view/disarm-estimator}.}
\end{abstract}

\section{Introduction}

Optimizing an expectation of a cost function of discrete variables with respect to the parameters of their distribution is a frequently encountered problem in machine learning. This problem is challenging because the gradient of the objective, like the objective itself, is an expectation over an exponentially large space of joint configurations of the variables. As the number of variables increases, these expectations quickly become intractable and thus are typically approximated using Monte Carlo sampling, trading a reduction in computation time for variance in the estimates.
When such stochastic gradient estimates are used for learning, their variance determines the largest learning rate that can be used without making training unstable. Thus finding estimators with lower variance leads directly to faster training by allowing higher learning rates. For example, the use of the reparameterization trick to yield low-variance gradient estimates has been essential to the success of variational autoencoders \citep{kingma2014auto,rezende2014stochastic}. However, this estimator, also known as the pathwise derivative estimator \citep{glasserman2013monte}, can only be applied to continuous random variables.

For discrete random variables, there are two common strategies for stochastic gradient estimation. The first one involves replacing discrete variables with continuous ones that approximate them as closely as possible \citep{maddison2017concrete,jang2017categorical} and training the resulting relaxed system with the reparameterization trick. However, as after training, the system is evaluated with discrete variables, this approach is not guaranteed to perform well and requires a careful choice of the continuous relaxation. Moreover, evaluating the cost function at the relaxed values instead of the discrete ones is not always desirable or even possible. 
The second strategy, involves using the REINFORCE estimator \citep{williams1992simple}, also known as the score-function \citep{rubinstein1990optimization} or likelihood-ratio  \citep{glynn1990likelihood} estimator, which, having fewer requirements than the reparameterization trick, also works with discrete random variables. As the simplest versions of this estimator tend to exhibit high variance, they are typically combined with variance reduction techniques. Some of the most effective such estimators \citep{tucker2017rebar,grathwohl2018backpropagation}, incorporate the gradient information provided by the continuous relaxation, while keeping the estimator unbiased \wrt the original discrete system.

The recently introduced Augment-REINFORCE-Merge (ARM) \citep{yin2018arm} estimator for binary variables and Augment-REINFORCE-Swap (ARS) and Augment-REINFORCE-Swap-Merge (ARSM) estimators \citep{yin2019arsm} for categorical variables provide a promising alternative to relaxation-based estimators.
However, when compared to a simpler baseline approach (REINFORCE with a leave-one-out-baseline~\citep[RLOO; ][]{kool2019buy}), ARM underperforms in the binary setting~\citep{dong2020disarm}, and we similarly demonstrate that ARS and ARSM underperform in the categorical setting. \citet{dong2020disarm} and \citet{yin2020probabilistic} independently developed an estimator that uses Rao-Blackwellization to improve ARM and outperforms RLOO, providing state-of-the-art performance in the binary setting.

In this paper, we explore how to devise a performant estimator in the categorical setting. A natural first approach is to apply the ideas from \citep{dong2020disarm,yin2020probabilistic} to ARS and ARSM. However, empirically, we find that this is insufficient to close the gap between ARS/ARSM and RLOO. Due to space constraints, we defer the details to Appendix~\ref{sec:arsm}. Instead, we develop a novel derivation of DisARM/U2G~\citep{dong2020disarm,yin2020probabilistic} using importance sampling which is simpler and more direct while providing a natural extension to the categorical case. This estimator requires constructing a \emph{coupling} on categorical variables, which we do using a stick-breaking process and antithetic Bernoulli variables. Motivated by this construction, we also consider estimators based on reparameterizing the problem with a sequence of binary variables. We systematically evaluate these estimators and their underlying design choices and find that they outperform RLOO across a range of problems without requiring more computation.

\section{Background}
We consider the problem of optimizing
\begin{equation}
\E_{q_\theta(z)} \left[ f_\theta(z) \right], 
\label{eq:obj}
\end{equation}
with respect to the parameters $\theta$ of a factorial categorical distribution $q_\theta(z) = \prod_k \Cat(z_k; \alpha_{\theta, k})$ where $k$ indexes dimension and $\alpha_{\theta, k}$ is the vector of logits of the categorical distribution with $C$ choices.\footnote{To simplify notation, we omit the subscripted $\theta$ on $\alpha$ when it is clear from context.} This situation covers many problems with discrete latent variables, for example, in variational inference $f_\theta(z)$ could be the instantaneous ELBO~\citep{jordan1999introduction} and $q_\theta(z)$ the variational posterior.

The gradient with respect to $\theta$ is
\begin{equation}
\nabla_\theta \E_{q_\theta(z)} \left[ f_\theta(z) \right] = \E_{q_\theta(z)} \left[ f_\theta(z) \nabla_\theta \log q_\theta(z)  + \nabla_\theta f_\theta(z) \right].
\label{eq:reinforce}
\end{equation}
It typically suffices to estimate the second term with a single Monte Carlo sample, so for notational clarity, we omit the dependence of $f$ on $\theta$ in the following sections. Monte Carlo estimates of the first term can have large variance. Low-variance, unbiased estimators of the first term will be our focus.

\subsection{DisARM/U2G}
\citet{dong2020disarm} and \citet{yin2020probabilistic} derived a Rao-Blackwellized estimator for \emph{binary} variables based on the coupled estimator of~\citet{yin2018arm}. In particular, if $q_\theta(b) \sim \prod_k \Bernoulli(\alpha_{\theta, k})$ with $\theta$ parameterizing the logits $\alpha_{\theta_k}$ of the Bernoulli distribution, and $b$ and $\tilde{b}$ are antithetic samples\footnote{Antithetic Bernoulli samples can be defined by the following process: sample $u \sim \text{Uniform}(0, 1)$, then set $b = u < p$ and $\tilde{b} = (1 - u) < p$, where $p$ is the probability parameter of the Bernoulli variable.} from $q_\theta(b)$ (independent across dimension $k$), then the estimator
\begin{equation*}
\frac{1}{2} \left(f(b) - f(\tilde{b})\right)\left(
(-1)^{\tilde{b}_{k}}  \1_{b_{k} \ne \tilde{b}_{k}}  \sigma(|(\alpha_{\theta,k})|)
\right) 
\end{equation*}            
is an unbiased estimator of the gradient $\nabla_{\alpha_{\theta, k}} \E_{q_\theta(b)}\left[f(b)\right]$. This estimator has been shown to outperform RLOO, but is limited to the binary variable case.

\section{Methods}
\label{sec:methods}
First, we provide a novel derivation of DisARM~\citep{dong2020disarm} / U2G~\citep{yin2020probabilistic} from an importance-sampling perspective, which naturally extends to the categorical case. Starting with the (2-sample) REINFORCE LOO estimator~\citep{kool2019buy} and applying importance sampling, we have
\begin{align*} 
\nabla_{\theta_k} & \E_{q_\theta(z)} \left[ f(z) \right] = \frac{1}{2}\E_{q_\theta(z)q_\theta(\tilde{z})}\left[ \overbrace{(f(z) - f(\tilde{z}))\left(\nabla_{\theta_k} \log q_\theta(z_k) - \nabla_{\theta_k} \log q_\theta(\tilde{z}_k) \right)}^{\text{RLOO}(z, \tilde{z})} \right] \\
&= \frac{1}{2}\E_{p_\theta(z, \tilde{z})}\left[ \frac{q_\theta(z)q_\theta(\tilde{z})}{p_\theta(z, \tilde{z})}(f(z) - f(\tilde{z}))\left(\nabla_{\theta_k} \log q_\theta(z_k) - \nabla_{\theta_k} \log q_\theta(\tilde{z}_k) \right) \right].
\end{align*}
To reduce the variance of the estimator, we can use the joint distribution $p_\theta(z, \tilde{z})$ to emphasize terms that have high magnitude. However, controlling the weights can be challenging for high dimensional $z$. We can sidestep this issue by taking advantage of the structure of the integrand and requiring that $p_\theta$ be a properly supported\footnote{As is standard for importance sampling, to ensure unbiasedness, we require that either $q(z, \tilde{z}) = 0$ or the integrand is $0$ whenever $p(z, \tilde{z}) = 0$.} \emph{coupling} that is independent across dimensions (i.e., a joint distribution $p(z, \tilde{z}) = \prod_k p(z_k, \tilde{z}_k)$ such that the marginals are maintained $p(z_k) = p(\tilde{z_k}) = q(z_k)$). Then with $(z, \tilde{z}) \sim p$, the following will be an unbiased estimator of the gradient (see Appendix~\ref{app:iw-derivation} for the derivation)
\begin{equation}
{g_\text{DisARM-IW}}_k = \frac{1}{2}\frac{q_\theta(z_k)q_\theta(\tilde{z_k})}{p_\theta(z_k, \tilde{z_k})}(f(z) - f(\tilde{z}))\left(\nabla_{\theta_k} \log q_\theta(z_k) - \nabla_{\theta_k} \log q_\theta(\tilde{z}_k) \right).
\label{eq:iw_estimator}
\end{equation}
Critically, because the coupling maintains the marginal distributions, we only need to importance weight a single dimension at a time, which ensures the weights are reasonable. The estimator is unbiased, so the coupling can be designed to reduce variance. In the binary case, with an antithetic Bernoulli coupling, it is straightforward to see that this precisely recovers DisARM/U2G. We note that~\citet{dimitriev2021arms} independently and concurrently discovered a similar result. Conveniently, this estimator is also valid in the categorical case, however, the choice of coupling affects the performance of the estimator.

\subsection*{Stick-breaking Coupling}
The ideal variance-reducing coupling would take into account the magnitude of $(f(z) - f(\tilde{z}))\left(\nabla_{\theta_k} \log q_\theta(z_k) - \nabla_{\theta_k} \log q_\theta(\tilde{z}_k) \right)$ which we do not know a priori. However, we do know that when $z_k = \tilde{z}_k$, the last multiplicative term of Eq.~\ref{eq:iw_estimator} vanishes, so moving mass away from this configuration will reduce variance. Furthermore, for the estimator to be valid, the coupling must put non-zero mass on all $z_k \neq \tilde{z_k}$ configurations.\footnote{Technically, the coupling can put zero mass on configurations as long as the expectation is zero across those configurations, however, this is hard to ensure for a general $f$.}

In the binary case, there is a natural construction of an antithetic coupling which minimizes the probability of $z_k = \tilde{z}_k$. We can extend this construction using formulations of categorical variables as a function of a sequence of binary decisions. The stick-breaking construction for categorical variables~\citep{khan2012stick} provides such an approach. Suppose we have a categorical distribution $q(z_k)$ and we construct a sequence of independent binary variables $b_{k,1}, \ldots, b_{k, C}$ with $b_{k, i} \sim \Bernoulli(q(z_k=i)/\sum_{j=i}^C q(z_k=j))$ and define $z_k \coloneqq \SB(b_{k,1}, \ldots, b_{k,C}) \coloneqq \min i$ s.t. $b_{k,i} = 1$; then $z_k \sim q(z_k)$. Given each of $b_{k,1}, \ldots, b_{k, C}$, we can independently sample antithetic Bernoulli variables $\tilde{b}_{k, 1}, \ldots, \tilde{b}_{k, C}$ and let $\tilde{z} \coloneqq \SB(\tilde{b}_{k, 1}, \ldots, \tilde{b}_{k, C})$. This process defines a joint distribution on $(z_k, \tilde{z}_k)$, which we call the \emph{stick-breaking coupling}. By construction, the joint distribution preserves the marginal distributions. Note that the stick-breaking coupling depends on the an arbitrary ordering of the categories.

Without reordering the categories, the coupling arising from the default ordering may not put non-zero mass on all $z_k \neq \tilde{z}_k$ configurations. When $p(b_{k, i}) \geq 0.5$, then $p(b_{k, i} = 0, \tilde{b}_{k, i} = 0) = 0$, which implies that $p(z_k > i, \tilde{z}_k > i) = 0$ by definition. So if $p(b_{k, i}) > 0.5$ for $i < C$, this violates the condition that the coupling put mass on all $z_k \neq \tilde{z}_k$ configurations. We can avoid this situation by relabeling the categories in the ascending order of probability (as this guarantees that $q(z_k=i)/\sum_{j=i}^C q(z_k=j) \leq 0.5$). With this ordering, computing the required importance weights with the stick-breaking coupling is straightforward when $z_k \neq \tilde{z}_k$
\begin{equation*}
    \frac{q_\theta(z_k)q_\theta(\tilde{z}_k)}{p_\theta(z_k, \tilde{z}_k)} = \prod_{i=1}^{\min(z_k, \tilde{z_k})-1}\frac{\sigma(-\alpha_{k, i})^2}{1-2\sigma(\alpha_{k, i})} \sigma\left(-\alpha_{\min(z_k, \tilde{z_k})}\right),
\end{equation*} 
where $\alpha_{k, i} = \logit \frac{q(z_k=i)}{\sum_{j=i}^C q(z_k=j)}$ (see Appendix~\ref{app:stick-breaking} for details). Note that due to the ascending ordering, the $\sigma(\alpha_{k, i})$ term in the product is always $\leq \frac{1}{3}$, ensuring that the weights are bounded.

Finally, we note that a mixture of couplings is also a coupling, so we can take any coupling and mix it with the independent coupling to ensure the support condition is met. Because the estimator is unbiased for any choice of nonzero mixing coefficient, the mixing coefficients can be optimized following the control variate tuning process described in~\citep{ruiz2016overdispersed,tucker2017rebar,grathwohl2018backpropagation}. This also ensures that the performance of the estimator will be at least as strong as RLOO. Exploring this idea could be an interesting future direction.

\subsection{Binary Reparameterization}
Motivated by the stick-breaking construction, we can eschew importance sampling and directly reparameterize the problem in terms of the binary variables and apply DisARM/U2G. In particular, if we have binary variables $b_{k, 1} \sim \Bernoulli(\sigma(\alpha_{k, 1})), \ldots, b_{k, C}\sim \Bernoulli(\sigma(\alpha_{k, C}))$ such that $z_k \coloneqq h(b_{k, 1}, \ldots, b_{k, C}) \sim q_\theta(z_k)$ where $h$ is a deterministic function that does not depend on $\theta$, then we can rewrite the gradient as 
\begin{align*} 
\nabla_{\alpha_{k, c}} \E_{q_\theta(z)}[f(h(b))] &= \nabla_{\alpha_{k, c}} \E_{b}[f(h(b))] =  \E_{b}[f(h(b)) \nabla_{\alpha_{k, c}} \log q_\theta(b)] \\
&= \E_{b}[f(h(b)) \nabla_{\alpha_{k, c}} \log q_\theta(b_{k, c})],
\end{align*}
which is an expression that DisARM/U2G can be applied to. 

We can additionally leverage the structure of the stick-breaking construction to further improve the estimator. Intuitively, by the definition of $z_k$ in terms of $\SB(\cdots)$, if we know that $b_{k, c} = 1$, then the values of $b_{k, c'}$ for $c' > c$ do not change the value of $z_k$, which implies the gradient terms vanishing. In particular, given antithetically coupled Bernoulli samples, the following is an unbiased estimator (see Appendix~\ref{app:stick-breaking} for details):
\begin{equation}
    {g_\text{DisARM-SB}}_{k, c} = \begin{cases}
         \frac{1}{2} \left(f(z) - f(\tilde{z})\right)\left(
                    (-1)^{\tilde{b}_{k, c}}  \1_{b_{k, c} \ne \tilde{b}_{k, c}}  \sigma(|(\alpha_{k, c}|)
                \right) & c \le \min(z_k, \tilde{z}_{k}) \\
    \frac{1}{2}\left(f(\tilde{z}) - f(z)\right)\nabla_{\alpha_{k, c}} \log q_\theta(\tilde{b}_{k, c}) & z_k < c \leq \tilde{z}_k \\
    \frac{1}{2}\left(f(z) - f(\tilde{z})\right)\nabla_{\alpha_{k, c}} \log q_\theta(b_{k, c}) & \tilde{z}_k < c \leq z_k \\
    0 & c > \max(z_k, \tilde{z}_k)
    \end{cases}.
    \label{eq:disarm_cat_sb}
\end{equation}
In contrast to the previous section, \emph{any} choice of ordering for the stick-breaking construction results in an unbiased estimator, and we experimentally evaluated several choices.

Alternatively, we can use a sequence of independent Bernoulli variables arranged as a binary tree to construct categorical variables, with each leaf node representing a category~\citep{zoubin2010treestructure}.  For simplicity, we only consider balanced binary trees (i.e., $C$ a power of $2$).  The binary variables encode the internal routing decisions through the tree, defining a deterministic mapping from the binary variables to the categorical variable $z \coloneqq T(b_1, \ldots, b_{C-1})$. It is straightforward to derive  the Bernoulli probabilities from $q_\theta(z)$ so that $z \sim q_\theta(z)$, so we defer the details to Appendix~\ref{app:tree}.

As above, we can additionally leverage the structure of the tree construction to further improve the estimator. Given binary variables $b_1, \ldots, b_{C-1}$, let $I(b_1, \ldots, b_{C-1})$ be the set of variables used in routing decisions ($|I(b_1, \ldots, b_{C-1})| = \log_2 C$). By the definition of $z_k$ in terms of $T(\ldots)$, we know that $z_k$ is unaffected by binary decisions that occur outside $I(b_{\cdot})$, which implies the gradient terms vanishing.  With a pair of antithetically sampled binary sequences, the following is an unbiased estimator (see Appendix~\ref{app:tree} for details):
\begin{equation}
    {g_\text{DisARM-Tree}}_{k,c} = \begin{cases}
         \frac{1}{2} \left(f(z) - f(\tilde{z})\right)\left(
                    (-1)^{\tilde{b}_{k,c}}  \1_{b_{k,c} \ne \tilde{b}_{k,c}}  \sigma(|(\alpha_{k,c}|)
                \right) & c \in I(b_{k, \cdot}) \cap I(\tilde{b}_{k, \cdot}) \\
    \frac{1}{2}\left(f(\tilde{z}) - f(z)\right)\nabla_{\alpha_{k,c}} \log q_\theta(\tilde{b}_{k,c}) & c \in I(\tilde{b}_{k, \cdot}) - I(b_{k, \cdot}) \\
    \frac{1}{2}\left(f(z) - f(\tilde{z})\right)\nabla_{\alpha_{k,c}} \log q_\theta(b_{k,c}) & c \in I(b_{k, \cdot}) - I(\tilde{b}_{k, \cdot}) \\
    0 & c \notin I(b_{k, \cdot}) \cup I(\tilde{b}_{k, \cdot}) 
    \end{cases}.
    \label{eq:disarm_cat_tree}
\end{equation}
Again, the choice of ordering affects the coupling, however, we leave investigating the optimal choice to future work.

Notably both estimators include terms that are reminiscent of the LOO estimator, however, they use \emph{dependent} $z, \tilde{z}$ and are still unbiased due to careful construction.

\subsection{Multi-sample extension}
\label{sec:multisample}
We can extend the proposed estimators to the case when we have $n$-coupled pairs $(z^1, \tilde{z}^1), \ldots, (z^n, \tilde{z}^n)$. \citet{dimitriev2021arms} note that the $2n$-sample RLOO estimator can be written as an average over all $2$-sample RLOO estimators. In other words, given $2n$ independent samples $z^1, \ldots, z^{2n}$, 
\[ g_{\text{RLOO}}(z^1, \ldots, z^{2n}) = \frac{1}{2n(2n-1)}\sum_{i \neq j} g_{\text{RLOO}}(z^i, z^j). \]
If instead of $2n$ independent samples, we used $n$-coupled pairs in the estimator, most 2-sample RLOO estimators will be with independent samples, however, $n$ estimators will use coupled pairs which introduces a bias. We form the unbiased $n$-coupled pair estimator by adding a correction term that replaces those terms with a DisARM estimator
\[ g_{\text{RLOO}}(z^1, \tilde{z}^1, \ldots, z^n, \tilde{z}^n) + \frac{2}{2n(2n-1)}\sum_{i=1}^n (g_{\text{DisARM-*}}(z^i, \tilde{z}^i) - g_{\text{RLOO}}(z^i, \tilde{z}^i)), \]
where we can use any of the previous DisARM-* estimators.

\section{Related Work}
Almost all unbiased gradient estimators for discrete random variables (RVs), including our proposed estimators, belong to the REINFORCE / score function / likelihood-ratio estimator family \citep{williams1992simple,rubinstein1990optimization,glynn1990likelihood}. 
The primary difference between such estimators is the variance reduction techniques they incorporate, trading additional computation for lower variance of the estimate. Averaging over multiple independent samples is the simplest but relatively inefficient variance reduction technique, as the variance of the estimate is inversely proportional to the number of samples and thus decreases slowly. REINFORCE with the leave-out-out baseline \citep[RLOO;][]{kool2019buy} provides a much more effective way of utilizing multiple independent samples, by using the average over all the other samples as the baseline for each sample. \citet{kool2020estimating} extend this approach by sampling \emph{without} replacement, however, they find that in the high-dimensional spaces we consider, we almost never see such duplicates, resulting in little to no improvement in sampling without replacement compared to sampling with replacement (Figures 2(b) and 4(b) in~\citep{kool2020estimating}).

The estimators we develop use a technique called \emph{coupling} \citep{mcbook} to make better use of multiple samples by making them dependent. Recently, this kind of approach has been applied to binary RVs, resulting in the Augment-REINFORCE-Merge \citep[ARM;][]{yin2018arm} and DisARM \citep{dong2020disarm} / U2G \citep{yin2020probabilistic} estimators. ARM is constructed by reparameterizing Bernoulli RVs in terms of Logistic variables and applying antithetic sampling, which is the simplest kind of coupling, to the resulting REINFORCE estimator. DisARM / U2G is obtained by Rao-Blackwellizing the ARM estimator to eliminate its dependence on the values of the underlying Logistic variables, reducing its variance.
We show how to employ DisARM with categorical RVs by representing them as sequences of binary decisions and applying DisARM to the resulting system. By considering the sequences obtained through stick-breaking and tree-structured decisions, we obtain the DisARM-SB and DisARM-Tree estimators respectively.
We obtain DisARM-IW by generalizing the two-sample version of RLOO to allow the two samples to be coupled and using an importance-weighting correction to keep the estimator unbiased. DisARM-IW reduces to DisARM in the case of Bernoulli RVs and antithetic coupling, and thus can be seen as a generalization. \citet{dimitriev2021arms} has concurrently and independently developed an estimator identical to DisARM-IW, though they considered only the case of Bernoulli random variables.

Augment-REINFORCE-Swap estimator \citep[ARS; ][]{yin2019arsm} is a generalization of ARM to categorical RVs; it reparameterizes the categorical variables in terms of Dirichlet RVs and applies REINFORCE to them, using a coupling based on a swapping operation. Augment-REINFORCE-Swap-Merge \citep[ARSM; ][]{yin2019arsm} goes one step further by averaging over all possible choices of the pivot dimension that ARS chooses arbitrarily. 
Our Rao-Blackwellized ARS+ and ARSM+ estimators are obtained by analytically integrating out the randomness introduced by the augmenting Dirichlet dimensions from the ARS and ARSM estimators respectively, following the strategy that was used to derive DisARM from ARM (see Appendix~\ref{sec:arsm} for details). 

Rao-Blackwellization has been a popular variance reduction technique in the literature on gradient estimation with discrete random variables, having been used in BBVI \citep{ranganath2014black}, NVIL \citep{mnih2014neural}, LEG \citep{titsias2015local}, as well as more recently by \citet{liu2019rao} and \citet{paulus2020rao}.
Stick-breaking has previously been used in the context of VAEs to allow inferring the latent space dimensionality ~\citep{nalisnick2016stick}.

In this paper we use variational training of latent variable models as a benchmark for unbiased gradient estimators for general expectations rather than a goal in itself. As a result, we do not compare to specialized techniques for training models with discrete latent variables, such as Reweighted Wake Sleep \citep{bornschein2015reweighted} and Joint Stochastic Approximation \citep{jsa2020}, which do not provide such unbiased estimators and optimize different objectives.

\section{Experiments}
\label{sec:experiments}
\begin{table}[t]
\caption{Mean variational lower bounds and the standard error of the mean computed based on 5 runs of $5\times 10^5$-steps training from different random initializations. The best performing method (up to the standard error) for each task is in bold. Top: comparing the performance of the proposed estimators and the RLOO baseline on three datasets. Bottom: comparing the effect of arranging the categories in the ascending or descending order to the default ordering for DisARM-SB.} 
\label{tab:exp_comparison}
\begin{adjustbox}{center}
{\scriptsize
\begin{tabular}{l cccc | c}
\toprule
\multicolumn{6}{c}{Estimator Comparison with Train ELBO}\\
\midrule
\midrule
{} & DisARM-IW &  DisARM-Tree & DisARM-SB & RLOO & RELAX \\
\midrule
DynamicMNIST &  $-103.41\pm0.23$ & $\bf{-103.10\pm0.25}$ &   $-103.35\pm0.17$ &  $-104.03\pm0.23$ & $-102.22\pm0.34$ \\
FashionMNIST  &  $-240.08\pm0.17$ &  $\bf{-239.83\pm0.29}$ &   $\bf{-239.66\pm0.20}$ &  $-240.89\pm0.49$ & $-238.81\pm0.34$ \\
Omniglot &  $\bf{-122.25\pm0.35}$ &  $-123.08\pm0.43$ &  $\bf{-121.99\pm0.17}$ &  $\bf{-122.70\pm0.80}$ & $-120.40\pm0.16$
\\
\bottomrule
\end{tabular}
}
\end{adjustbox}
\begin{adjustbox}{center}
{\scriptsize
\begin{tabular}{l ccc}
\toprule
\multicolumn{4}{c}{Ordering Comparison with Train ELBO}\\
\midrule
\midrule
{} & Ascending & Descending & Default \\
\midrule
DynamicMNIST &   $\bf{-103.35\pm0.17}$ &    $\bf{-103.56\pm0.28}$ &  $\bf{-103.46\pm0.17}$ \\
FashionMNIST &   $\bf{-239.66\pm0.20}$ &    $-240.04\pm0.24$ &  $-239.99\pm0.20$ \\
Omniglot &   $\bf{-121.99\pm0.17}$ &    $-122.79\pm0.12$ &  $-122.75\pm0.15$ \\
\bottomrule
\end{tabular}
}
\end{adjustbox}
\end{table}
\begin{table}[t]
\caption{Evaluating the estimators and the effect of ordering on FashionMNIST with different model sizes. Mean variational lower bounds and the standard error of the mean computed based on 5 runs of $5\times 10^5$-steps training from different random initializations. We consider three  combinations of the number of categories (C) and the number of latent variables (V). The best performing method (up to the standard error) for each task is in bold. Top: evaluating the performance of the proposed estimators and the RLOO baseline. Bottom: comparing the effect of arranging the categories in the ascending or descending order to the default ordering for DisARM-SB.} 
\label{tab:exp_model_size}
\begin{adjustbox}{center}
{\footnotesize
\begin{tabular}{l cccc}
\toprule
\multicolumn{5}{c}{Estimator Comparison with Train ELBO}\\
\midrule
\midrule
{} & DisARM-IW & DisARM-Tree & DisARM-SB & RLOO \\
\midrule
C64~/~V5  &  $\bf{-244.75\pm0.53}$ &  $\bf{-244.29\pm0.30}$ &   $\bf{-244.81\pm0.64}$ &  $-245.27\pm0.30$ \\
C64~/~V32 &  $-240.08\pm0.17$ &  $\bf{-239.83\pm0.29}$ &   $\bf{-239.66\pm0.20}$ &  $-240.89\pm0.49$  \\
C16~/~V32 &  $\bf{-239.94\pm0.14}$ &  $-240.52\pm0.22$ &   $-240.43\pm0.13$ &  $-240.99\pm0.25$ \\
\bottomrule
\end{tabular}
}
\end{adjustbox}
\begin{adjustbox}{center}
{\footnotesize
\begin{tabular}{l ccc}
\toprule
\multicolumn{4}{c}{Ordering Comparison with Train ELBO}\\
\midrule
\midrule
{} & Ascending & Descending & Default \\
\midrule
C64~/~V5  &   $\bf{-244.81\pm0.64}$ &    $\bf{-244.95\pm0.36}$ &  $\bf{-244.61\pm0.38}$ \\
C64~/~V32 &   $\bf{-239.66\pm0.20}$ &    $-240.04\pm0.24$ &  $-239.99\pm0.20$ \\
C16~/~V32 &   $\bf{-240.43\pm0.13}$ &    $\bf{-240.39\pm0.21}$ &  $\bf{-240.50\pm0.13}$ \\
\bottomrule
\end{tabular}
}
\end{adjustbox}
\end{table}

Our goal was to derive unbiased, low-variance gradient estimators for the categorical setting, which do not rely on continuous relaxation. We primarily compared against the REINFORCE estimator with a leave-one-out baseline~\citep[RLOO,][]{kool2019buy}. In the high dimensional setting that we evaluated, RLOO has been shown to be a strong baseline algorithm, performing at least as well as more complex methods, such as RELAX, REBAR, STORB, and UnOrd, as shown in~\citep{richter2020vargrad,kool2020estimating}. As in prior work~\citep{yin2019arsm}, we benchmark the estimators by training variational auto-encoders~\citep{kingma2014auto,rezende2014stochastic} (VAE) with \emph{categorical} latent variables on three datasets: binarized MNIST\footnote{Creative Commons Attribution-Share Alike 3.0 license}~\citep{lecun2010mnist}, FashionMNIST\footnote{MIT license}~\citep{xiao2017/online}, and Omniglot\footnote{MIT license}~\citep{lake2015human}. As we sought to evaluate optimization performance, we use dynamic binarization to avoid overfitting and largely find that training performance mirrors test performance. We use the standard split into train, validation, and test sets.

To allow direct comparison, we use the same model structure as in~\citep{yin2019arsm}.
Briefly, the model has a single layer of categorical latent variables which are then represented as one-hot vectors and decoded to Bernoulli logits for the observation through either a linear transformation or an MLP with two hidden layers of $256$ and $512$ of LeakyReLU units~\citep{xu2015leakyrelu}. The encoder mirrors the structure, with two hidden layers of $512$ and $256$ LeakyReLU units and outputs producing the softmax logits. For most experiments, we used 32 latent variables with 64 categories unless specified otherwise. See Appendix~\ref{app:exp-details} for more details.

\subsection{Evaluating DisARM-IW, DisARM-SB \& DisARM-Tree}
First, we evaluated the importance-weighted estimator (DisARM-IW), and DisARM estimators with the stick breaking construction (DisARM-SB) and the tree structured construction (DisARM-Tree). These estimators require only $2$ expensive function evaluations regardless of the number of categories ($C$), unlike ARS/ARSM which require up to $O(C)/O(C^2)$ evaluations, so we use the 2-independent sample RLOO estimator as a baseline. 

\begin{figure}[h]
    \centering
    \rotatebox[origin=l]{90}{{\scriptsize {\quad \quad DynamicMNIST}}}
    \;
    \includegraphics[width=0.31\linewidth]{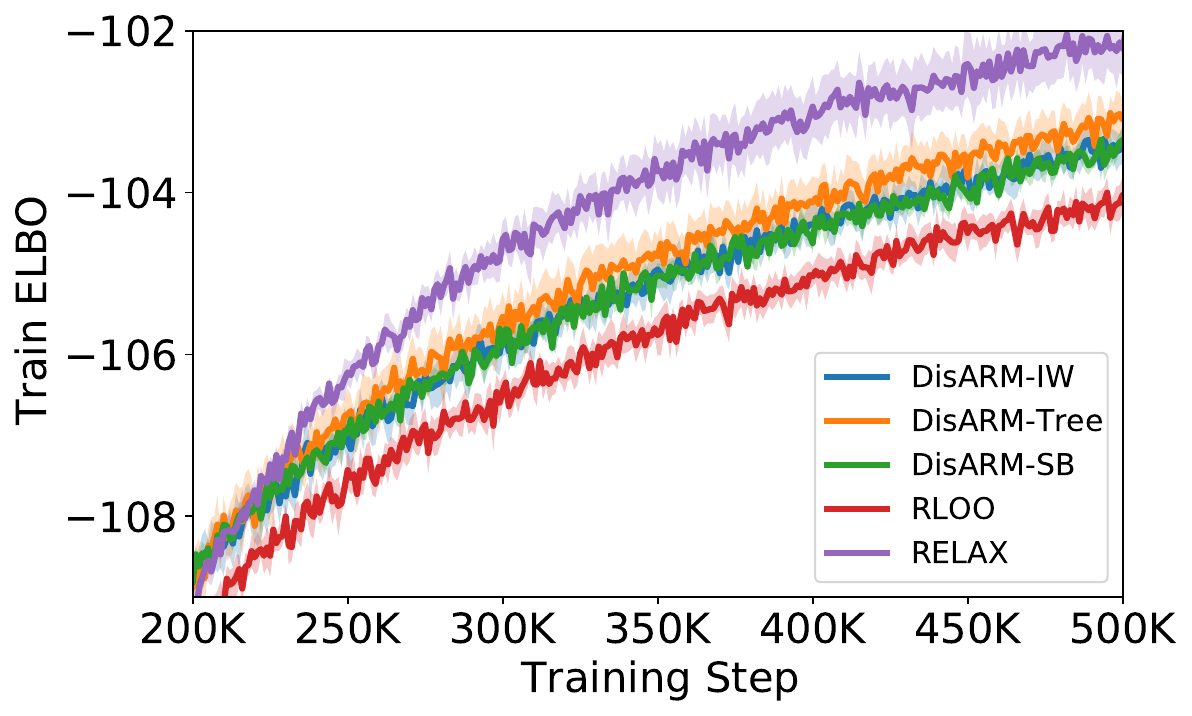}
    \;
    \includegraphics[width=0.31\linewidth]{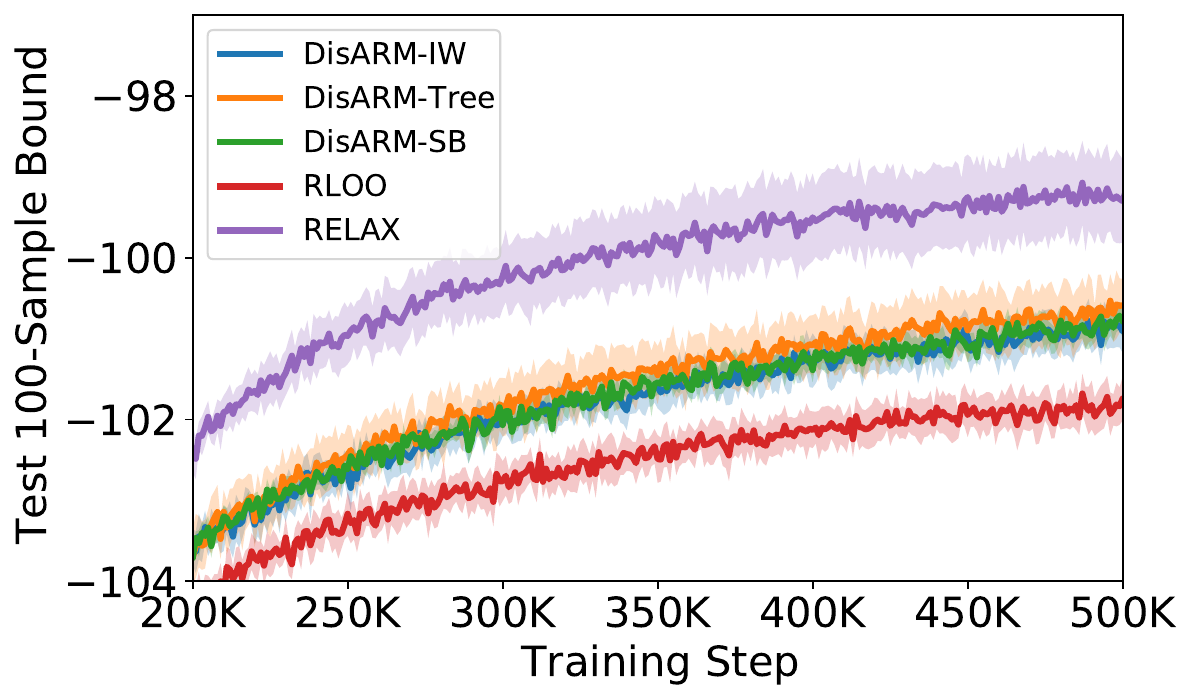}
    \;
    \includegraphics[width=0.29\linewidth]{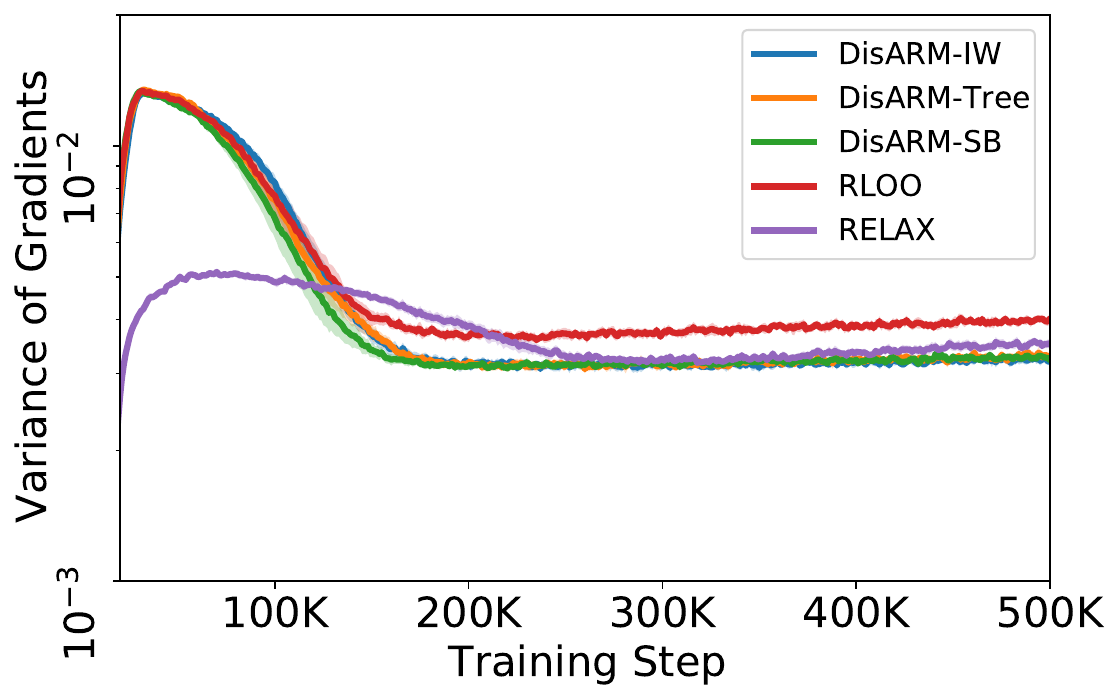}
    \\
    \rotatebox[origin=l]{90}{{\scriptsize {\quad \quad FashionMNIST}}}
    \;
    \includegraphics[width=0.31\linewidth]{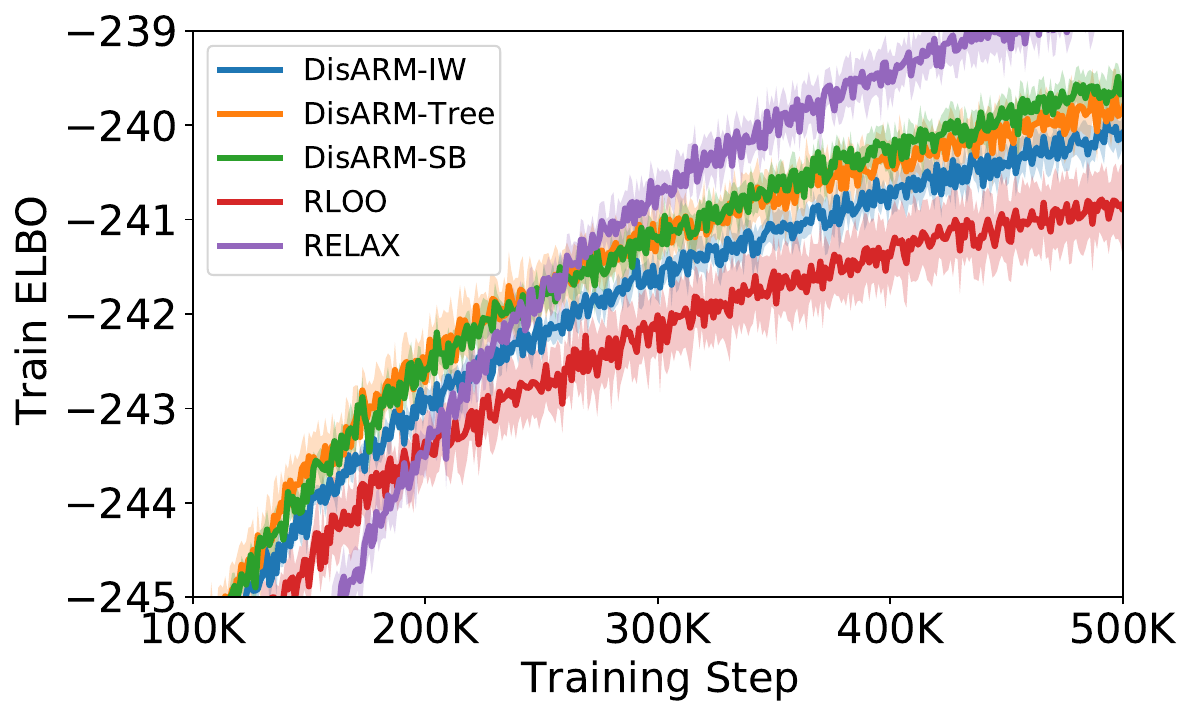}
    \;
    \includegraphics[width=0.31\linewidth]{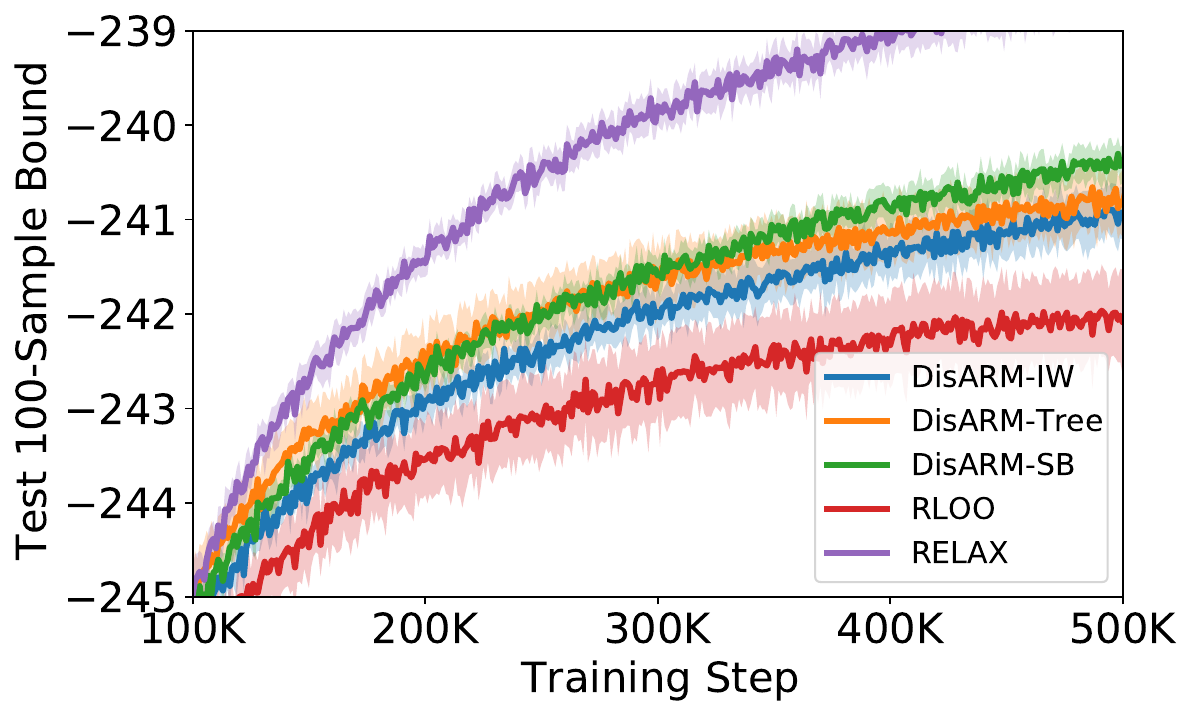}
    \;
    \includegraphics[width=0.29\linewidth]{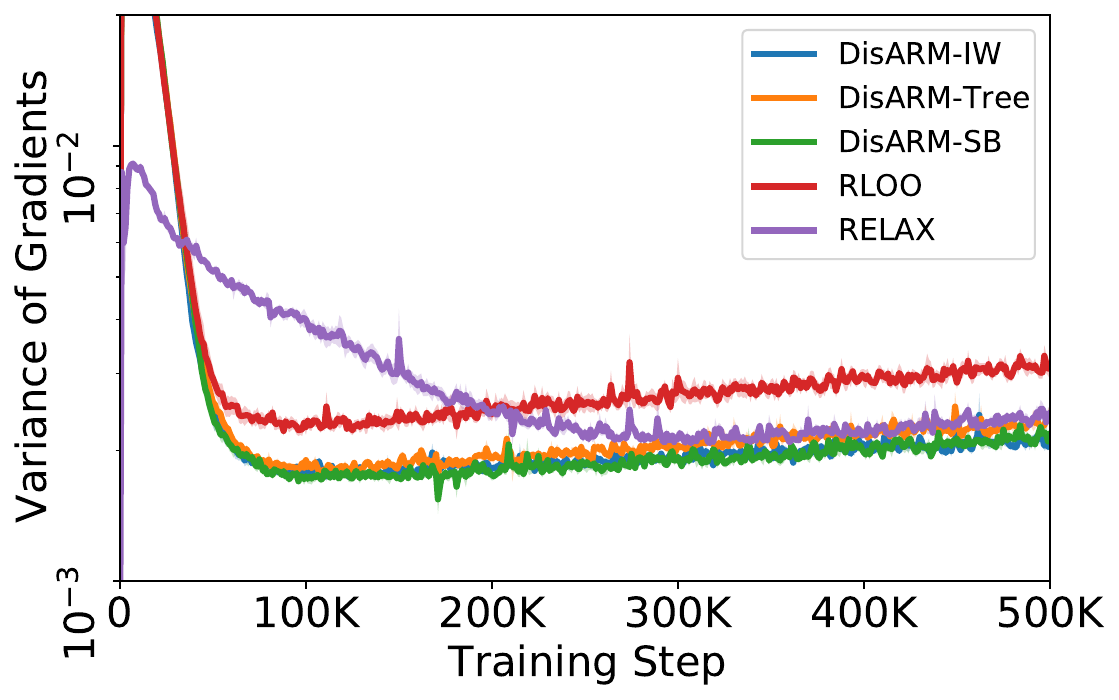}
    \\
    \rotatebox[origin=l]{90}{{\scriptsize {\quad \quad \quad Omniglot}}}
    \;
    \includegraphics[width=0.31\linewidth]{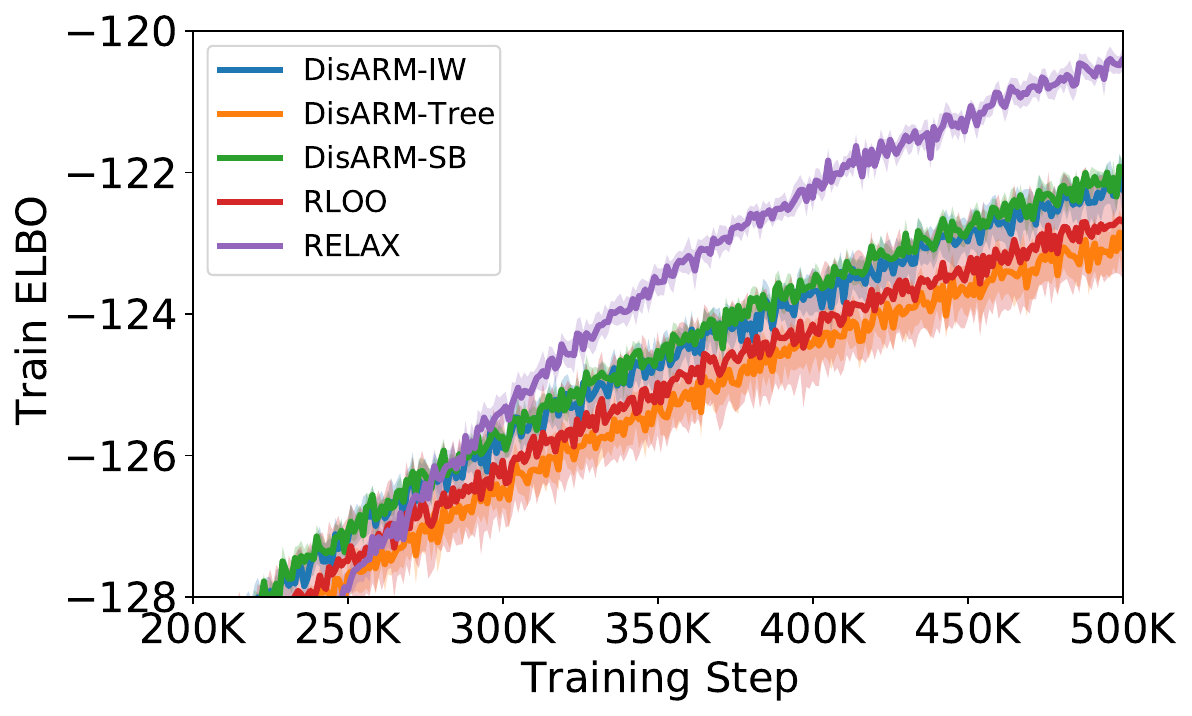}
    \;
    \includegraphics[width=0.31\linewidth]{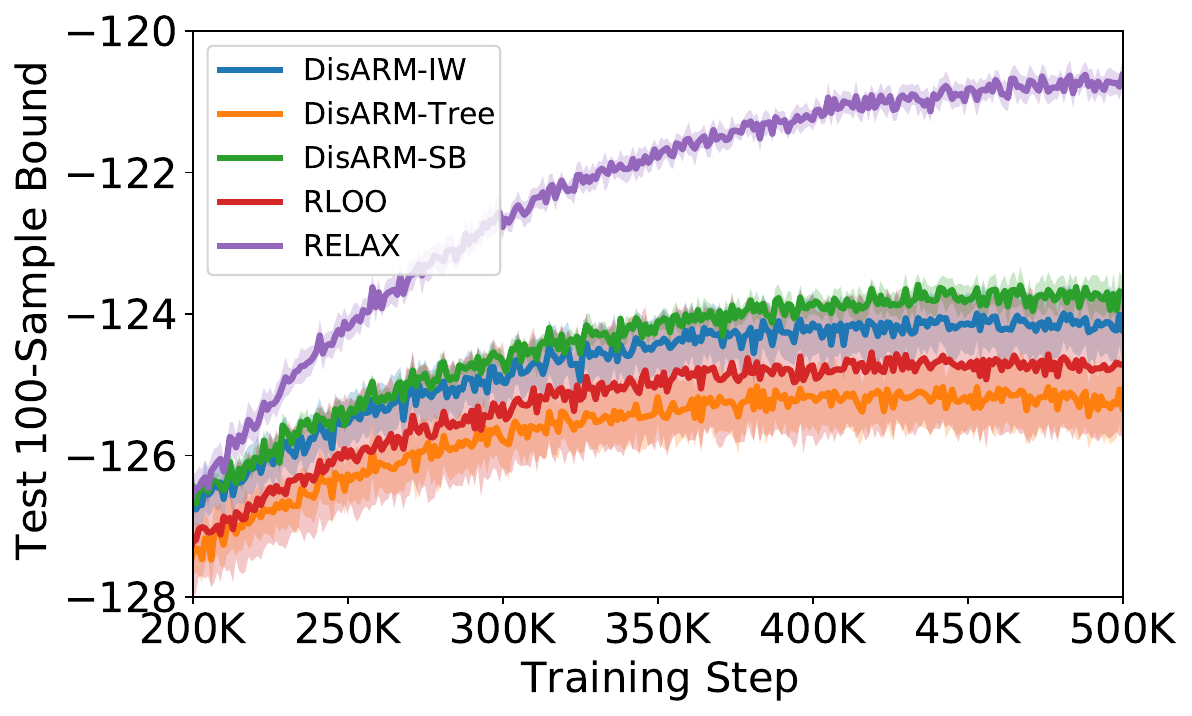}
    \;
    \includegraphics[width=0.29\linewidth]{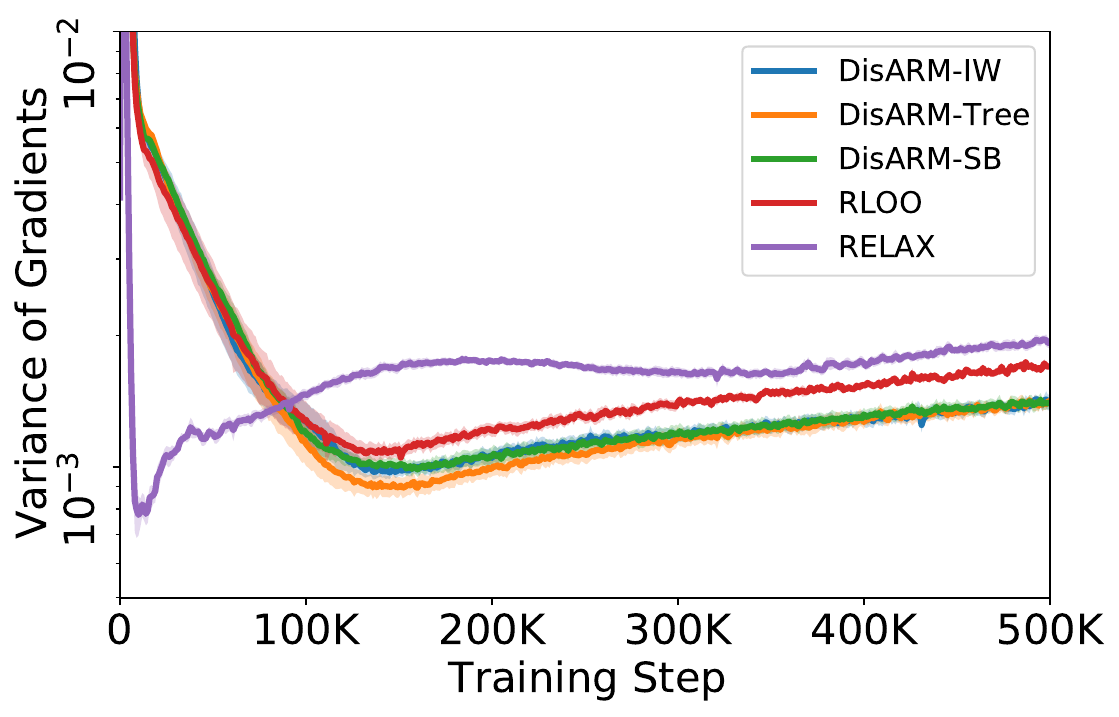}
    \caption{Training a non-linear categorical VAE with $32$ latent variables with $64$ categories on dynamically binarized MNIST, FashionMNIST, Omniglot datasets by maximizing the ELBO. We plot the train ELBO (left column), test 100-sample bound (middle column), and the variance of gradient estimator (right column). For a fair comparison, the variance of all the gradient estimators was computed along the training trajectory of the RLOO estimator. We plot the mean and one standard error based on $5$ runs from different random initializations.}
    \label{fig:experiment-benchmarking}
\end{figure}

\begin{figure}[h]
    \captionsetup[subfigure]{labelformat=empty}
    \centering
    \begin{subfigure}[b]{0.32\textwidth}
        \includegraphics[width=\textwidth]{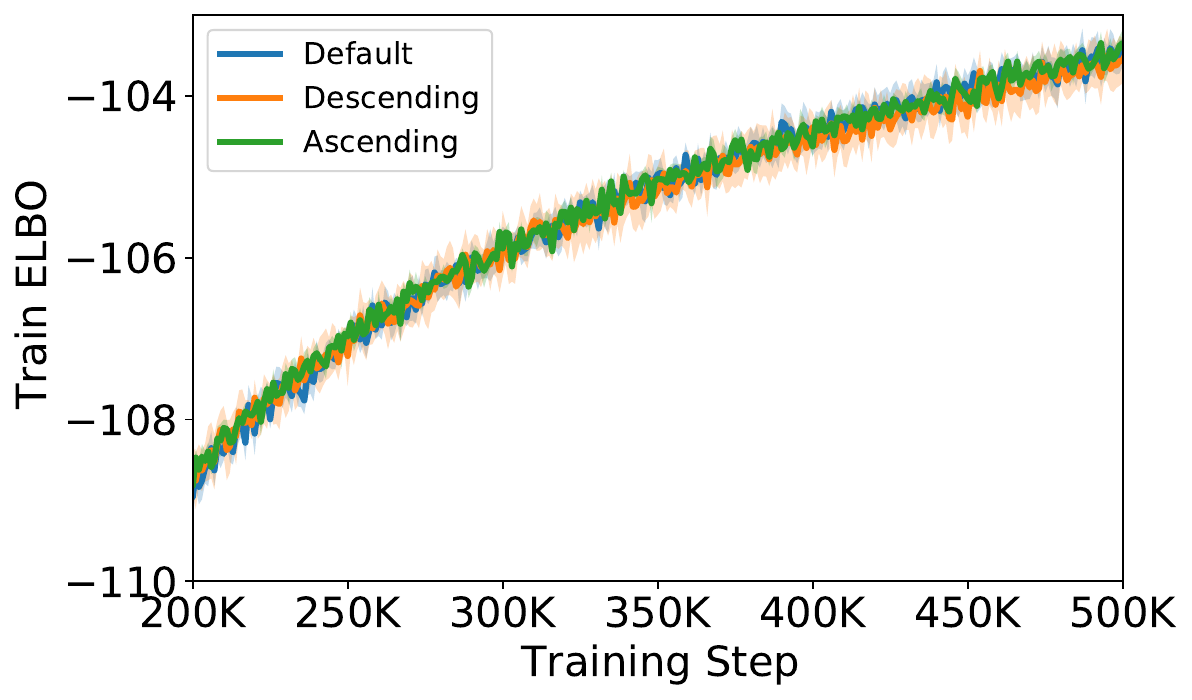}
        \caption{DynamicMNIST}
    \end{subfigure}
    \begin{subfigure}[b]{0.32\textwidth}
        \includegraphics[width=\textwidth]{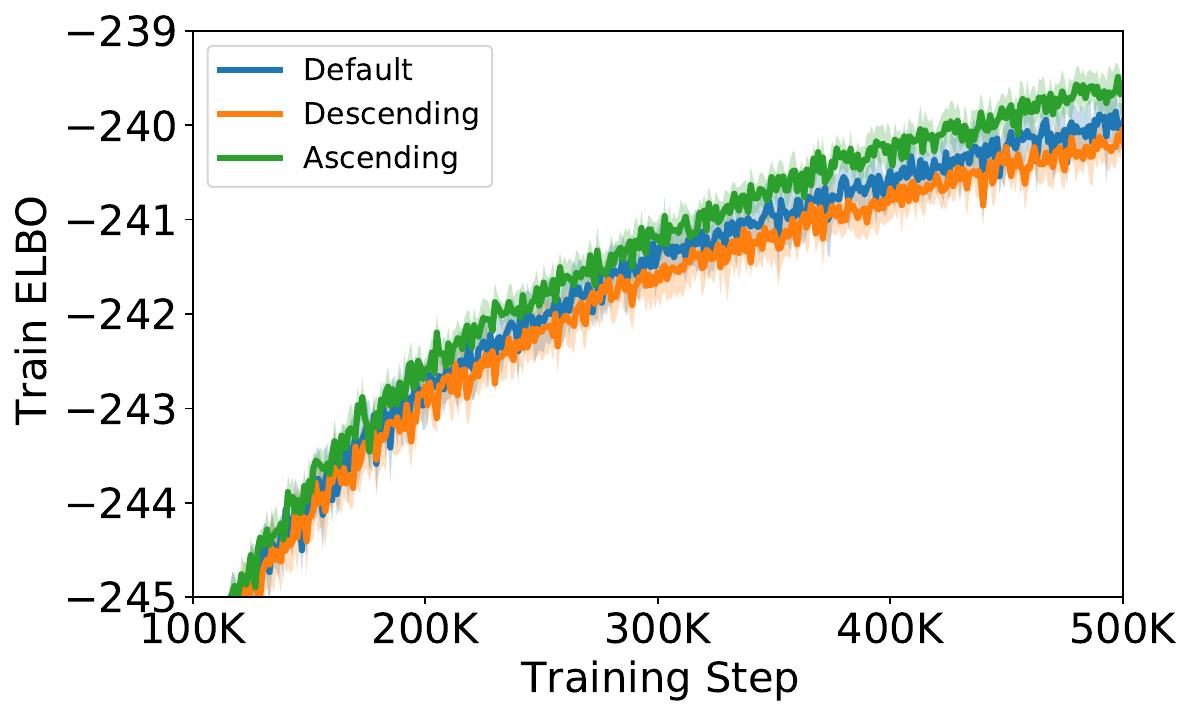}
        \caption{FashionMNIST}
    \end{subfigure}
    \begin{subfigure}[b]{0.32\textwidth}
        \includegraphics[width=\textwidth]{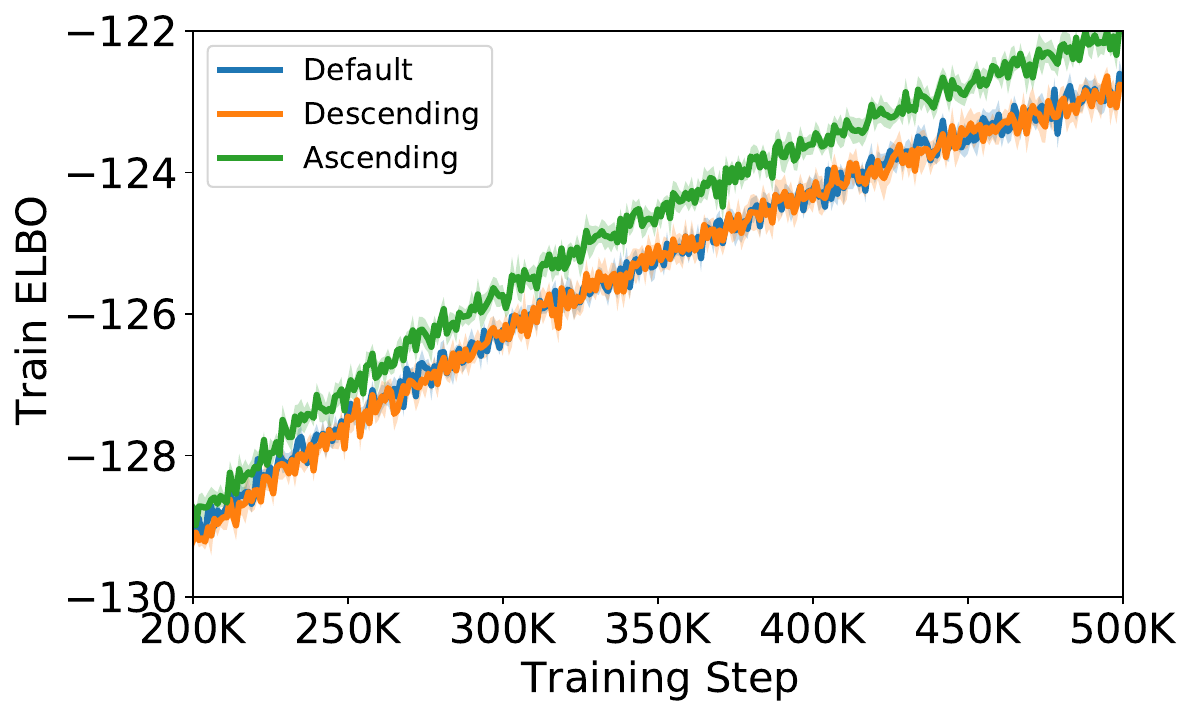}
        \caption{Omniglot}
    \end{subfigure}
    \caption{The effect of logit ordering on the performance of DisARM-SB. We sort the encoder logits in the ascending or descending order and compare against the default ordering.}
    \label{fig:experiment-ordering}
\end{figure}

In the case of nonlinear models (\Cref{fig:experiment-benchmarking}), DisARM-IW and DisARM-SB perform similarly, with lower gradient variance and better performance than the baseline estimator (RLOO) across all datasets. While for DisARM-IW, we have to use the ascending order for the stick-breaking construction to ensure unbiasedness, for DisARM-SB any ordering results in an unbiased estimator. We experimented with ordering the categories in the ascending and descending order by probability as well as using the default ordering (\Cref{fig:experiment-ordering}). We found that the estimator with the ascending order consistently outperforms the other ordering schemes, though the gain is dataset-dependent. However, when we varied the number of categories and variables (\Cref{fig:experiment-modelsize}), we found that no ordering dominates. Given that ordering introduces an additional complexity, we recommend no ordering in practice.

In the case of linear models, we found that all estimators performed similarly (Appendix~\ref{app:experiment}). \citet{dong2020disarm} found that for multi-layer linear models in the binary case, DisARM showed increasing improvement over RLOO for models with deeper hierarchies. As nonlinear models are more common and expressive, we leave exploring whether this trend holds in the categorical case to future work.

We further verified that the observed improvements of the proposed estimators \wrt the RLOO baseline are consistent for models with different sizes of the latent space. As shown in~Appendix \Cref{fig:experiment-modelsize}, we find that the best choice of the proposed estimators depends on the model size, however all three estimators outperform RLOO in all cases.

\subsection{Evaluating Multi-sample Estimators}

\begin{figure}[h]
    \captionsetup[subfigure]{labelformat=empty}
    \centering
    \rotatebox[origin=l]{90}{{\scriptsize {\quad \quad \quad 5 Pairs}}}
    \;
    \begin{subfigure}[b]{0.3\textwidth}
        \includegraphics[width=\textwidth]{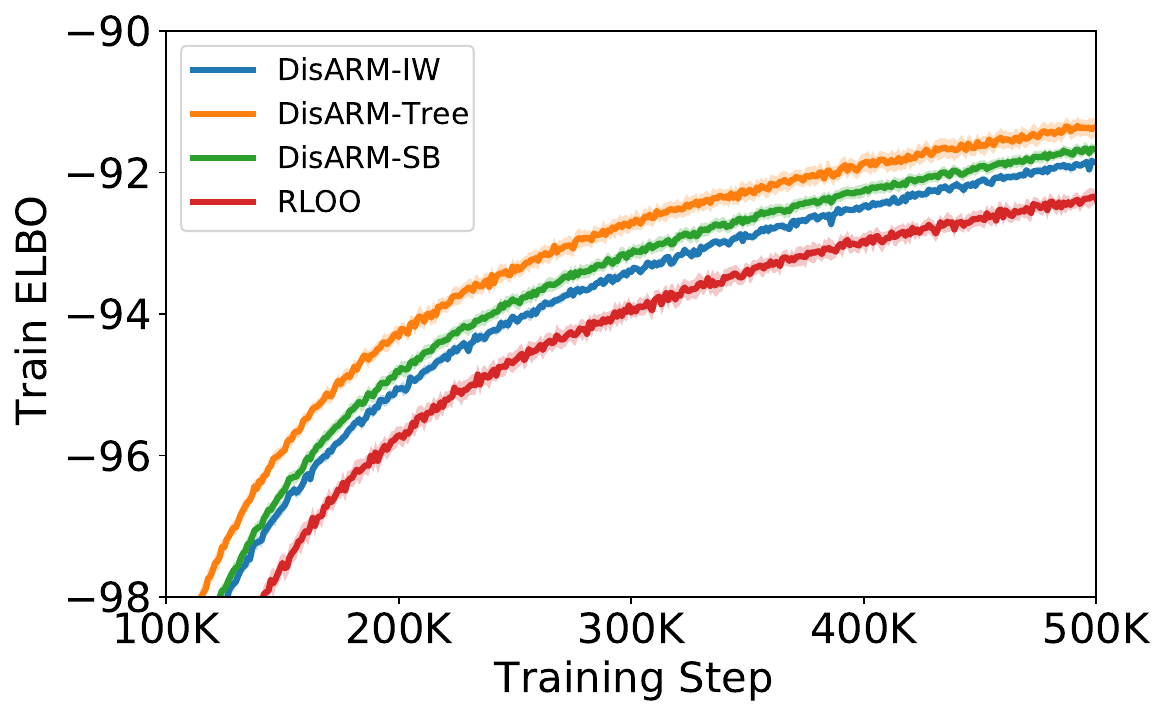}
    \end{subfigure}
    \begin{subfigure}[b]{0.3\textwidth}
        \includegraphics[width=\textwidth]{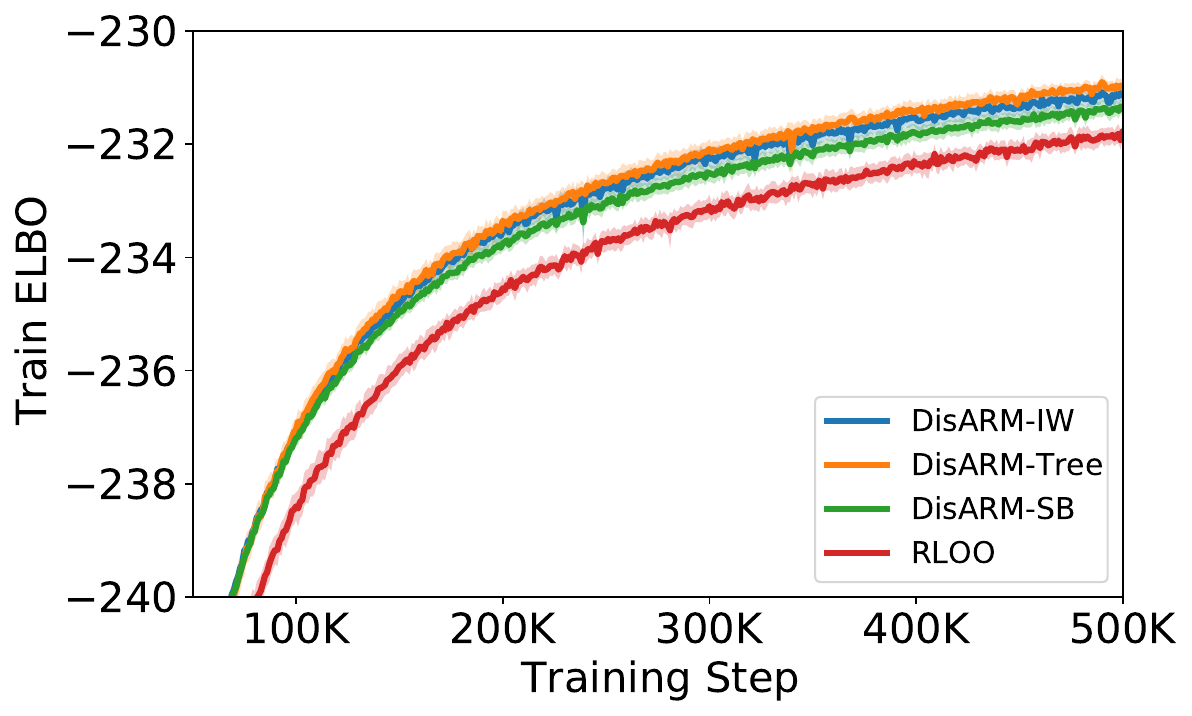}
    \end{subfigure}
    \begin{subfigure}[b]{0.3\textwidth}
        \includegraphics[width=\textwidth]{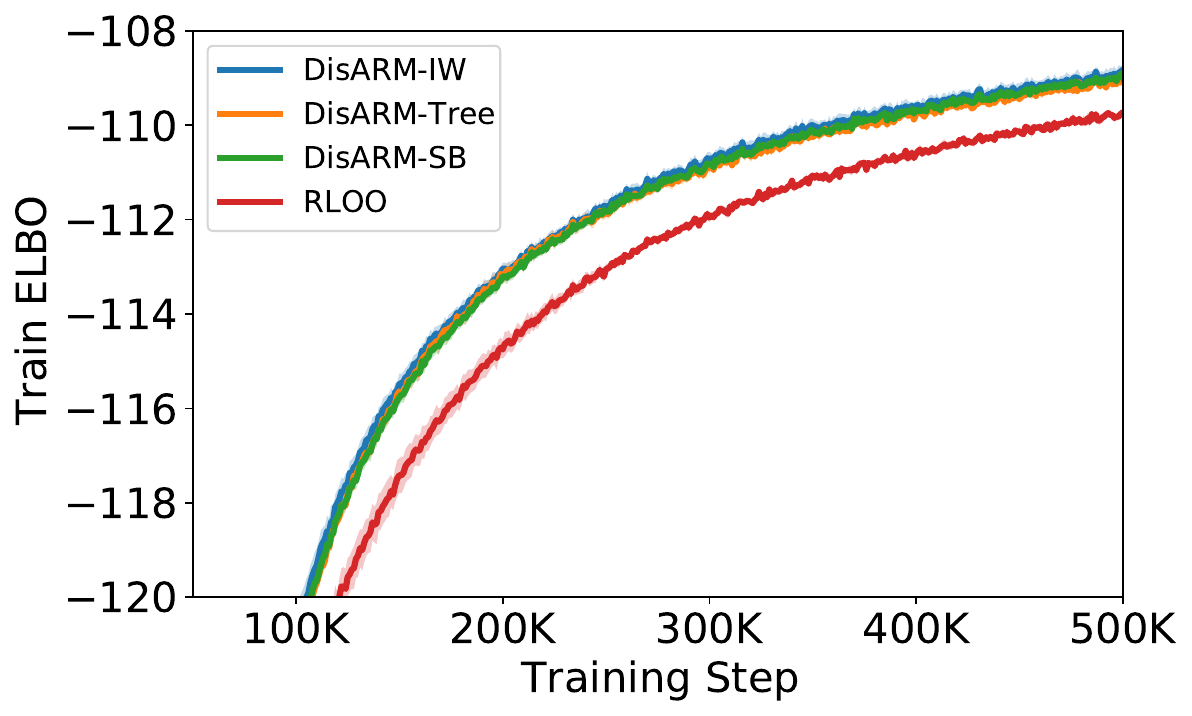}
    \end{subfigure}
    \\
    \rotatebox[origin=l]{90}{{\scriptsize {\qquad \qquad \qquad 10 Pairs}}}
    \;
    \begin{subfigure}[b]{0.3\textwidth}
        \includegraphics[width=\textwidth]{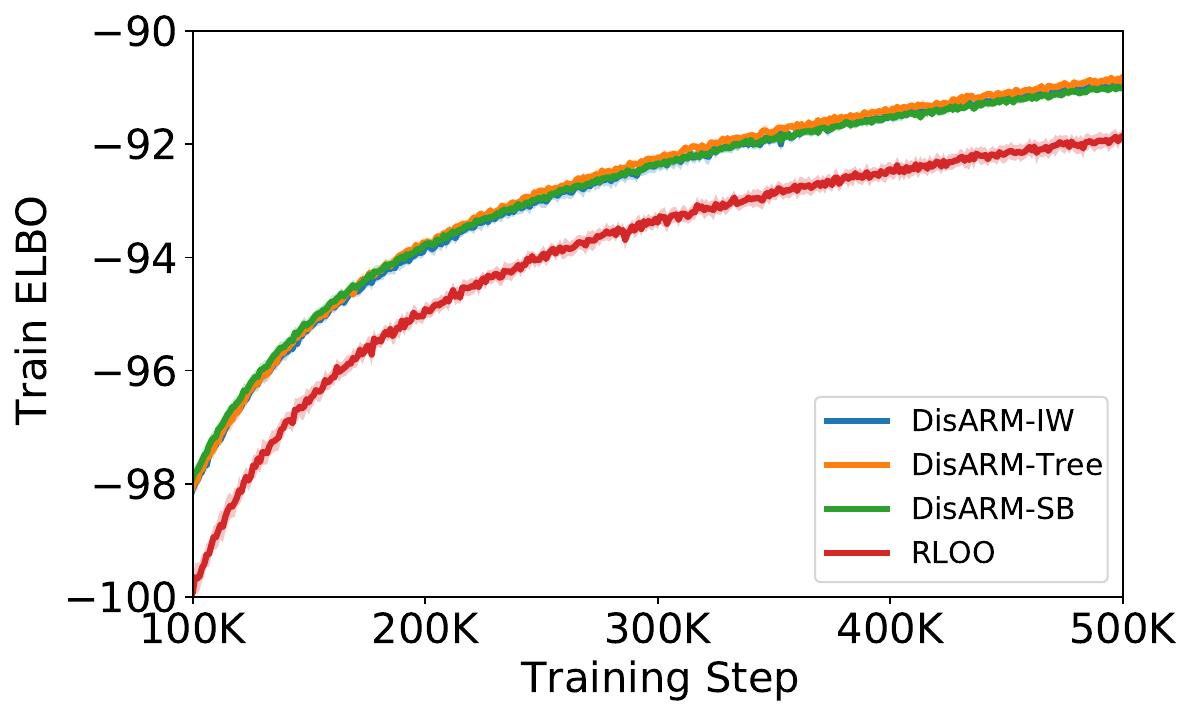}
        \caption{DynamicMNIST}
    \end{subfigure}
    \begin{subfigure}[b]{0.3\textwidth}
        \includegraphics[width=\textwidth]{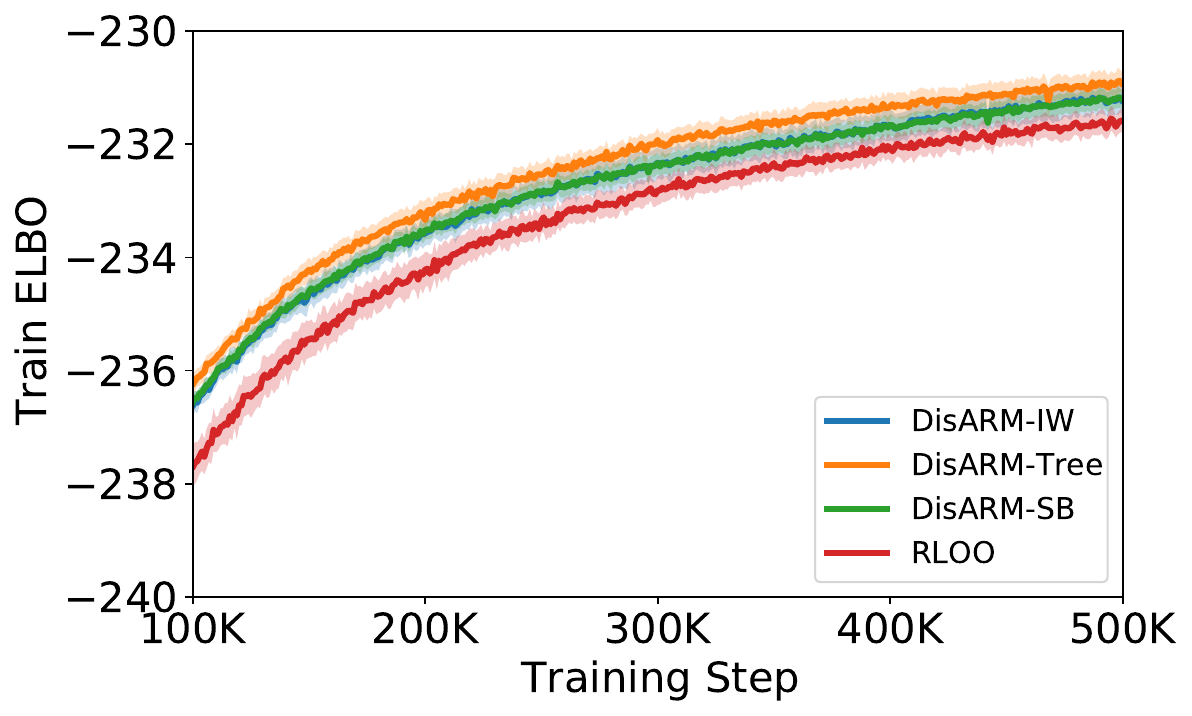}
        \caption{FashionMNIST}
    \end{subfigure}
    \begin{subfigure}[b]{0.3\textwidth}
        \includegraphics[width=\textwidth]{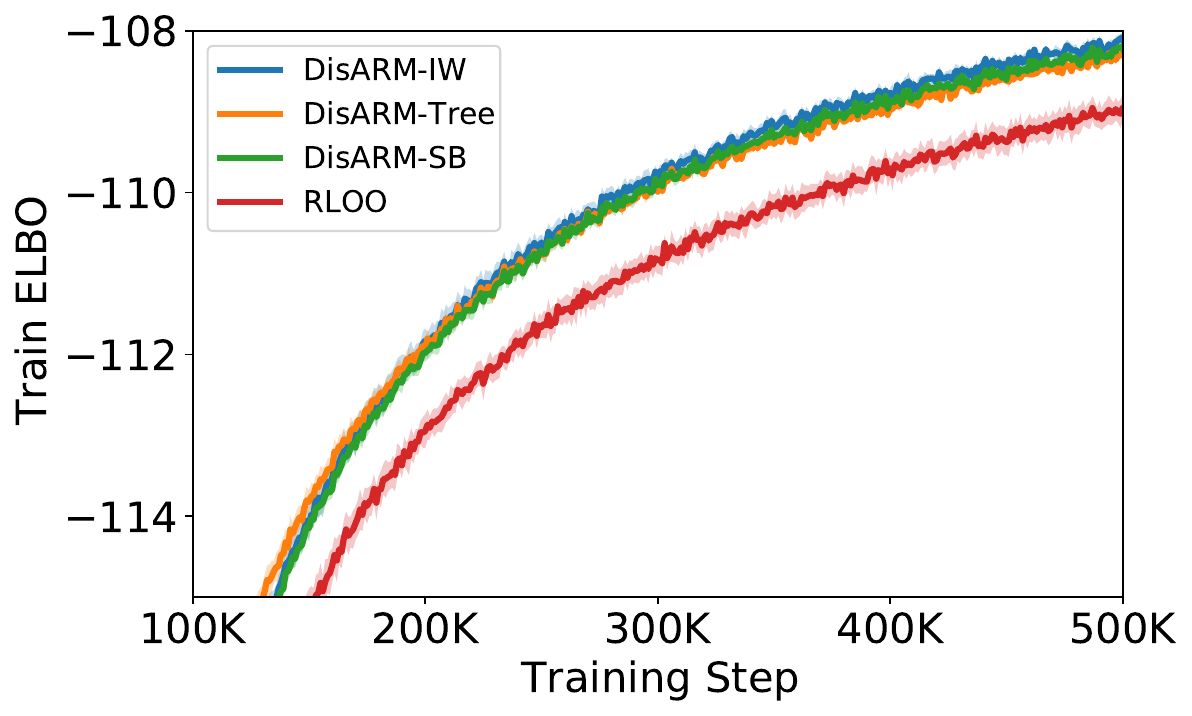}
        \caption{Omniglot}
    \end{subfigure}
    \caption{Training a non-linear categorical VAE with $128$ latent variables and $16$ categories, using multi-sample estimators, on dynamically binarized MNIST, FashionMNIST, Omniglot datasets by maximizing the ELBO. The examples in the top row are using 5 pairs of samples, while the ones in the bottom row are using 10 pairs.}
    \label{fig:experiment-multisample}
\end{figure}

We evaluate the multi-sample extension of the DisARM-* estimators on three benchmark datasets, by training a non-linear categorical VAE. The VAE has a stochastic hidden layer with 128 categorical latent variables, each with 16 categories. We run experiments with 5 antithetic pairs (10 samples) and 10 antithetic pairs (20 samples), and found that the proposed estimators outperform RLOO with a comparable number of independent samples. Even with the increasing number of samples, the performance improvement of the proposed estimators w.r.t. RLOO still holds, as shown in \Cref{fig:experiment-multisample}, Appendix Figure~\ref{fig-appendix:experiment-multisample}, and Appendix Table~\ref{tab:multisample}.

\section{Conclusion}
\label{sec:discussion}
We introduce a novel derivation of DisARM/U2G estimator based on importance sampling and statistical couplings, and naturally extend it to the categorical setting, calling the resulting estimator DisARM-IW. Motivated by the construction of a stick-breaking coupling, we introduce two estimators, DisARM-SB and DisARM-Tree, by reparameterizing the problem with a sequence of binary variables and performing Rao-Blackwellization. 

With systematic experiments, we demonstrate that the proposed estimators provide state-of-the-art performance. We find that the proposed estimators usually perform similarly and all outperform RLOO, with the winner depending on the dataset and the model. As the importance sampling estimator DisARM-IW is simpler to implement, more natural to understand, and easier to generalize, we recommend it in practice. We expect that this estimator can be further improved through better couplings, which is something we intend to explore in the future.

A limitation of the introduced categorical couplings is that they impose structure on the categorical space (i.e., an ordering or tree structure), which is not fully satisfactory because in most settings there is no such natural structure for categorical spaces. Developing coupling-based estimators that do not rely on such a structure would be interesting future work which might lead to further improvements. While we see that our coupling-based estimators generally outperform or perform at least as well as RLOO in our experiments, using coupled samples instead of independent samples is not guaranteed to lead to better performance. Learning the couplings would provide a way of ensuring an improvement over RLOO. Finally, the estimators we propose in this paper, like all multi-sample estimators, can be used for RL only if the environment is simulated or we have a model of it, as they require being able to perform multiple rollouts from the same state.

Discrete latent variables have particular applicability to interpretable models and sparse/conditional computation. Improving the foundational tools to train such models will make them more widely available. While interpretable systems are typically viewed as a positive, they only give a partial view of a complex system, and they can be misused to give a false sense of understanding. While sparse/conditional models can reduce the environmental cost of learned models~\citep{patterson2021carbon}, the unintended implications of large models must also be considered~\citep{bender2021dangers}.

\begin{ack}
We thank Chris J. Maddison and Ben Poole for helpful comments. The authors did not receive any third-party funding or third-party support for this work, and have no financial relationships with entities that could potentially be perceived to influence what they wrote in the submitted work.
\end{ack}

\bibliography{neurips_2021.bib}
\bibliographystyle{apalike}

\newpage
\appendix

\section{Appendix}

\begin{figure}[!h]
    \centering
    \rotatebox[origin=l]{90}{{\scriptsize {\quad \quad DynamicMNIST}}}
    \;
    \includegraphics[width=0.31\linewidth]{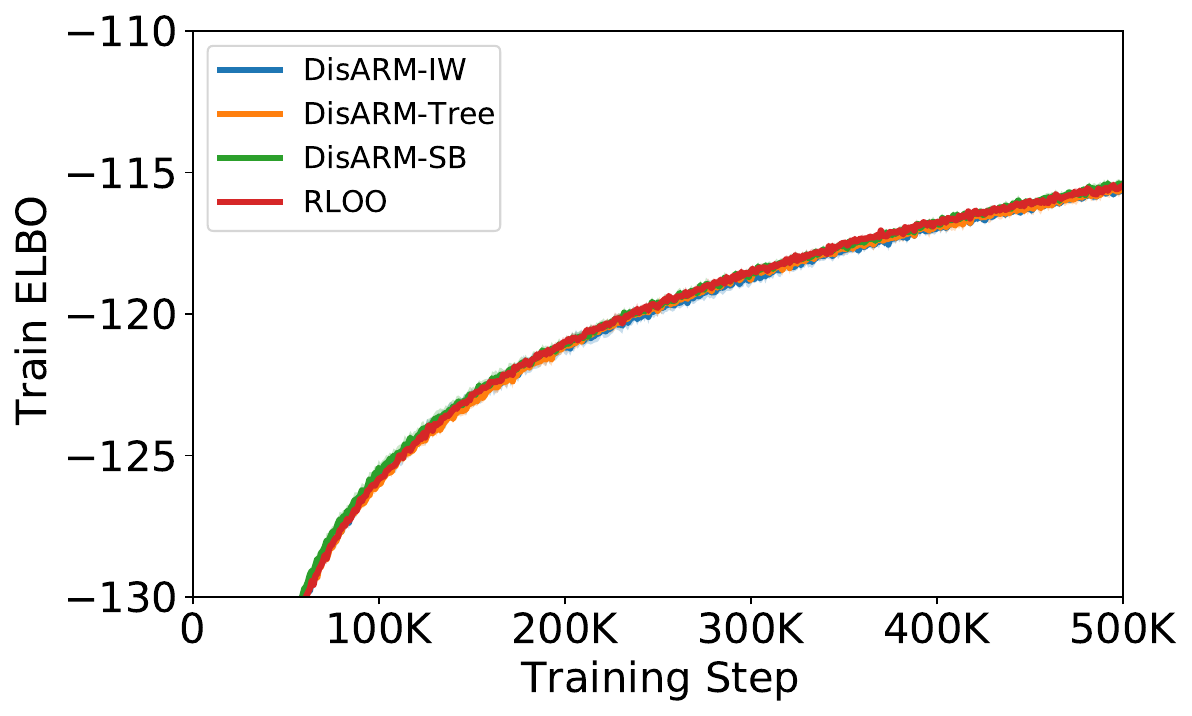}
    \;
    \includegraphics[width=0.31\linewidth]{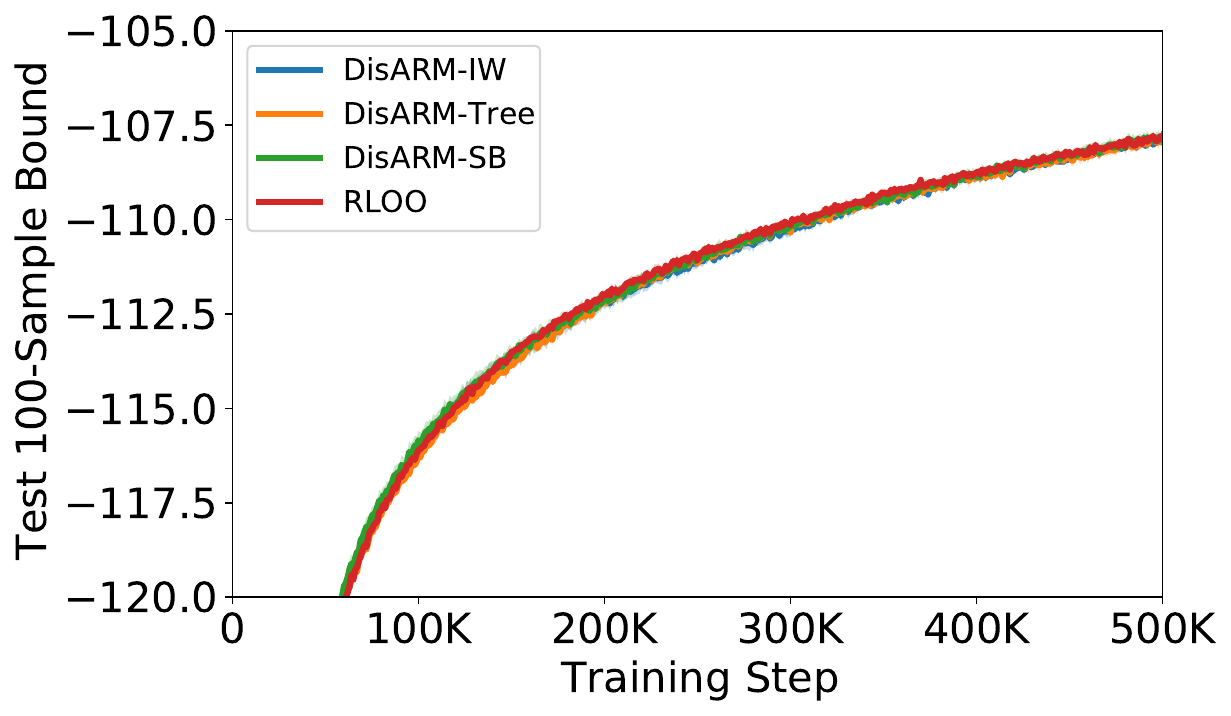}
    \;
    \includegraphics[width=0.29\linewidth]{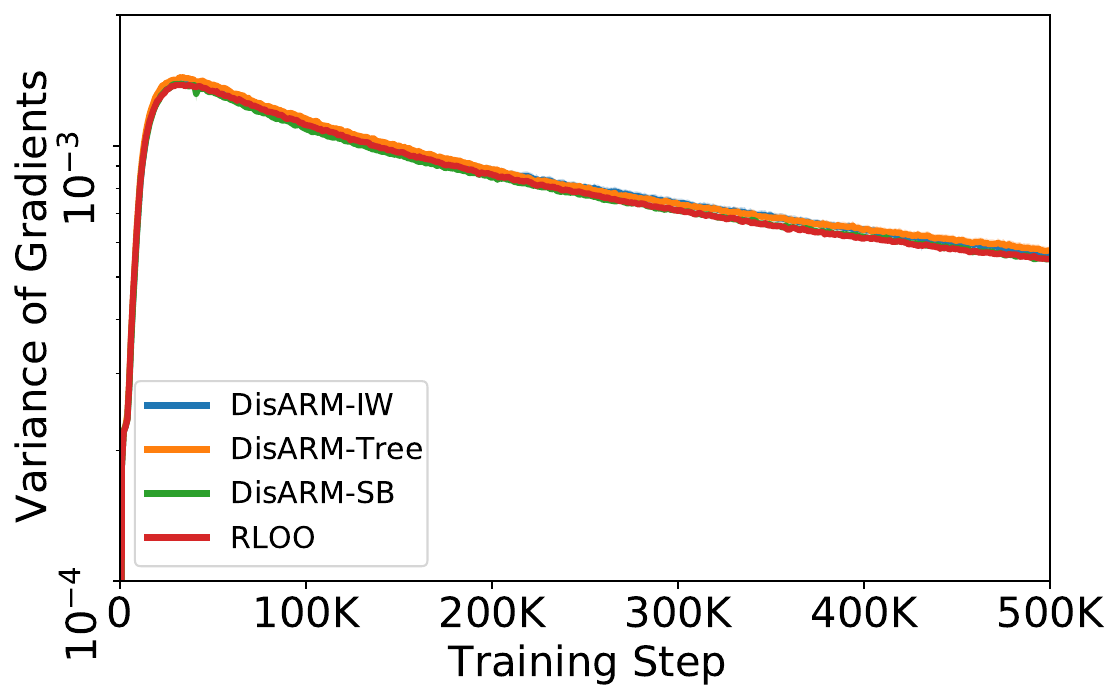}
    \\
    \rotatebox[origin=l]{90}{{\scriptsize {\quad \quad FashionMNIST}}}
    \;
    \includegraphics[width=0.31\linewidth]{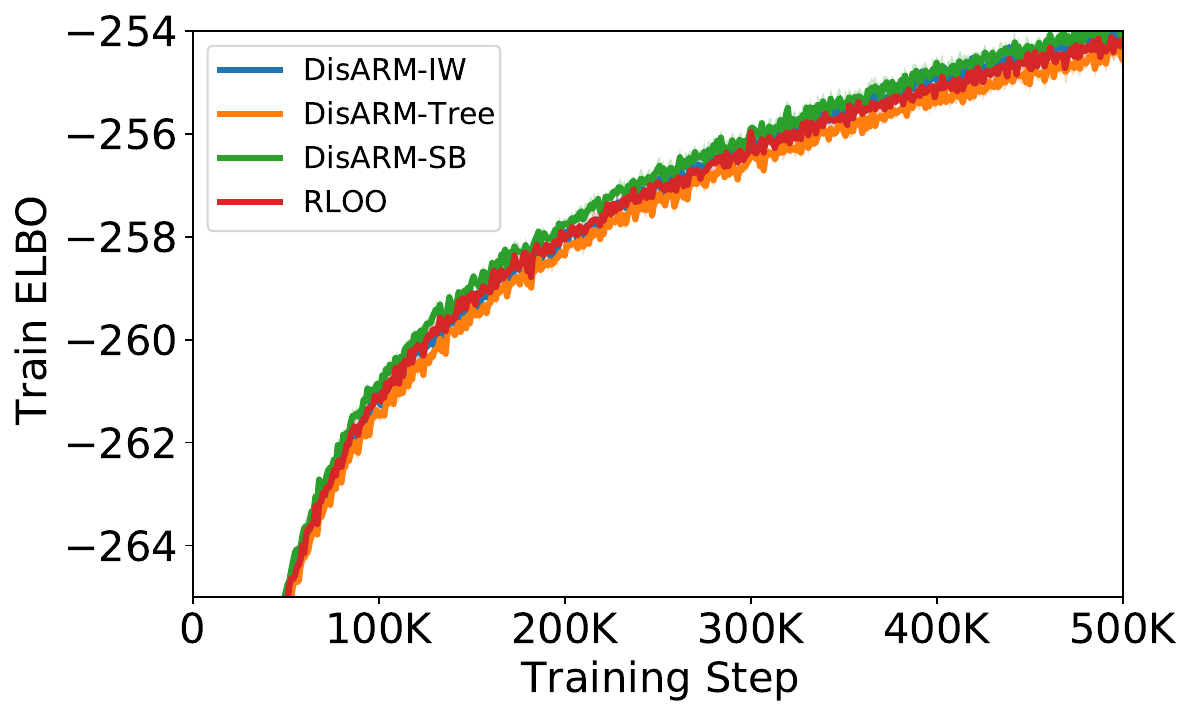}
    \;
    \includegraphics[width=0.31\linewidth]{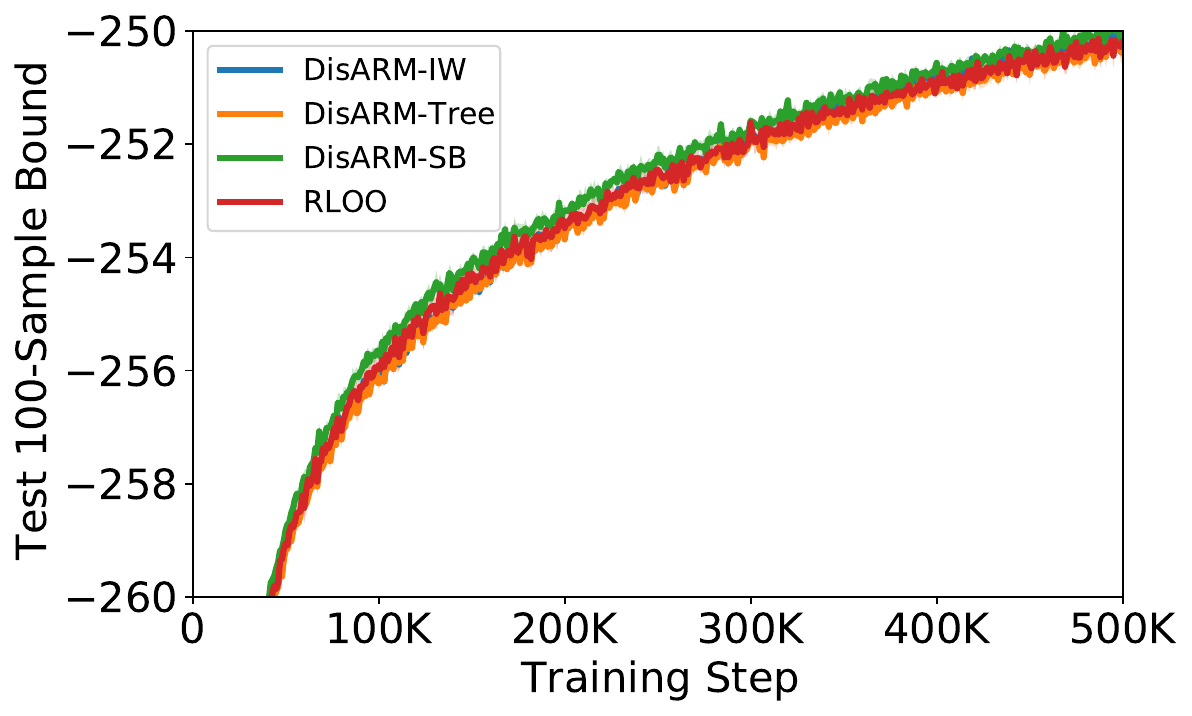}
    \;
    \includegraphics[width=0.29\linewidth]{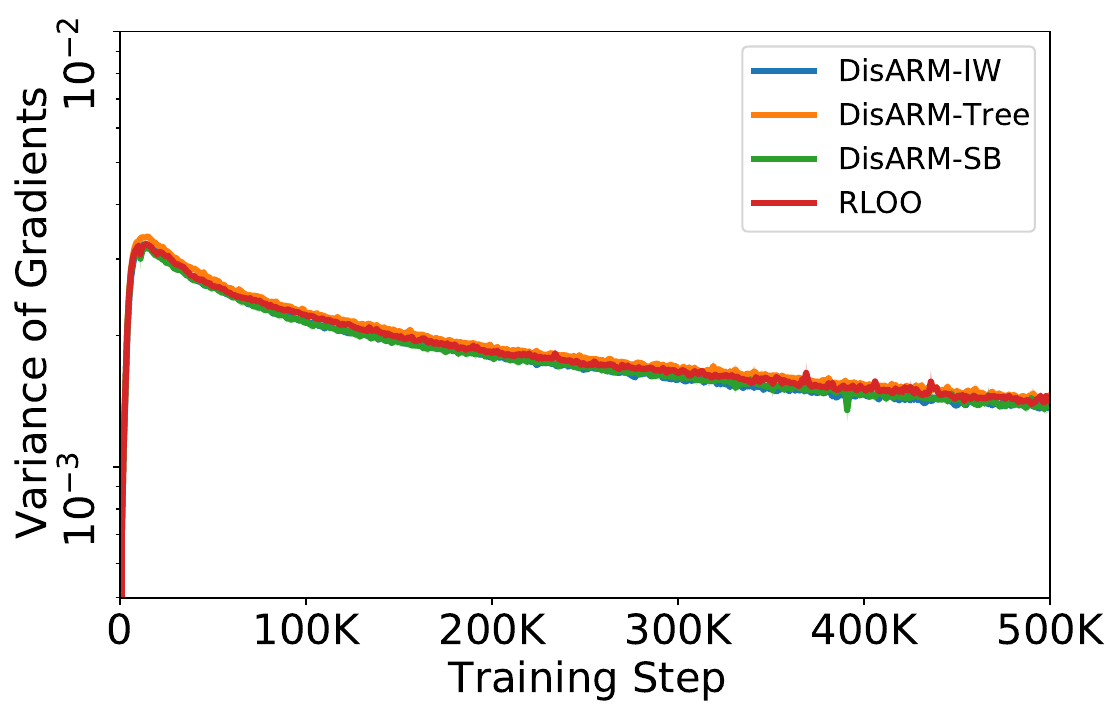}
    \\
    \rotatebox[origin=l]{90}{{\scriptsize {\quad \quad \quad Omniglot}}}
    \;
    \includegraphics[width=0.31\linewidth]{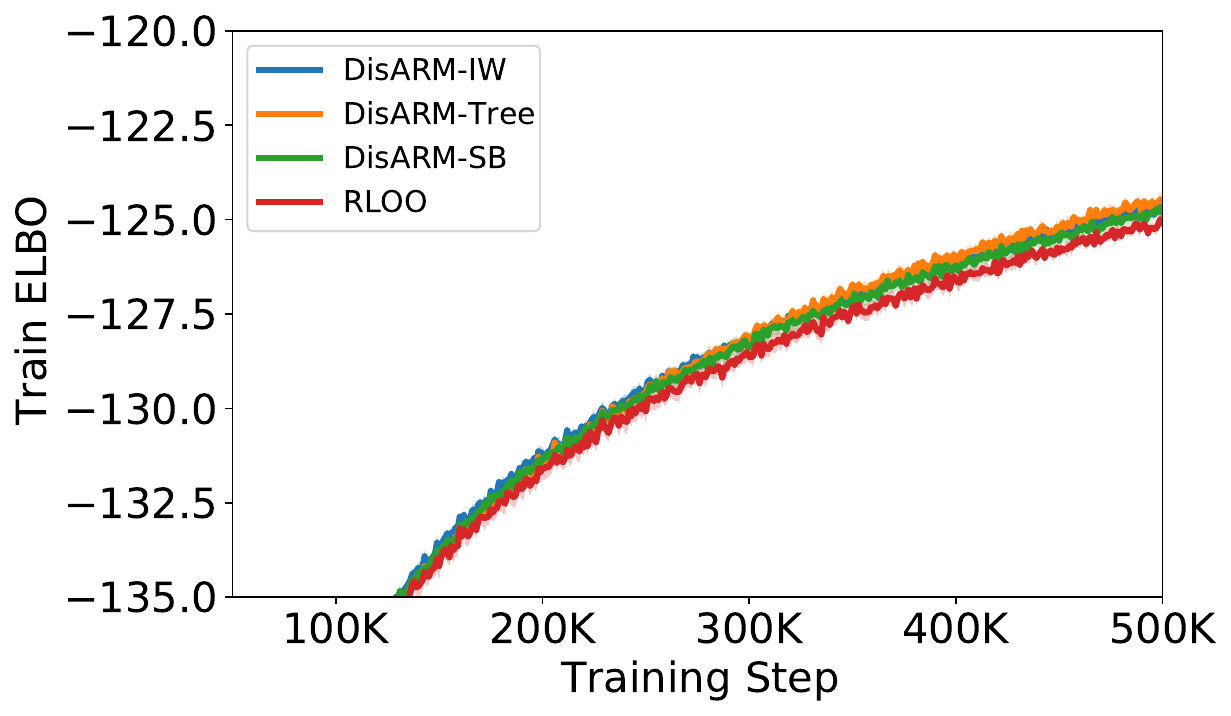}
    \;
    \includegraphics[width=0.31\linewidth]{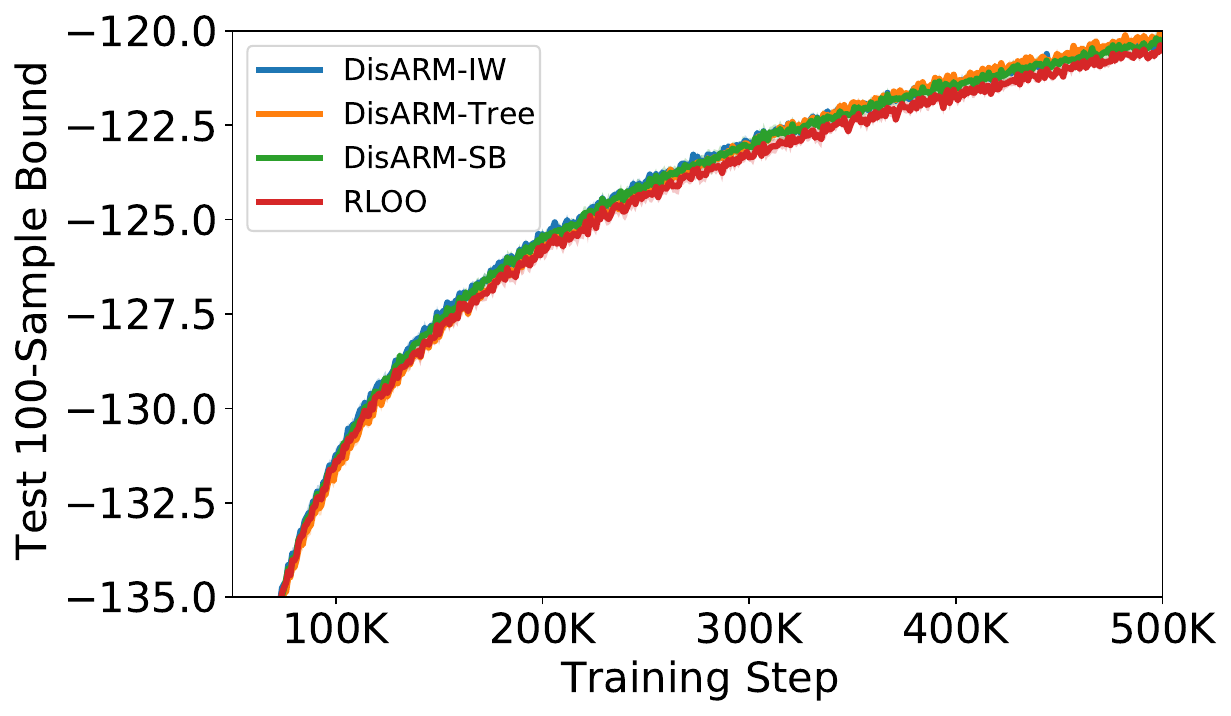}
    \;
    \includegraphics[width=0.29\linewidth]{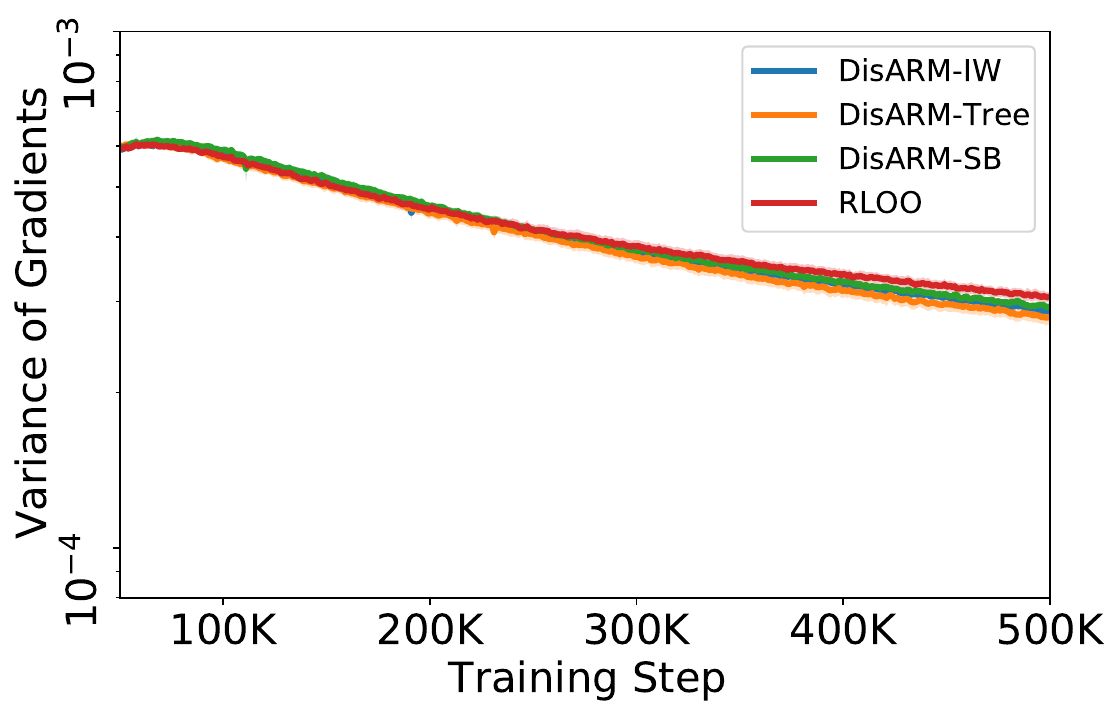}
    \caption{Training a linear VAE with $32$ latent variables with $64$ categories on dynamically binarized MNIST, FashionMNIST, and Omniglot datasets by maximizing the ELBO. We plot the train ELBO (left column), the test 100-sample bound (middle column), and the variance of the gradient estimator (right column). For a fair comparison, the variance of all the gradient estimators was computed along the training trajectory of the RLOO estimator. We plot the mean and one standard error based on $5$ runs from different random initializations.}
    \label{fig-appendix:experiment-linear}
\end{figure}

\begin{figure}[!h]
    \centering
    \rotatebox[origin=l]{90}{{\scriptsize {\quad \quad DynamicMNIST}}}
    \;
    \includegraphics[width=0.31\linewidth]{figures/dynamic_nonlinear_train_eval_C64V20-Ordering_legend_true.pdf}
    \;
    \includegraphics[width=0.31\linewidth]{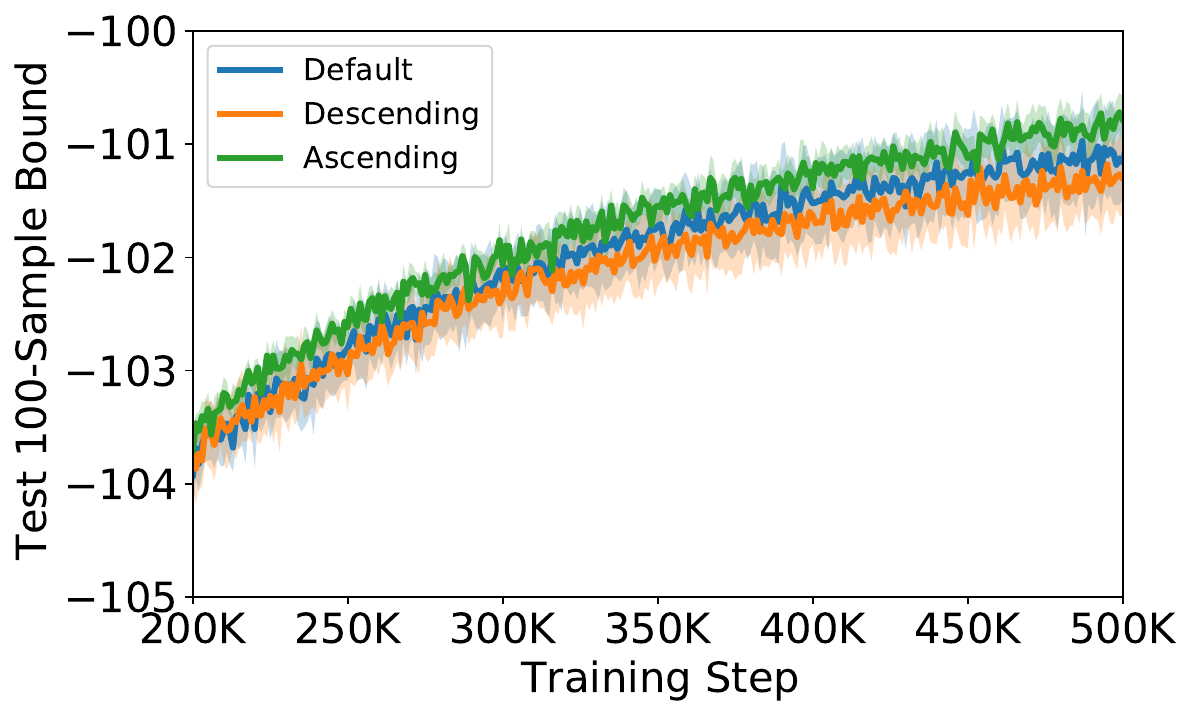}
    \;
    \includegraphics[width=0.29\linewidth]{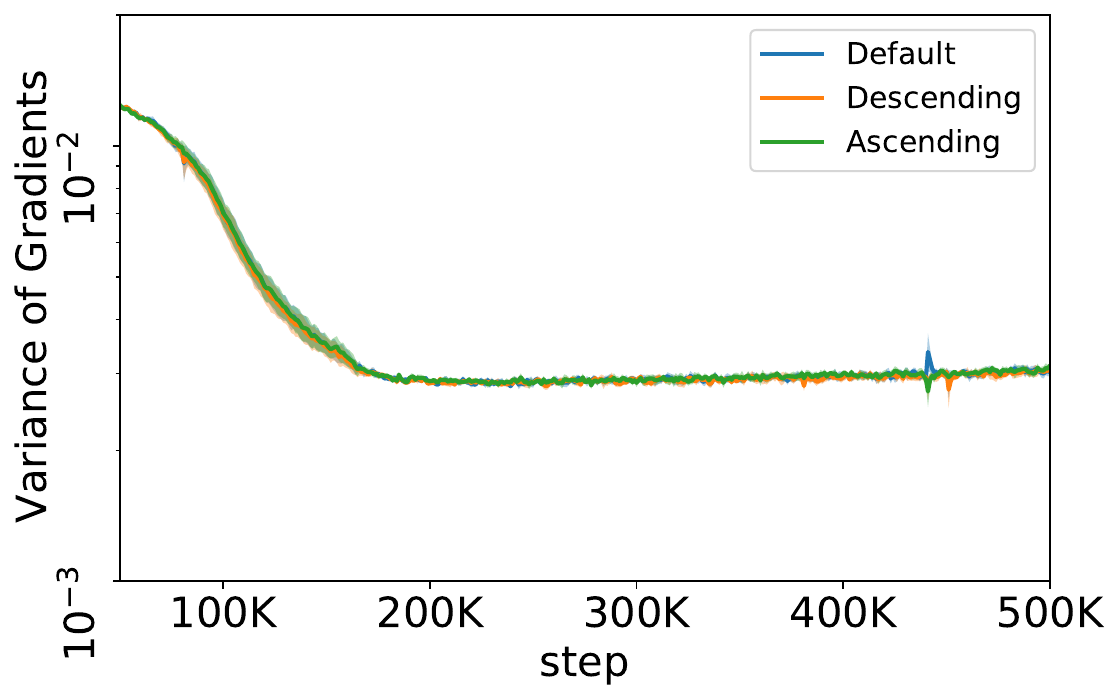}
    \\
    \rotatebox[origin=l]{90}{{\scriptsize {\quad \quad FashionMNIST}}}
    \;
    \includegraphics[width=0.31\linewidth]{figures/fashion_nonlinear_train_eval_C64V20-Ordering_legend_true.pdf}
    \;
    \includegraphics[width=0.31\linewidth]{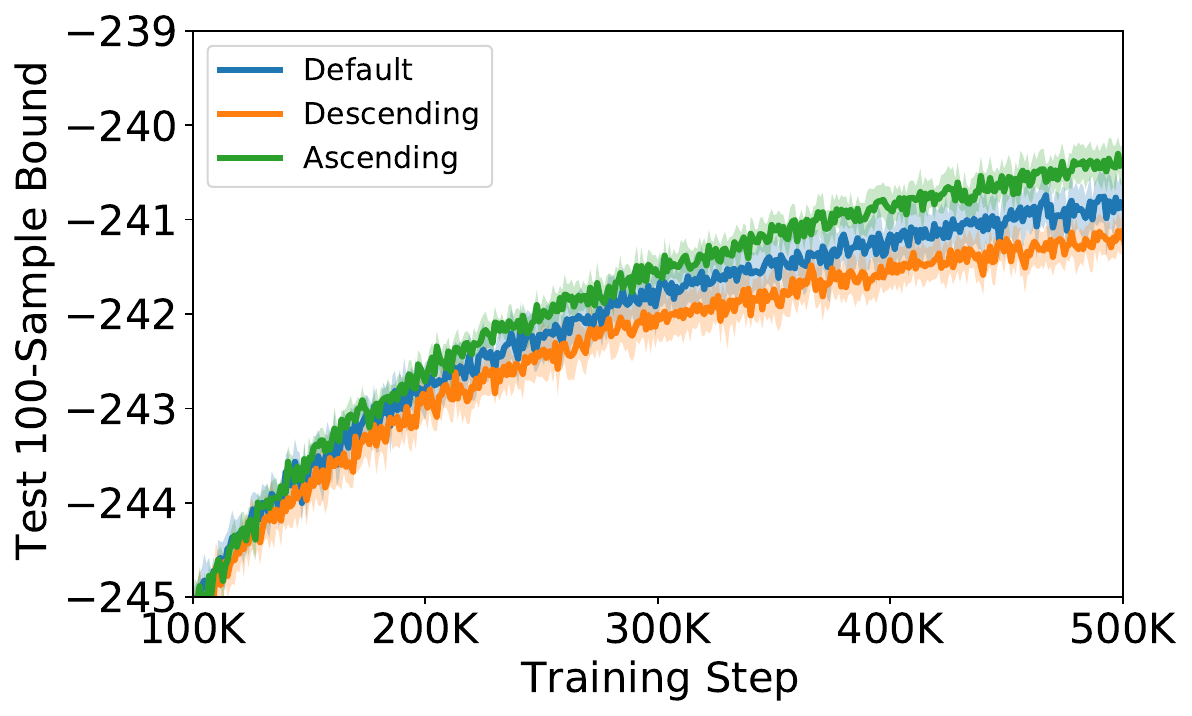}
    \;
    \includegraphics[width=0.29\linewidth]{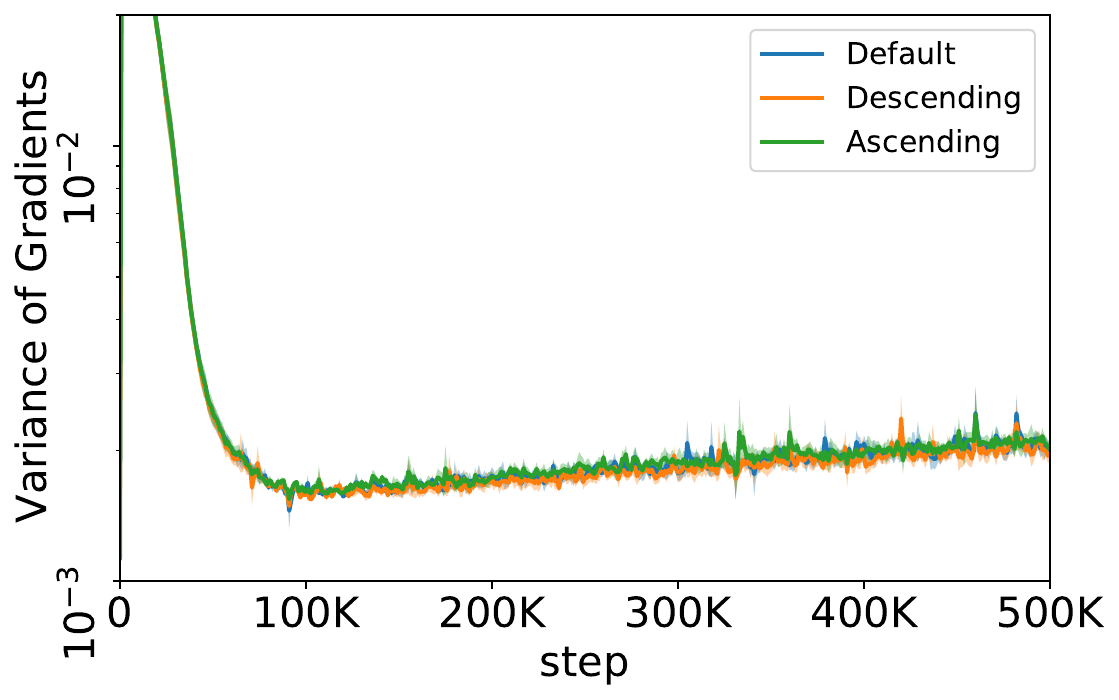}
    \\
    \rotatebox[origin=l]{90}{{\scriptsize {\quad \quad \quad Omniglot}}}
    \;
    \includegraphics[width=0.31\linewidth]{figures/omniglot_nonlinear_train_eval_C64V20-Ordering_legend_true.pdf}
    \;
    \includegraphics[width=0.31\linewidth]{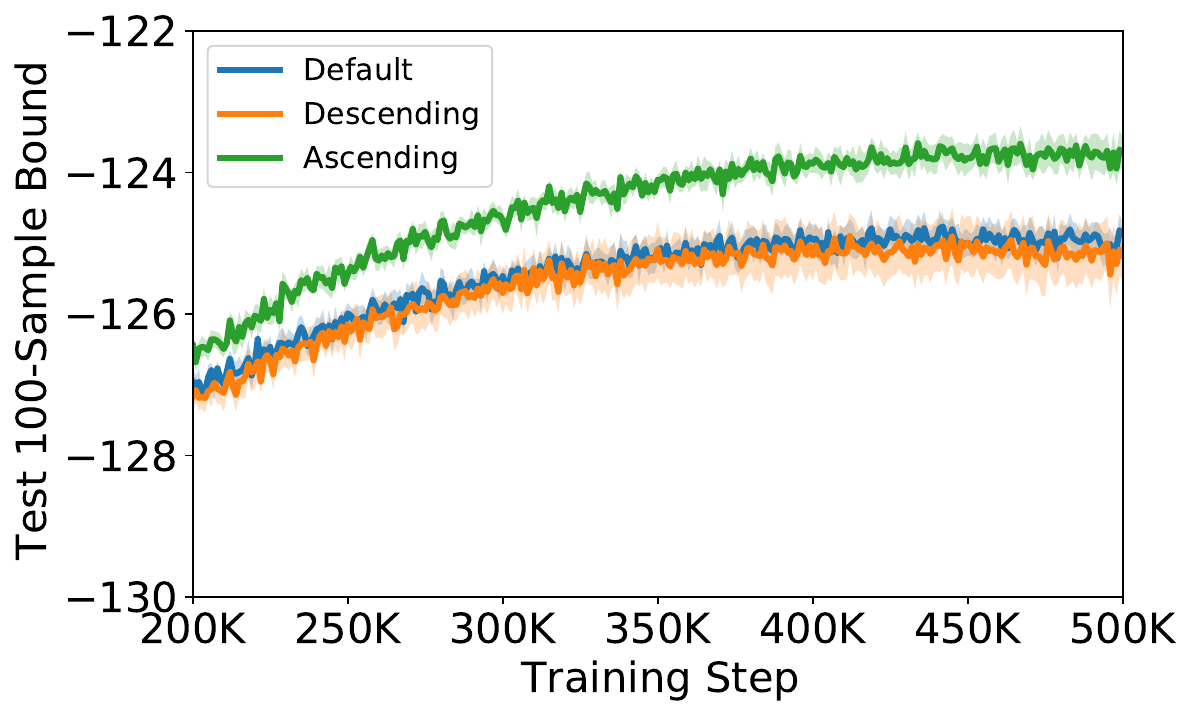}
    \;
    \includegraphics[width=0.29\linewidth]{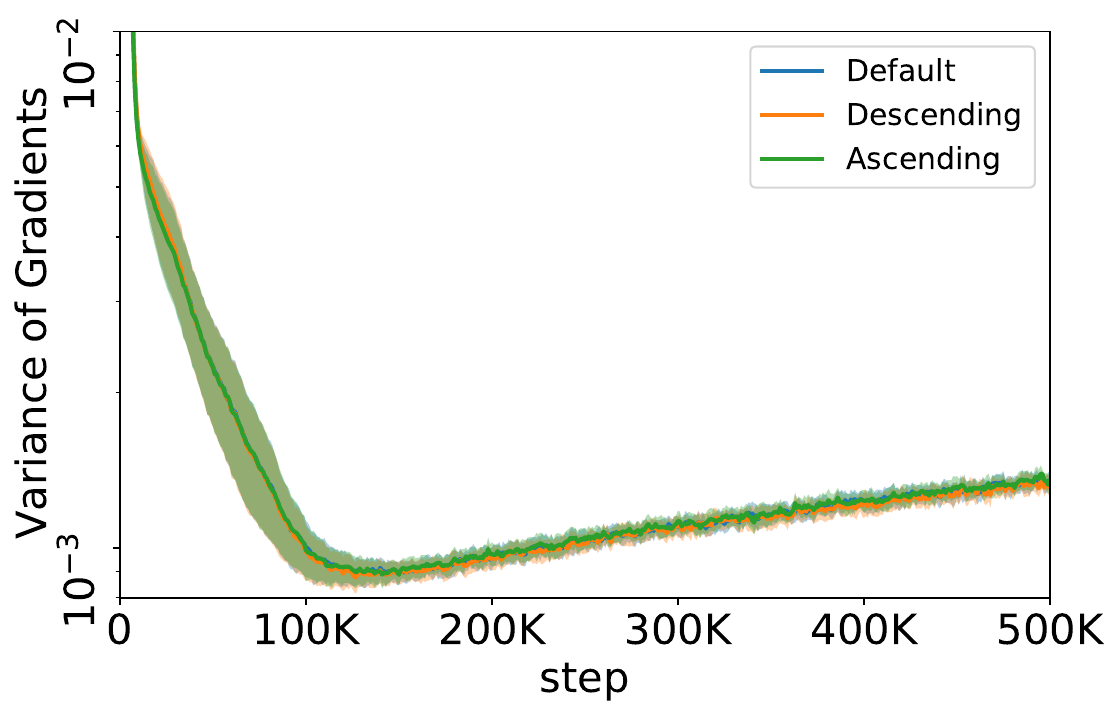}
    \caption{The effect of logit ordering on the performance of DisARM-SB. We sort the encoder logits in the ascending (Green) or descending (Red) order, and compare against the default ordering (Blue).}
    \label{fig-appendix:experiment-ordering}
\end{figure}

\begin{figure}[h]
    \captionsetup[subfigure]{labelformat=empty}
    \centering
    {Comparison of Estimators}\\
    \begin{subfigure}[b]{0.32\textwidth}
        \includegraphics[width=\textwidth]{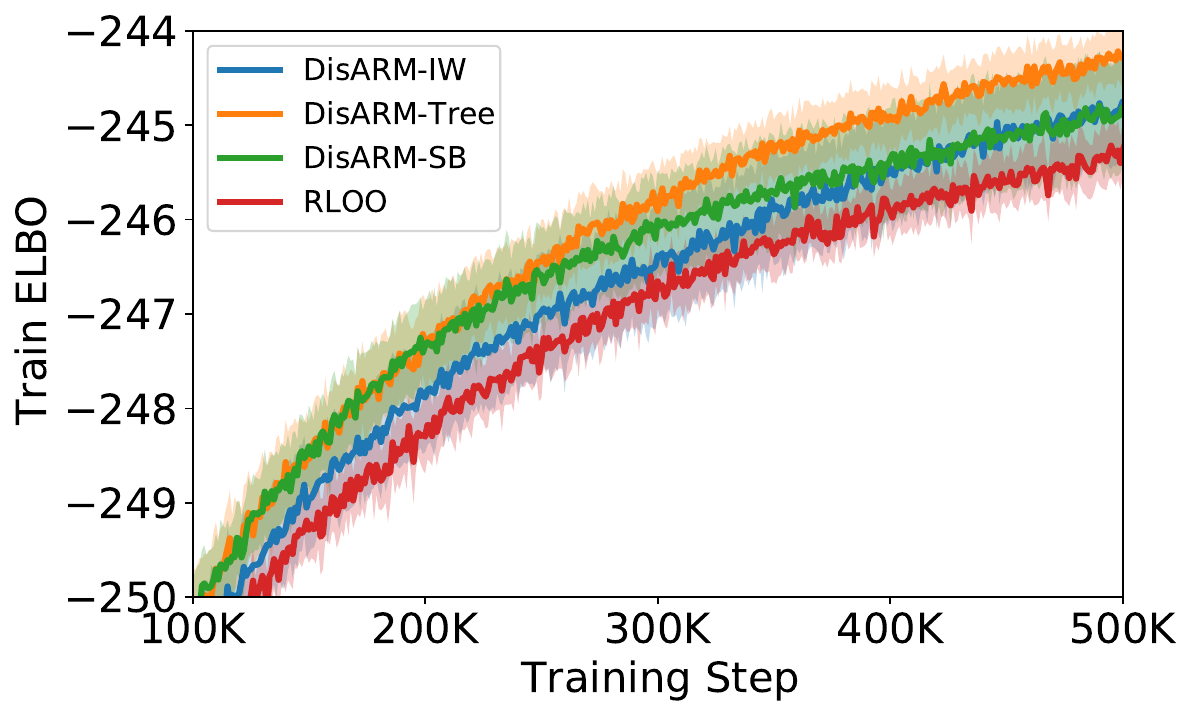}
    \end{subfigure}
    \begin{subfigure}[b]{0.32\textwidth}
        \includegraphics[width=\textwidth]{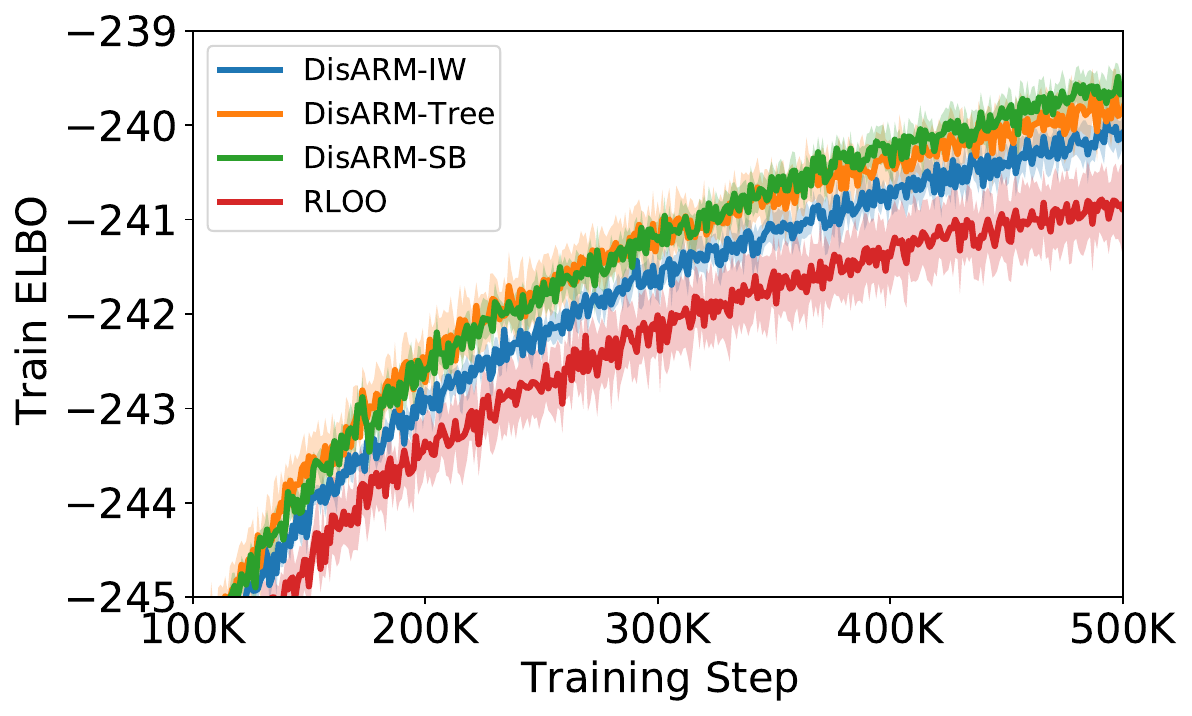}
    \end{subfigure}
    \begin{subfigure}[b]{0.32\textwidth}
        \includegraphics[width=\textwidth]{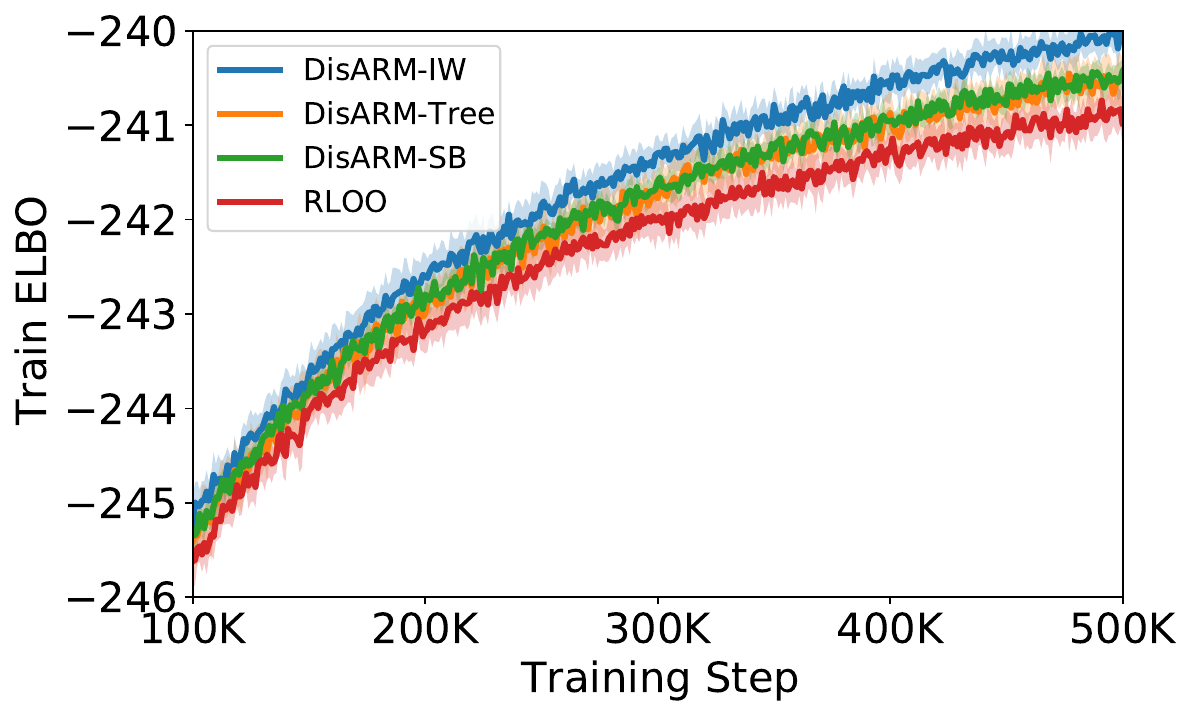}
    \end{subfigure} \\
    {Comparison of Ordering for DisARM-SB}\\
    \begin{subfigure}[b]{0.32\textwidth}
        \includegraphics[width=\textwidth]{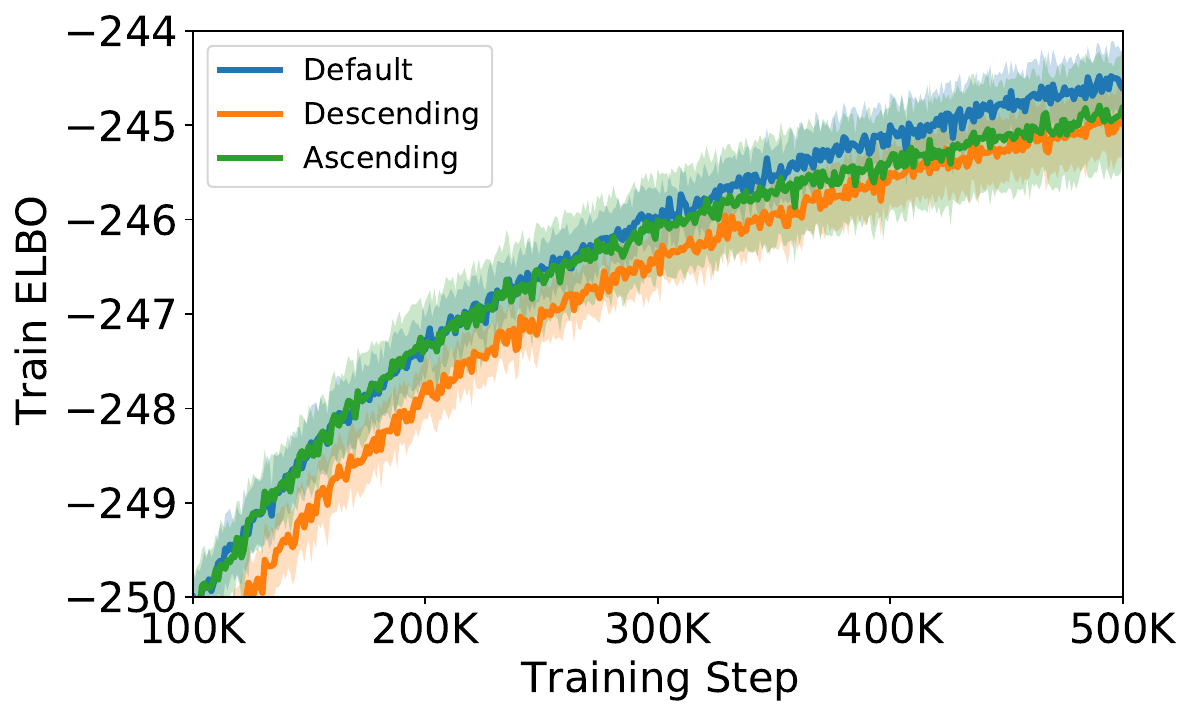}
        \caption{C64/V5}
    \end{subfigure}
    \begin{subfigure}[b]{0.32\textwidth}
        \includegraphics[width=\textwidth]{figures/fashion_nonlinear_train_eval_C64V20-Ordering_legend_true.pdf}
        \caption{C64/V32}
    \end{subfigure}
    \begin{subfigure}[b]{0.32\textwidth}
        \includegraphics[width=\textwidth]{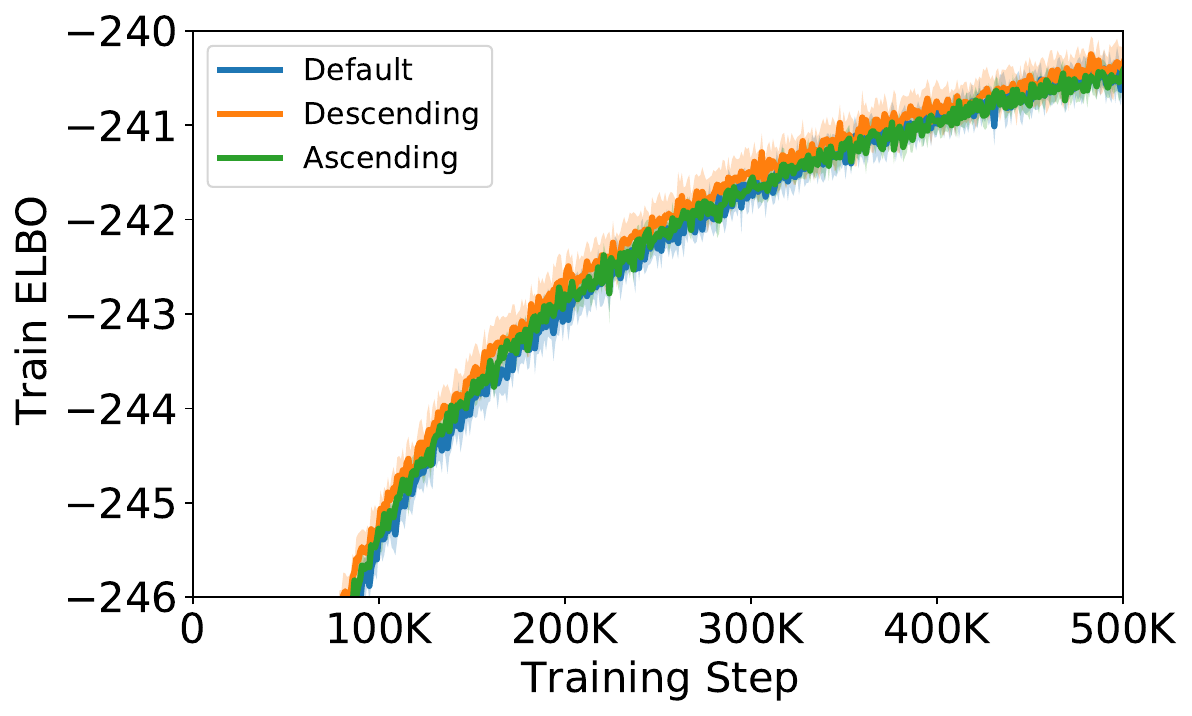}
        \caption{C16/V32}
    \end{subfigure}
    \caption{Training non-linear categorical VAEs with different model sizes on dynamically binarized FashionMNIST. Left: 5 latent variables of 64 categories. Middle: with 32 latent variables with 64 categories. Right: 32 latent variables with 16 categories.}
    \label{fig:experiment-modelsize}
\end{figure}

\begin{figure}[h]
    \captionsetup[subfigure]{labelformat=empty}
    \centering
    {5 Pairs} \\
    \rotatebox[origin=l]{90}{{\scriptsize {\quad \quad \quad Train ELBO}}}
    \;
    \begin{subfigure}[b]{0.3\textwidth}
        \includegraphics[width=\textwidth]{figures/dynamic_train_eval_C16V128-5Pairs.pdf}
    \end{subfigure}
    \begin{subfigure}[b]{0.3\textwidth}
        \includegraphics[width=\textwidth]{figures/fashion_train_eval_C16V128-5Pairs.pdf}
    \end{subfigure}
    \begin{subfigure}[b]{0.3\textwidth}
        \includegraphics[width=\textwidth]{figures/omniglot_train_eval_C16V128-5Pairs.pdf}
    \end{subfigure}
    \\
    \rotatebox[origin=l]{90}{{\scriptsize {\qquad \qquad Variance}}}
    \;
    \begin{subfigure}[b]{0.3\textwidth}
        \includegraphics[width=\textwidth]{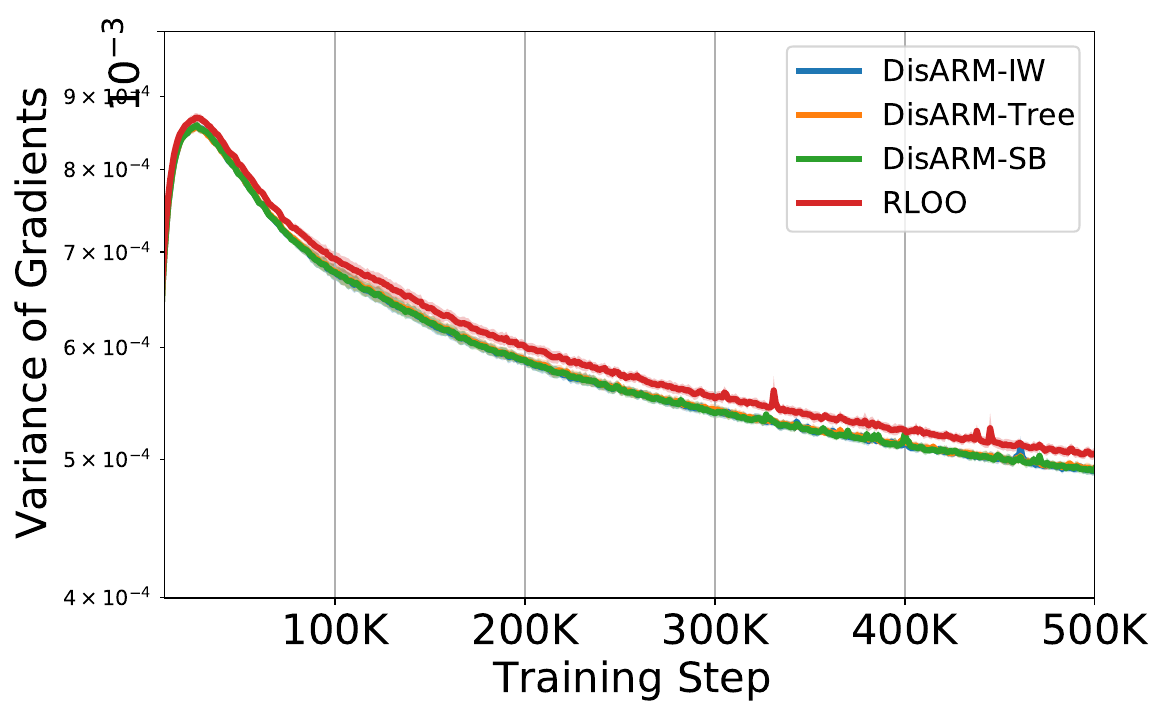}
    \end{subfigure}
    \begin{subfigure}[b]{0.3\textwidth}
        \includegraphics[width=\textwidth]{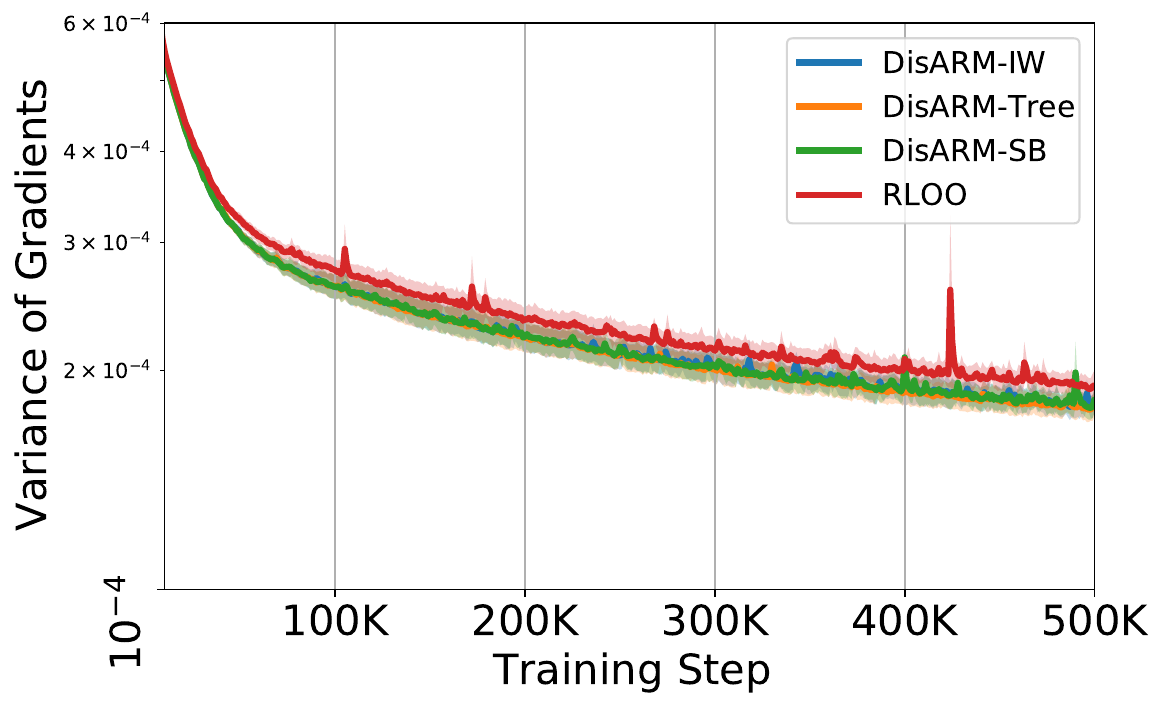}
    \end{subfigure}
    \begin{subfigure}[b]{0.3\textwidth}
        \includegraphics[width=\textwidth]{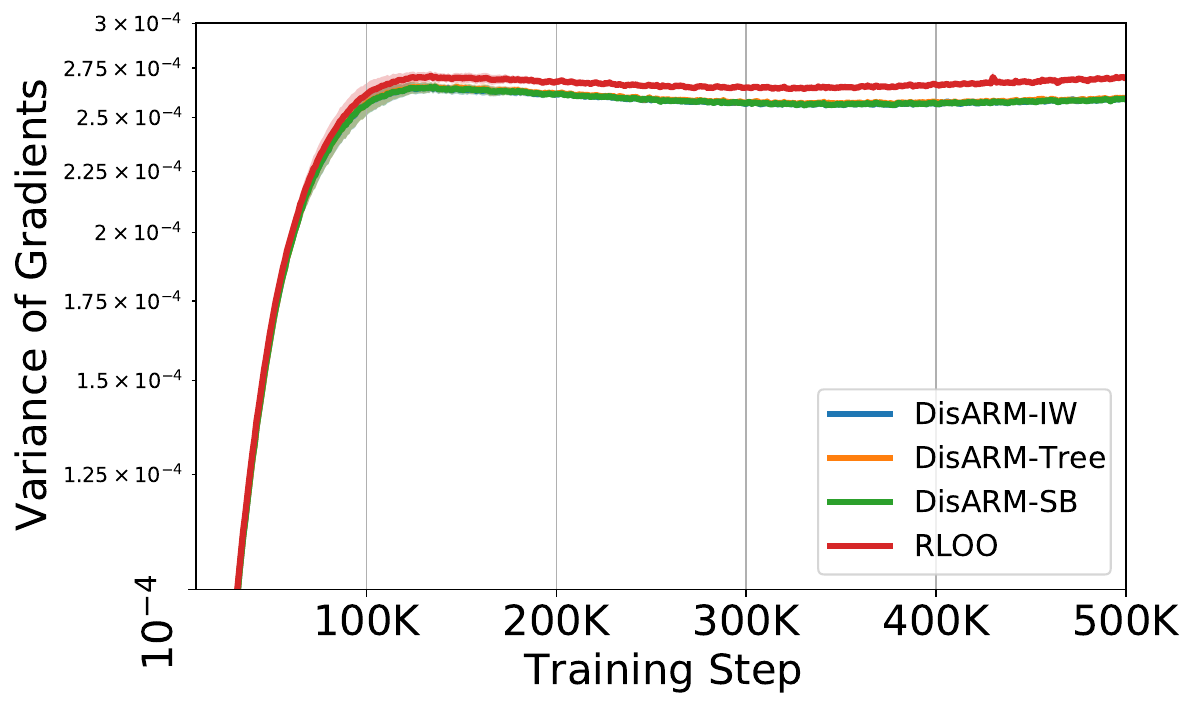}
    \end{subfigure} \\
    \centering
    {10 Pairs} \\
    \rotatebox[origin=l]{90}{{\scriptsize {\quad \quad \quad Train ELBO}}}
    \;
    \begin{subfigure}[b]{0.3\textwidth}
        \includegraphics[width=\textwidth]{figures/dynamic_train_eval_C16V128-10Pairs.pdf}
    \end{subfigure}
    \begin{subfigure}[b]{0.3\textwidth}
        \includegraphics[width=\textwidth]{figures/fashion_train_eval_C16V128-10Pairs.pdf}
    \end{subfigure}
    \begin{subfigure}[b]{0.3\textwidth}
        \includegraphics[width=\textwidth]{figures/omniglot_train_eval_C16V128-10Pairs.pdf}
    \end{subfigure}
    \\
    \rotatebox[origin=l]{90}{{\scriptsize {\qquad \qquad \qquad Variance}}}
    \;
    \begin{subfigure}[b]{0.3\textwidth}
        \includegraphics[width=\textwidth]{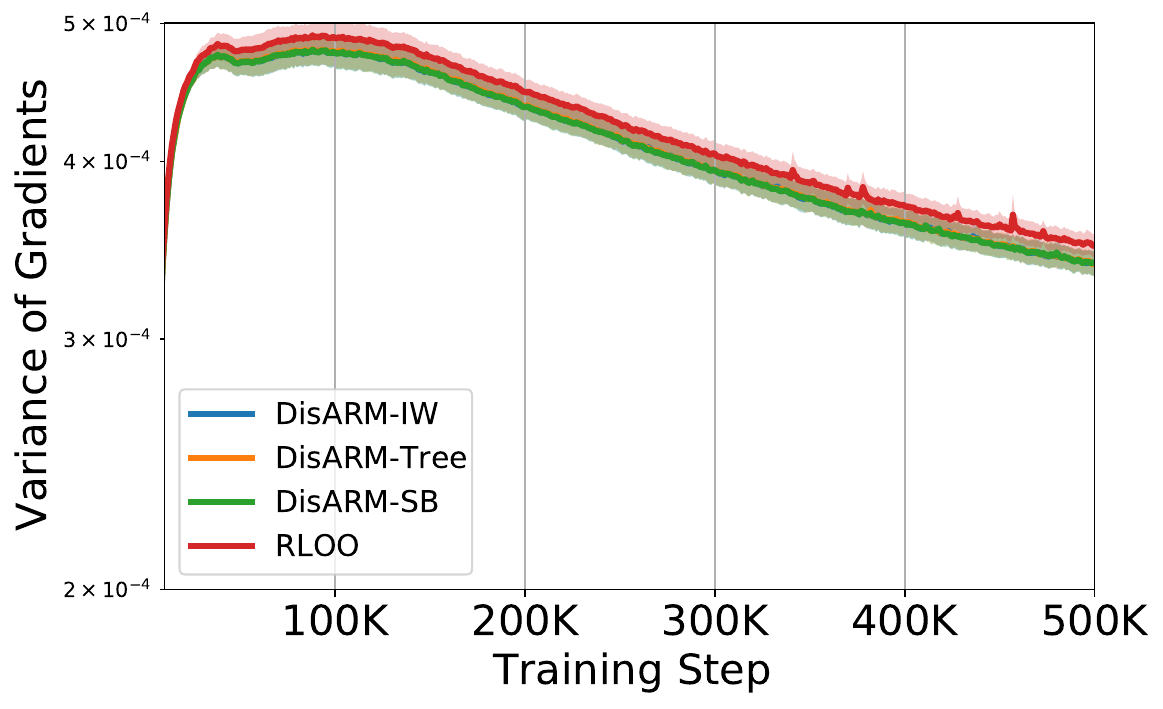}
        \caption{DynamicMNIST}
    \end{subfigure}
    \begin{subfigure}[b]{0.3\textwidth}
        \includegraphics[width=\textwidth]{figures/fashion_C16V128-10Pairs_variance.pdf}
        \caption{FashionMNIST}
    \end{subfigure}
    \begin{subfigure}[b]{0.3\textwidth}
        \includegraphics[width=\textwidth]{figures/omniglot_C16V128-10Pairs_variance.pdf}
        \caption{Omniglot}
    \end{subfigure}
    \caption{Training a non-linear categorical VAE with $128$ latent variables and $16$ categories, using multi-sample objectives, on dynamically binarized MNIST, FashionMNIST, Omniglot datasets by maximizing the ELBO. The examples in the top 2 rows are using 5 pairs of samples, while the ones in the bottom 2 rows are using 10 pairs.}
    \label{fig-appendix:experiment-multisample}
\end{figure}

\begin{table}[h]
\caption{Training a non-linear VAE with categorical latents using multi-sample estimators. Mean variational lower bounds and the standard error of the mean computed based on 5 runs of $5\times 10^5$-steps training from different random initializations. The best performing method (up to the standard error) for each task is in bold.  } 
\label{tab:multisample}
\begin{adjustbox}{center}{
    \footnotesize
    \begin{tabular}{c l c c c c}
    \toprule
    {\#Pairs} & {Dataset} &     DisARM-Tree & DisARM-SB &      DisARM-IW &        RLOO \\
    \midrule
    \multirow{3}{*}{5} 
    & Dynamic MNIST &   ${\bf -91.36\pm0.11}$ &   $-91.67\pm0.08$ &   $-91.85\pm0.04$ &   $-92.35\pm0.10$ \\
    & Fashion MNIST  &  ${\bf -231.01\pm0.14}$ &  $-231.38\pm0.14$ &  ${\bf -231.14\pm0.17}$ &  $-231.78\pm0.16$ \\
    & Omniglot &  ${\bf -109.00\pm0.10}$ &  ${\bf -108.90\pm0.09}$ &  ${\bf -108.83\pm0.11}$ &  $-109.73\pm0.07$ \\
    \midrule
    \multirow{3}{*}{10} 
    & Dynamic MNIST  &   ${\bf -90.81\pm0.06}$ &   ${\bf -91.01\pm0.06}$ &   ${\bf -90.95\pm0.08}$ &   $-91.86\pm0.15$ \\
    & Fashion MNIST  &  ${\bf -230.95\pm0.21}$ &  $-231.21\pm0.18$ &  $-231.25\pm0.29$ &  $-231.59\pm0.21$ \\
    & Omniglot &  $-108.27\pm0.03$ &  $-108.19\pm0.07$ &  ${\bf -108.08\pm0.06}$ &  $-108.95\pm0.17$ \\
    \bottomrule
    \end{tabular}}
\end{adjustbox}
\end{table}

\subsection{Experiments with linear categorical VAEs}
\label{app:experiment}
We evaluate the three proposed gradient estimators, DisARM-IW, DisARM-SB, and  DisARM-Tree, by training linear variational auto-encoders with categorical latent variables on dynamically binarized MNIST, FashionMNIST, and Omniglot datasets. As in~\citep{dong2020disarm}, we benchmark the proposed estimators against the 2-sample REINFORCE estimator with the leave-one-out baseline~\citep[RLOO;][]{kool2019buy}. The linear model has a single layer of $32$ categorical latent variables, each with $64$ categories. We find no significant difference in performance between the proposed estimators and the RLOO baseline in this setting (Appendix Table~\ref{tab-appendix:exp_comparison} and Appendix Figure~\ref{fig-appendix:experiment-linear}).  However, as we noted in the maintext, \citet{dong2020disarm} found that for multi-layer linear models in the binary case, DisARM showed increasing improvement over RLOO for models with deeper hierarchies. So it would be interesting to see whether this holds for the categorical case in future work. 

\begin{table}[h]
\caption{Training a linear VAE with categorical latents using the proposed estimators and the RLOO baseline. Mean variational lower bounds and the standard error of the mean computed based on 5 runs of $5\times 10^5$-steps training from different random initializations. The best performing method (up to the standard error) for each task is in bold.  } 
\label{tab-appendix:exp_comparison}
\begin{adjustbox}{center}
{\footnotesize
\begin{tabular}{l cccc}
\toprule
\multicolumn{5}{c}{Training set ELBO}\\
\midrule
\midrule
{} & DisARM-IW &  DisARM-Tree & DisARM-SB & RLOO \\
\midrule
Dynamic MNIST &  $-115.64\pm0.09$ &  $\bf{-115.59\pm0.16}$ &   $\bf{-115.48\pm0.14}$ &  $\bf{-115.50\pm0.18}$ \\
Fashion MNIST &  $\bf{-254.23\pm0.19}$ &  $-254.56\pm0.22$ &   $\bf{-254.05\pm0.20}$ &  $\bf{-254.18\pm0.15}$ \\
Omniglot &  $-124.64\pm0.02$ &  $\bf{-124.52\pm0.11}$ &   $-124.82\pm0.21$ &  $-125.10\pm0.11$ \\
\bottomrule
\end{tabular}
}
\end{adjustbox}
\end{table}

\subsection{Experimental Details}
\label{app:exp-details}
We use the same model structure as in~\citep{yin2019arsm}.
The model has a single layer of categorical latent variables which are mapped to Bernoulli logits using an MLP with two hidden layers of $256$ and $512$ of LeakyReLU units~\citep{xu2015leakyrelu} with negative slope of $0.2$. The encoder mirrors the structure, having two hidden layers of $512$ and $256$ LeakyReLU units.

For a fair comparison of the variance of the gradient estimators, we train a model with the RLOO estimator and evaluate the variance of all the estimators at each step along the training trajectory of this model. Based on preliminary experiments, the results were independent of the gradient estimator used to generate the model trajectory. We report the average per-parameter variance based on the parameter moments estimated with an exponential moving average with decay rate 0.999.

Each experiment run takes around $12$ hours on an NVIDIA Tesla P100 GPU. Our implementation was biased towards readability instead of computational efficiency, so we expect that significant improvements in runtime could be achieved.

\subsection{Importance Weighting Derivation}
\label{app:iw-derivation}
We consider proposal distributions that factorize across dimensions $p(z, \tilde{z}) = \prod_k p(z_k, \tilde{z}_k)$ and that are couplings such that the marginals are maintained $p(z_k) = p(\tilde{z_k}) = q(z_k)$. In general, $q_\theta(z)q_\theta(\tilde{z})$ has full support, but we know that the integrand $q_\theta(z)q_\theta(\tilde{z})(f(z) - f(\tilde{z}))\left(\nabla_{\theta_k} \log q_\theta(z_k) - \nabla_{\theta_k} \log q_\theta(\tilde{z}_k) \right)$ vanishes when $z_k = \tilde{z}_k$ by inspection, so we allow $p(z_k, \tilde{z}_k)$ to put zero mass on $z_k = \tilde{z}_k$ configurations, and require $p(z_k, \tilde{z}_k) > 0$ otherwise.

Then with $z, \tilde{z} \sim p$, we show that $g_\text{DisARM-IW}$ (Eq.~\ref{eq:iw_estimator}) is an unbiased estimator. First,
\begin{align*}
\E_{p(z, \tilde{z})}&\left[\frac{1}{2}\frac{q_\theta(z_k)q_\theta(\tilde{z_k})}{p_\theta(z_k, \tilde{z_k})}(f(z) - f(\tilde{z}))\left(\nabla_{\theta_k} \log q_\theta(z_k) - \nabla_{\theta_k} \log q_\theta(\tilde{z}_k) \right)\right] \\
&= \E_{p(z_{-k}, \tilde{z}_{-k})}\E_{p(z_k, \tilde{z}_k)}\left[\frac{1}{2}\frac{q_\theta(z_k)q_\theta(\tilde{z_k})}{p_\theta(z_k, \tilde{z_k})}(f(z) - f(\tilde{z}))\left(\nabla_{\theta_k} \log q_\theta(z_k) - \nabla_{\theta_k} \log q_\theta(\tilde{z}_k) \right)\right] \\
&= \E_{p(z_{-k}, \tilde{z}_{-k})}\sum_{z_k, \tilde{z}_k \in \supp p(z_k, \tilde{z}_k)}\frac{1}{2}q_\theta(z_k)q_\theta(\tilde{z_k})(f(z) - f(\tilde{z}))\left(\nabla_{\theta_k} \log q_\theta(z_k) - \nabla_{\theta_k} \log q_\theta(\tilde{z}_k) \right) \\
&= \E_{p(z_{-k}, \tilde{z}_{-k})}\sum_{z_k, \tilde{z}_k}\frac{1}{2}q_\theta(z_k)q_\theta(\tilde{z_k})(f(z) - f(\tilde{z}))\left(\nabla_{\theta_k} \log q_\theta(z_k) - \nabla_{\theta_k} \log q_\theta(\tilde{z}_k) \right) \\
&= \E_{p(z_{-k}, \tilde{z}_{-k})}\E_{q_\theta(z_k)q_\theta(\tilde{z}_k)}\left[\frac{1}{2}(f(z) - f(\tilde{z}))\left(\nabla_{\theta_k} \log q_\theta(z_k) - \nabla_{\theta_k} \log q_\theta(\tilde{z}_k) \right)\right],
\end{align*}
using the fact that the integrand vanishes outside the support of $p(z_k, \tilde{z}_k)$. Then,
\begin{align*}
\E_{p(z_{-k}, \tilde{z}_{-k})}&\E_{q_\theta(z_k)q_\theta(\tilde{z}_k)}\left[\frac{1}{2}(f(z) - f(\tilde{z}))\left(\nabla_{\theta_k} \log q_\theta(z_k) - \nabla_{\theta_k} \log q_\theta(\tilde{z}_k) \right)\right] \\
&= \E_{q_\theta(z_k)q_\theta(\tilde{z}_k)}\left[\frac{1}{2}(\E_{p(z_{-k}, \tilde{z}_{-k})}\left[f(z) - f(\tilde{z})\right])\left(\nabla_{\theta_k} \log q_\theta(z_k) - \nabla_{\theta_k} \log q_\theta(\tilde{z}_k) \right)\right] \\
&= \E_{q(z_k)q(\tilde{z}_k)}\left[\frac{1}{2}(\E_{p(z_{-k}, \tilde{z}_{-k})}\left[f(z)\right] - \E_{p(z_{-k}, \tilde{z}_{-k})}\left[f(\tilde{z})\right])\left(\nabla_{\theta_k} \log q_\theta(z_k) - \nabla_{\theta_k} \log q_\theta(\tilde{z}_k) \right)\right] \\
&= \E_{q(z_k)q(\tilde{z}_k)}\left[\frac{1}{2}(\E_{q(z_{-k})}\left[f(z)\right] - \E_{q(\tilde{z}_{-k})}\left[f(\tilde{z})\right])\left(\nabla_{\theta_k} \log q_\theta(z_k) - \nabla_{\theta_k} \log q_\theta(\tilde{z}_k) \right)\right] \\
&= \E_{q(z)q(\tilde{z})}\left[\frac{1}{2}(f(z) - f(\tilde{z}))\left(\nabla_{\theta_k} \log q_\theta(z_k) - \nabla_{\theta_k} \log q_\theta(\tilde{z}_k) \right)\right],
\end{align*}
following from the linearity of the expectation and the fact that the coupling preserves marginals.

\subsection{Stick-Breaking Coupling}
\label{app:stick-breaking}
\subsubsection{Computing Importance Weights}
To simplify the notation in this subsection, we consider a single dimension at a time and omit the dimension index. To compute $\frac{q_\theta(z)q_\theta(\tilde{z})}{p_\theta(z, \tilde{z})}$, it is helpful to work with the logits of the binary variables $\alpha_{i} = \logit \frac{q(z=i)}{\sum_{j=i+1}^C q(z=j)}$. First, we know that $q(z = i) = \prod_{j=1}^{i-1} \sigma(-\alpha_j)\sigma(\alpha_i)$. For a pair of antithetic binary variables $(b_i, \tilde{b}_i)$, the coupling joint probability is
\[ p(b_i, \tilde{b}_i) = 
\begin{cases}
    \max(1 - 2\sigma(\alpha_i), 0) & b_i = \tilde{b}_i = 0 \\    
    \max(2\sigma(\alpha_i)-1, 0) & b_i = \tilde{b}_i = 1 \\    
    \sigma(-|\alpha_i|) & \text{o.w.}     
\end{cases}. \]
Since the categories are arranged in the ascending order of probability, $\alpha_i \leq 0$ for $i < C$, so the joint probability simplifies to
\[ p(b_i, \tilde{b}_i) = 
\begin{cases}
    1 - 2\sigma(\alpha_i) & b_i = \tilde{b}_i = 0 \\    
    0 & b_i = \tilde{b}_i = 1 \\    
    \sigma(\alpha_i) & \text{o.w.}     
\end{cases},\]
for $i < C$. We do not need to compute the entries for $z = \tilde{z}$ as the integrand already vanishes. Because of symmetry, without loss of generality, assume $z < \tilde{z}$. From the construction of $z$ and $\tilde{z}$ in terms of the binary variables, we can reason about their values. We know that for $i < z$, we must have $b_i = \tilde{b}_i = 0$. Then, for $i = z$, we must have $b_i = 1$ and for $z \leq i < \tilde{z}$, $\tilde{b}_i = 0$, and finally, for $i = \tilde{z}$, $\tilde{b} = 1$. Putting this together yields
\[ p(z, \tilde{z}) = \left[\prod_{i=1}^{z-1}(1 - 2\sigma(\alpha_i))\right]\sigma(\alpha_z)\left[\prod_{j=z+1}^{\tilde{z}-1}\sigma(-\alpha_j)\right]\sigma(\alpha_{\tilde{z}}). \]
Thus, the importance weights are
\begin{align*}
   \frac{q_\theta(z)q_\theta(\tilde{z})}{p_\theta(z, \tilde{z})} &= \frac{\left[\prod_{i=1}^{z-1} \sigma(-\alpha_i)\right]\sigma(\alpha_z)\left[\prod_{j=1}^{\tilde{z}-1} \sigma(-\alpha_j)\right]\sigma(\alpha_{\tilde{z}})}{\left[\prod_{i=1}^{z-1}(1 - 2\sigma(\alpha_i))\right]\sigma(\alpha_z)\left[\prod_{j=z+1}^{\tilde{z}-1}\sigma(-\alpha_j)\right]\sigma(\alpha_{\tilde{z}})} = \frac{\left[\prod_{i=1}^{z-1} \sigma(-\alpha_i)^2\right]\sigma(-\alpha_z)}{\prod_{i=1}^{z-1}(1 - 2\sigma(\alpha_i))}.
\end{align*}

\subsubsection{Unbiasedness of $g_\text{DisARM-SB}$}
Recall the $g_\text{DisARM-SB}$ estimator. We have binary variables $b_{k, 1} \sim \Bernoulli(\sigma(\alpha_{k, 1})), \ldots, b_{k, C}\sim \Bernoulli(\sigma(\alpha_{k, C}))$ and independently sampled antithetic pairs $\{ \tilde{b}_{k, c} \}$ such that $z = h(b)$ and $\tilde{z} = h(\tilde{b})$. The estimator is
\begin{equation*}
    {g_\text{DisARM-SB}}_{k, c} = \begin{cases}
         \frac{1}{2} \left(f(z) - f(\tilde{z})\right)\left(
                    (-1)^{\tilde{b}_{k, c}}  \1_{b_{k, c} \ne \tilde{b}_{k, c}}  \sigma(|(\alpha_{k, c}|)
                \right) & c \le \min(z_k, \tilde{z}_{k}) \\
    \frac{1}{2}\left(f(\tilde{z}) - f(z)\right)\nabla_{\alpha_{k, c}} \log q_\theta(\tilde{b}_{k, c}) & z_k < c \leq \tilde{z}_k \\
    \frac{1}{2}\left(f(z) - f(\tilde{z})\right)\nabla_{\alpha_{k, c}} \log q_\theta(b_{k, c}) & \tilde{z}_k < c \leq z_k \\
    0 & c > \max(z_k, \tilde{z}_k)
    \end{cases}.
\end{equation*}

We claim that for any $b_{-k, c}$ and $\tilde{b}_{-k, c}$,
\[ \E_{b_{k, c}, \tilde{b}_{k, c}}\left[ {g_{\text{DisARM-SB}}}_{k, c} \right] = \frac{1}{2}\E_{b_{k, c}} \left[ f(h(b)) \nabla_{\alpha_{k, c}} \log q_\theta(b_{k, c}) \right] + \frac{1}{2}\E_{\tilde{b}_{k, c}} \left[ f(h(\tilde{b})) \nabla_{\alpha_{k, c}} \log q_\theta(\tilde{b}_{k, c}) \right], \]
which immediately implies unbiasedness. Importantly, the conditions defining the estimator can be determined solely based on $b_{-k,c}$ and $\tilde{b}_{-k, c}$, so it suffices to verify that the estimator is unbiased for each case separately. In the first case, $g_\text{DisARM-SB}$ is the DisARM estimator from~\citep{dong2020disarm} which was previously shown to be unbiased. The second and third cases are reminiscent of the 2-sample RLOO estimator, however, in the coupled case, such an estimator must be justified as unbiased. In the second case, 
\begin{align*}
    \E_{b_{k, c}, \tilde{b}_{k, c}} &\left[ \frac{1}{2}\left(f(\tilde{z}) - f(z)\right)\nabla_{\alpha_{k, c}} \log q_\theta(\tilde{b}_{k, c}) \right] \\
    &= \frac{1}{2}\E_{\tilde{b}_{k, c}} \left[ f(\tilde{z}) \nabla_{\alpha_{k, c}} \log q_\theta(\tilde{b}_{k, c}) \right] - \frac{1}{2}\E_{\tilde{b}_{k, c}} \left[ \E_{b_{k, c} | \tilde{b}_{k, c}}\left[ f(z) \right] \nabla_{\alpha_{k, c}} \log q_\theta(\tilde{b}_{k, c}) \right] \\
    &= \frac{1}{2}\E_{\tilde{b}_{k, c}} \left[ f(\tilde{z}) \nabla_{\alpha_{k, c}} \log q_\theta(\tilde{b}_{k, c}) \right] - \overbrace{\frac{1}{2}\E_{\tilde{b}_{k, c}} \left[ f(z) \nabla_{\alpha_{k, c}} \log q_\theta(\tilde{b}_{k, c}) \right]}^{0} \\
    &= \frac{1}{2}\E_{\tilde{b}_{k, c}} \left[ f(\tilde{z}) \nabla_{\alpha_{k, c}} \log q_\theta(\tilde{b}_{k, c}) \right] + \overbrace{\frac{1}{2}\E_{b_{k, c}} \left[ f(z) \nabla_{\alpha_{k, c}} \log q_\theta(b_{k, c}) \right]}^{0}.
\end{align*}
$f(z)$ depends on $\tilde{b}_{k, c}$ through the coupled sample $b_{k, c}$, however the condition $c > z_k$ implies that $z$ does not depend on $b_{k, c}$, so $f(z)$ is a constant with respect to both $b_{k, c}$ and $\tilde{b}_{k, c}$, resulting in the vanishing terms. The third case follows by symmetry. In the fourth case, the condition $c > \max(z_k, \tilde{z}_k)$ implies that neither $z$ nor $\tilde{z}$ depend on $b_{k, c}$ or $\tilde{b}_{k, c}$ hence the gradient vanishes.

\subsection{Tree-Structured Coupling}
\label{app:tree}
We construct a categorical sample based on a binary sequence arranged as a balanced binary tree. Considering a binary sequence $b = \left[ b_0, b_1, b_2, \cdots, b_{C-1} \right]$, we interpret as a binary tree recursively with root $b[0]$ and left subtree $b[1:len(b)//2+1]$ and right subtree $b[len(b)//2+1:]$. The binary variables correspond to internal routing decisions in the binary tree with the categories as the leaves, so that $T(b)$ is defined by the following recursive function (assuming $C$ is a power of $2$)
\begin{lstlisting}[language=Python]
def T(b):
    half = len(b)//2 + 1
    if not b:
        return 1
    elif b[0] == 0:
        return T(b[1:half])
    else:
        return half + T(b[half:])
\end{lstlisting}
For each binary variable $b_i$, we can find all the categories residing in its left subtree ($\mathrm{Left(i)}$) and all the categories in its right subtree $\mathrm{Right(i)}$. We would like the probability of traversing the right subtree $\sigma(\alpha_i)$ to be such that 
\[
\sigma(\alpha_i) = \frac{\sum_{j\in \mathrm{Right(i)}} q(z = j)} 
{\sum_{j\in \mathrm{Right(i)}} q(z = j) + \sum_{j\in \mathrm{Left(i)}} q(z = j)}
\]
Hence,  
\begin{align*}
\alpha_i = \log\frac{\sum_{j\in \mathrm{Right(i)}} q(z = j)}{\sum_{j\in \mathrm{Left(i)}} q(z = j)}.
\end{align*}

Recall, the estimator from the maintext. Given binary variables $b_1, \ldots, b_{C-1}$, let $I(b_1, \ldots, b_{C-1})$ be the set of variables used in routing decisions ($|I(b_1, \ldots, b_{C-1})| = \log_2 C$). With a pair of antithetically sampled binary sequences $(b, \tilde{b})$, the following is an unbiased estimator:
\begin{equation}
    {g_\text{DisARM-Tree}}_{k,c} = \begin{cases}
         \frac{1}{2} \left(f(z) - f(\tilde{z})\right)\left(
                    (-1)^{\tilde{b}_{k,c}}  \1_{b_{k,c} \ne \tilde{b}_{k,c}}  \sigma(|(\alpha_{k,c}|)
                \right) & c \in I(b_{k, \cdot}) \cap I(\tilde{b}_{k, \cdot}) \\
    \frac{1}{2}\left(f(\tilde{z}) - f(z)\right)\nabla_{\alpha_{k,c}} \log q_\theta(\tilde{b}_{k,c}) & c \in I(\tilde{b}_{k, \cdot}) - I(b_{k, \cdot}) \\
    \frac{1}{2}\left(f(z) - f(\tilde{z})\right)\nabla_{\alpha_{k,c}} \log q_\theta(b_{k,c}) & c \in I(b_{k, \cdot}) - I(\tilde{b}_{k, \cdot}) \\
    0 & c \notin I(b_{k, \cdot}) \cup I(\tilde{b}_{k, \cdot}) 
    \end{cases}.
\end{equation}
Closely following the argument from the previous section, we claim that for any $b_{-k, c}$ and $\tilde{b}_{-k, c}$,
\[ \E_{b_{k, c}, \tilde{b}_{k, c}}\left[ {g_\text{DisARM-Tree}}_{k, c} \right] = \frac{1}{2}\E_{b_{k, c}} \left[ f(h(b)) \nabla_{\alpha_{k, c}} \log q_\theta(b_{k, c}) \right] + \frac{1}{2}\E_{\tilde{b}_{k, c}} \left[ f(h(\tilde{b})) \nabla_{\alpha_{k, c}} \log q_\theta(\tilde{b}_{k, c}) \right], \]
which immediately implies unbiasedness. Importantly, the conditions defining the estimator can be determined solely based on $b_{-k,c}$ and $\tilde{b}_{-k, c}$, so it suffices to verify that the estimator is unbiased for each case separately. In the first case, $g_\text{DisARM-Tree}$ is the DisARM estimator from~\citep{dong2020disarm} which was previously shown to be unbiased. The second and third cases are reminiscent of the 2-sample RLOO estimator, however, in the coupled case, such an estimator must be justified as unbiased. In the second case,  
$f(z)$ depends on $\tilde{b}_{k, c}$ through the coupled sample $b_{k, c}$, however the condition $c \in I(\tilde{b}_{k, \cdot}) - I(b_{k, \cdot})$ implies that $z$ does not depend on $b_{k, c}$, so $f(z)$ is a constant with respect to both $b_{k, c}$ and $\tilde{b}_{k, c}$, resulting in the vanishing terms. Thus, we can apply the same argument as in the previous section. The third case follows by symmetry. In the fourth case, the condition $c \notin I(b_{k, \cdot}) \cup I(\tilde{b}_{k, \cdot})$ implies that neither $z$ nor $\tilde{z}$ depend on $b_{k, c}$ or $\tilde{b}_{k, c}$ hence the gradient vanishes.

\subsection{Rao-Blackwellized ARS \& ARSM}
\label{sec:arsm}
First, we briefly review the Augment-REINFORCE-Swap (ARS) \& Augment-REINFORCE-Swap-Merge (ARSM) estimators \citep{yin2019arsm}. \citet{yin2019arsm} use the fact that the discrete distribution can be reparameterized by an underlying continuous augmentation: if $\vpidist$ and $z_k \coloneqq \argmin_i \pi_{k, i} e^{-\alpha_{k,i}}$, then $z_k \sim \Cat(\alpha_k)$; and show that $\nabla_{\alpha_{k, c}}\E_{q_\theta(z)} \left[ f(z) \right] = \E_{\pi} \left[ f(z)(1 - C\pi_{k,c}) \right].$
Furthermore, they define a swapped probability matrix $\pi_k^{\swap{i}{j}}$ by swapping the entries at indices $i$ and $j$ in $\pi_k$
\begin{equation*}
    \pi^{\swap{i}{j}}_{k, c} 
    \coloneqq 
    \begin{cases}
    \pi_{k, i} & c = j \\
    \pi_{k, j} & c = i \\
    \pi_{k, c} & \text{o.w.}
    \end{cases},
\end{equation*}
and $\zs{i}{j}_k \coloneqq \argmin_c \pi^{\swap{i}{j}}_{k, c} e^{-\alpha_{k, c}}$. Using these constructions, they show an important identity 
\begin{align*}
\nabla_{\alpha_{k, c}}\E_{q_\theta(z)}  \left[ f(z) \right] &= \E_\pi \left[ {\gARS}_{k, c} \coloneqq \left[f(\zs{c}{j}) - \frac{1}{C}\sum_{m=1}^C f(\zs{m}{j}) \right](1-C\pi_{k,j}) \right],
\end{align*}
which shows that ${\gARS}_{k, c}$ is an unbiased estimator. 
To further improve the estimator, \citet{yin2019arsm} average over the choice of the reference $j$, resulting in the ARSM estimator
\begin{align*}
{\gARSM}_{k, c} \coloneqq \frac{1}{C} \sum_{j=1}^C \left[ f(\zs{c}{j}) - \frac{1}{C}\sum_{m=1}^C f(\zs{m}{j}) \right](1-C\pi_{k,j}).
\end{align*}
Notably, both ARS and ARSM only evaluate $f$ at discrete values, and thus  do not rely on a continuous relaxation. 

\subsubsection{Rao-Blackwellization}
Motivated by the approach of \citet{dong2020disarm}, we can derive improved versions of ARS and ARSM by integrating out the extra randomness due to the continuous variables. ARS and ARSM heavily rely on a continuous reparameterization of the problem, yet the original problem only depends on the discrete values.  Ideally, we would integrate out  $\pi | \zs{1}{j},...,\zs{C}{j}$, however, unlike in the binary case, computing the expectation analytically appears infeasible. Instead, we analytically integrate out two dimensions of $\pi$ and use Monte Carlo sampling to deal with the rest. This is a straightforward albeit tedious calculation

Starting with ARS, ideally, we would like to compute
\begin{align*}
    \E_{\pi | \zs{1}{j},...,\zs{C}{j}}\left[{\gARS}_{k, c} \right] &= \left[f(\zs{c}{j}) - \frac{1}{C}\sum_{m=1}^C f(\zs{m}{j}) \right](1-C\E_{\pi | \zs{1}{j},...,\zs{C}{j}}\left[\pi_{k, j}\right]) \\
    &= \left[f(\zs{c}{j}) - \frac{1}{C}\sum_{m=1}^C f(\zs{m}{j}) \right](1-C\E_{\pi_{k, j} | \zs{1}{j}_k,...,\zs{C}{j}_k}\left[\pi_{k, j}\right]),
\end{align*}
taking advantage of independence between dimensions (indexed by $k$). To reduce notational clutter, we omit the dimension index in the following derivation. 

We have reduced the problem to computing
\[ \E_{\pi_j | \zs{1}{j},...,\zs{C}{j}}\left[\pi_j\right], \]
but were unable to compute the expectation analytically. Instead, we analytically integrate out some dimensions of $\pi$ and use Monte Carlo sampling to deal with the rest. First, we know that $\sum_i \pi_i = 1$, so one variable is redundant (denote this choice by $l$). Next, we show how to integrate the reference index $j \neq l$ (i.e., compute $\E_{\pi_j | \pi_{-j, l}, \zs{1}{j},...,\zs{C}{j}}\left[\pi_j \right]$, where $\pi_{-j,l}$ denotes $\pi$ excluding its $j$-th and $l$-th elements.). 

The known values of $\pi_{-j,l}, \zs{1}{j},...,\zs{C}{j}$ imply lower and upper bounds on $\pi_j$. First, because $1 - \sum_{i \neq l} \pi_i = \pi_l \geq 0$, we conclude that $\pi_j \leq 1 - \sum_{i \neq j, l} \pi_i$. To determine the implications of the configurations $\zs{1}{j},...,\zs{C}{j}$, it is helpful to define some additional notation. Let $\vscj \coloneqq \pi^{\swap{c}{j}} e^{-\mathbf{\alpha}}$.
Let's look at what the value of $\zs{m}{j} \coloneqq \argmin_i \vsmj_i$ tells us about $\pi_j$.  We need to consider two cases:
\begin{itemize}
\item $\zs{m}{j} = m$: This means that $\vsmj_m = \pi_j e^{-\alpha_m}$ is the smallest entry in $\vsmj$: $\pi_j e^{-\alpha_m} \leq \min_{i \neq m} \vsmj_i$, which implies that $\pi_j \leq \min_{i \neq m} e^{\alpha_m} \vsmj_i$.

$e^{\alpha_m} \vsmj_i$ contains $\pi_l$ when $m = l$ and $i = j$ or $m \neq l$ and $i = l$. When $m = l$ and $i = j$, we have that $e^{\alpha_l} \vsmj_j = e^{\alpha_l} \pi_l e^{-\alpha_j} = (1 - \sum_{n \neq j, l}\pi_n - \pi_j)e^{\alpha_l - \alpha_j}$. Therefore, 
\[ \pi_j \leq \frac{(1 - \sum_{n \neq j, l}\pi_n)e^{ - \alpha_j}}{e^{-\alpha_j} + e^{-\alpha_l}}.\]
A similar computation is required for the case $m \neq l$ and $i = l$.
\item $\zs{m}{j} \neq m$: This means that $\pi_j e^{-\alpha_m}$ is larger than the smallest entry in $\vsmj$: $\pi_j e^{-\alpha_m} \geq \min_i \vsmj_i$ which implies that $\pi_j \geq \min_i e^{\alpha_m} \vsmj_i$. As above, we can eliminate $\pi_l$ from the bounds.
\end{itemize}
Finally, we aggregate the inequalities to compute the lower and upper bounds.  Because $\pi \sim \text{Dirichlet}(1_C)$ is a uniform distribution over the simplex, $\pi_j | \pi_{-j,l}, \zs{1}{j},...,\zs{C}{j}$ will be uniformly distributed over an interval
, which means that it suffices to compute the lower and upper bounds to compute the expectation.

We can apply the same ideas to ARSM, however, in preliminary experiments with ARS+, we found that leveraging the symmetry (described next) was responsible for most of the performance improvement for ARS+. So, for ARSM, we reduce the variance only by leveraging the symmetry and call the resulting estimator ARSM+.

Furthermore, When all of the swapped $z$s agree on a dimension (i.e., $\zs{1}{j}_k=\cdots =\zs{C}{j}_k$), then we will show that both ${\gARS}_{k, c}$ and ${\gARSM}_{k, c}$ vanish in expectation, so we can zero out these terms explicitly. The high level intuition is that even though they may disagree in other dimensions for a single sample because the other dimensions are independent and expectations are linear, in expectation they cancel out. Let $\delta_k = \mathbbm{1}_{\zs{1}{j}_k=\cdots =\zs{C}{j}_k}$. Then, we have
\begin{align*}
   \E_{\pi | \delta_k = 1} \left[ {\gARS}_{k,c} \right] &= \E_{\pi | \delta_k} \left[ \left[f(\zs{c}{j}) - \frac{1}{C}\sum_{m=1}^C f(\zs{m}{j}) \right](1-C\pi_{k,j}) \right] \\
   &= \E_{\pi_k | \delta_k = 1}\left[\left(\E_{\pi_{-k}} \left[ f(\zs{c}{j}) \right] - \frac{1}{C}\sum_{m=1}^C \E_{\pi_{-k}} \left[f(\zs{m}{j})\right] \right) (1-C\pi_{k,j}) \right].
\end{align*}
Now, we claim that inside the expectation $\E_{\pi_{-k}} \left[f(\zs{m}{j})\right]$ is constant with respect to $m$. First, we know that inside the expectation $\zs{1}{j}_k=\cdots =\zs{C}{j}_k$ and that the dimensions indexed by $k$ are independent. Because $\vpidist$, $\text{Dirichlet}(1_C)$ is symmetric, and we are taking an unconditional expectation over the remaining dimensions, the value is invariant to the swapping operation. As a result, the entire expression vanishes. Thus, we conclude that 
\[ (1 - \delta_k){\gARS}_{k,c}  \]
is still an unbiased estimator. A similar argument holds for $\gARSM$. This is complementary to the approach in the previous subsection and can done in combination
\[ {g_{\text{ARS+}}}_{k, c} \coloneqq \E_{\pi_{k,j} | \pi_{k,-jl}, \zs{1}{j}_k,...,\zs{C}{j}_k}\left[{(1 - \delta_k)\gARS}_{k,c} \right],\]
where we choose $l \neq j$ uniformly at random. This is the estimator we use in our experiments.

\subsubsection{Evaluating Rao-Blackwellized ARS \& ARSM}
We train models with $10$/$5$/$3$/$2$-category latent variables on dynamically binarized MNIST. For comparison, we train models with ARS, ARSM, and an $n$-sample RLOO. To match computation, RLOO uses $C$ samples for comparing against ARS/ARS+, and uses $C(C-1)/2+1$ samples for ARSM/ARSM+, where $C$ is the number of categories. Based on preliminary experiments with ARS+, we found that leveraging the symmetry led to most of the performance improvement for ARS+. So, for ARSM, we reduce the variance only by leveraging the symmetry and call the resulting estimator ARSM+. 

\begin{figure}[h]
    \centering
    \rotatebox[origin=l]{90}{{\scriptsize {\quad \qquad $5$-category}}}
    \;
    \includegraphics[width=0.31\linewidth]{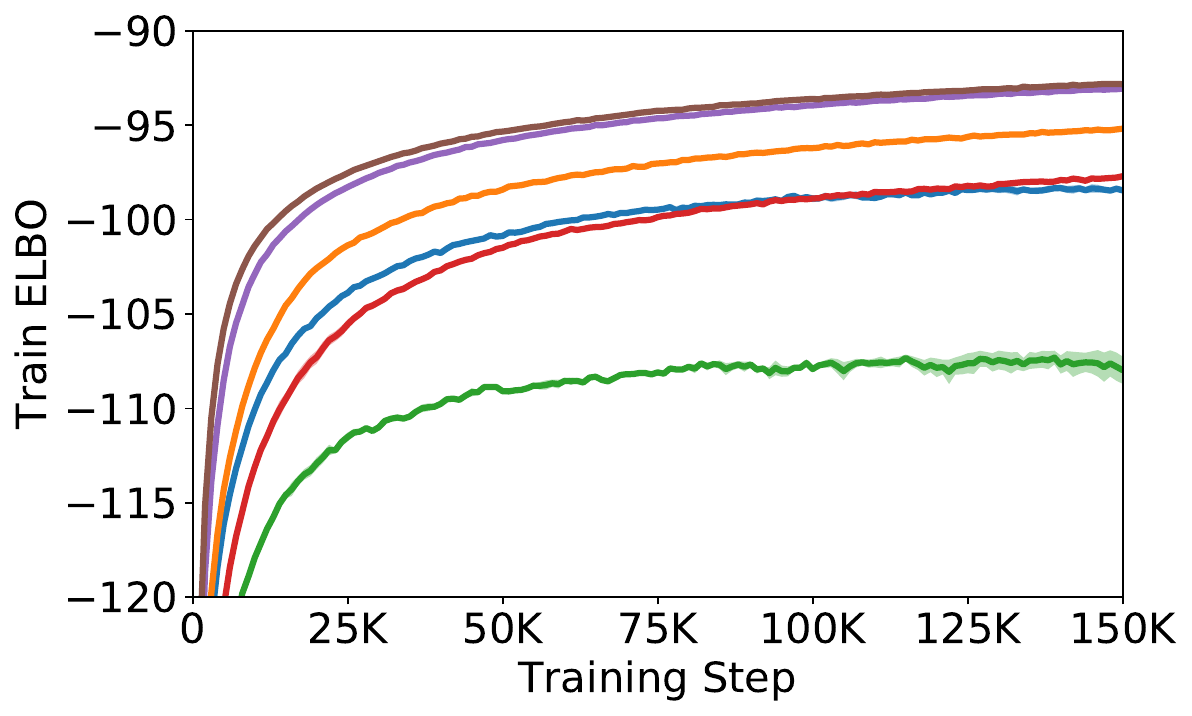}
    \;
    \includegraphics[width=0.31\linewidth]{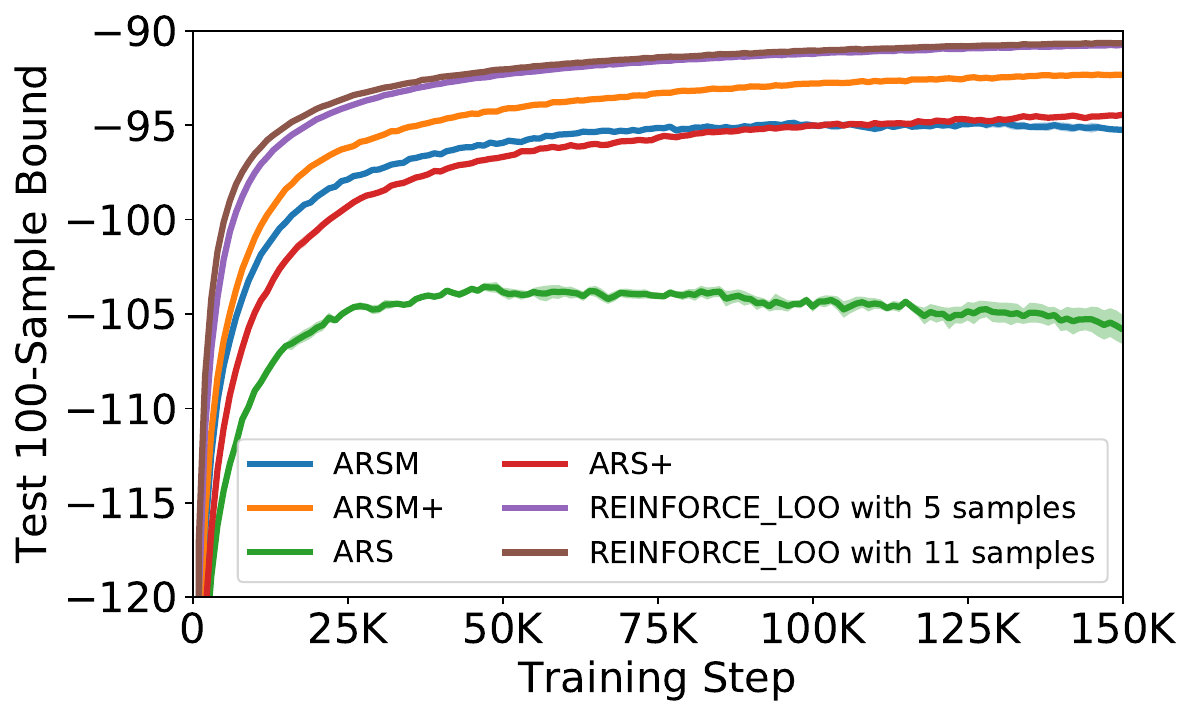}
    \;
    \includegraphics[width=0.29\linewidth]{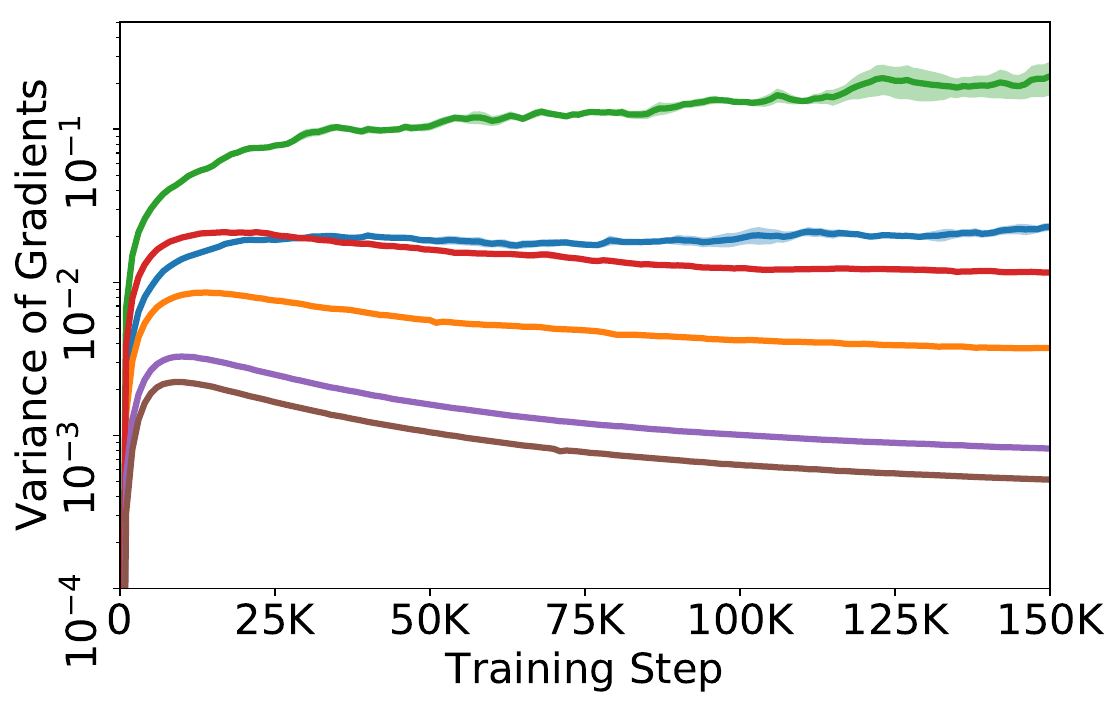}
    \caption{Training a non-linear categorical VAE with latent variables with $5$ categories on dynamically binarized MNIST dataset by maximizing the ELBO. We plot the train ELBO (left column), test 100-sample bound (middle column), and the variance of gradient estimator (right column). We plot the mean and one standard error based on $5$ runs from different random initializations.}
    \label{fig:experiment-arsm}
\end{figure}

\begin{figure}[!h]
    \centering
    \rotatebox[origin=l]{90}{{\scriptsize {\quad \qquad $10$-category}}}
    \;
    \includegraphics[width=0.31\linewidth]{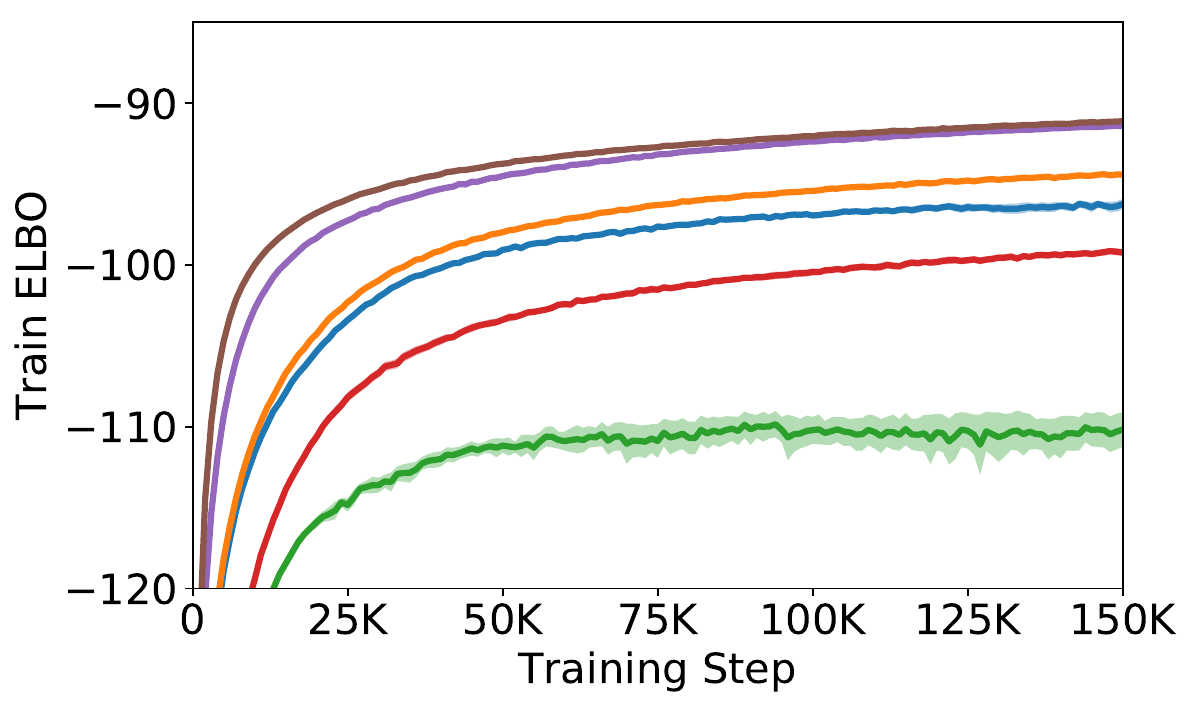}
    \;
    \includegraphics[width=0.31\linewidth]{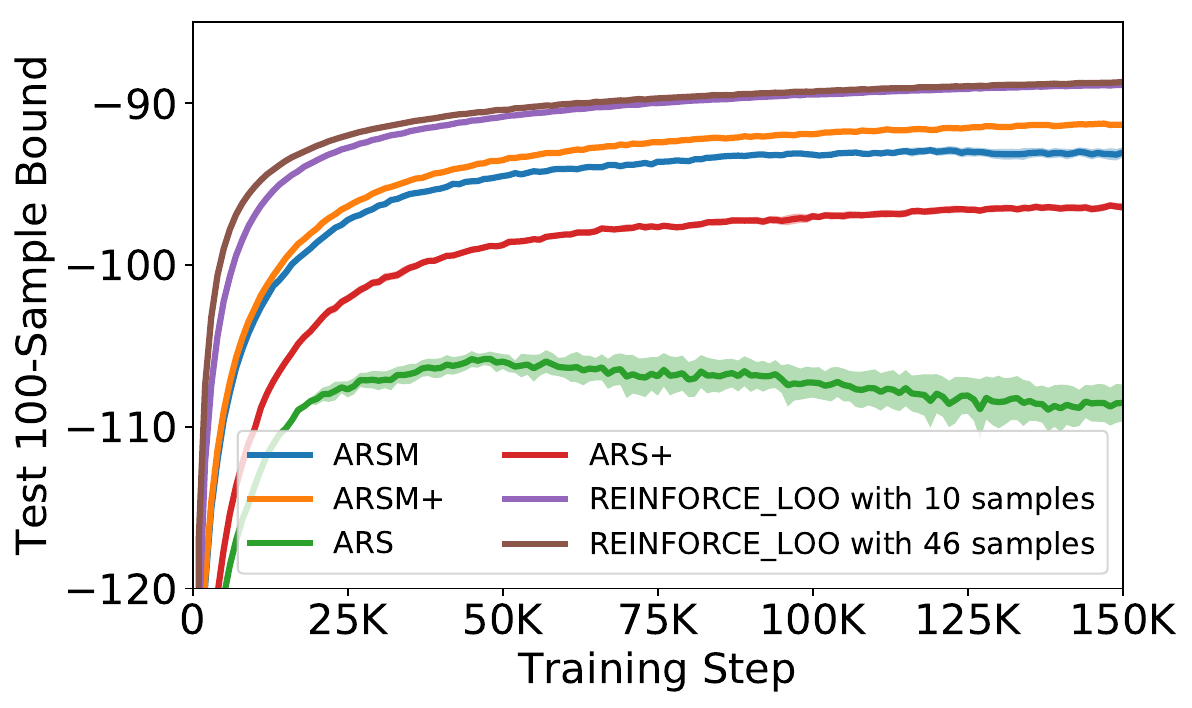}
    \;
    \includegraphics[width=0.29\linewidth]{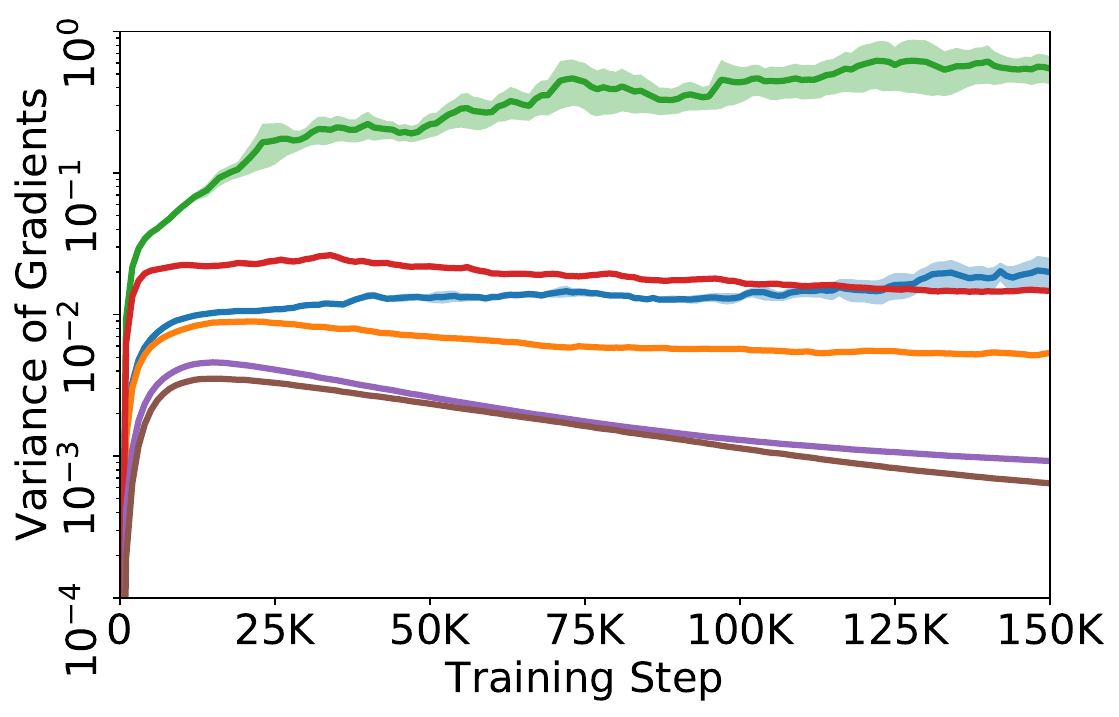}
    \\
    \rotatebox[origin=l]{90}{{\scriptsize {\quad \qquad $5$-category}}}
    \;
    \includegraphics[width=0.31\linewidth]{figures/dynamic_5-category_train_eval_legend_false.pdf}
    \;
    \includegraphics[width=0.31\linewidth]{figures/dynamic_5-category_test_eval_legend_true.pdf}
    \;
    \includegraphics[width=0.29\linewidth]{figures/dynamic_5-category_var_legend_false.pdf}
    \\
    \rotatebox[origin=l]{90}{{\scriptsize {\quad \qquad $3$-category}}}
    \;
    \includegraphics[width=0.31\linewidth]{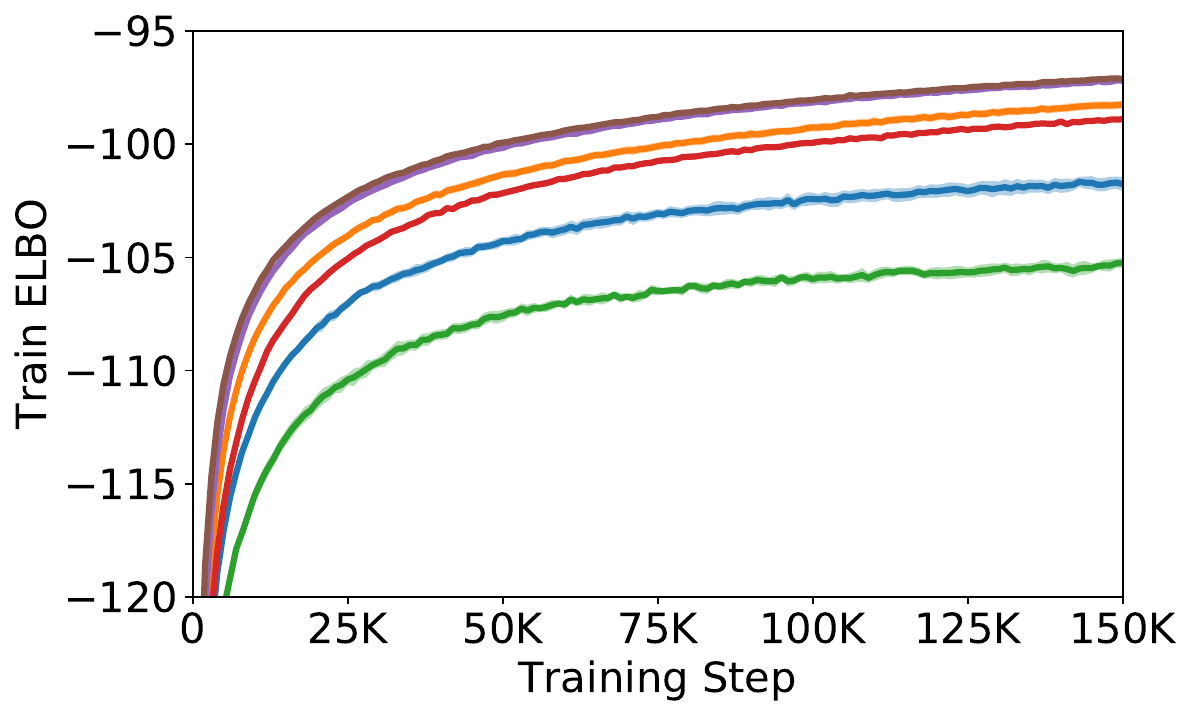}
    \;
    \includegraphics[width=0.31\linewidth]{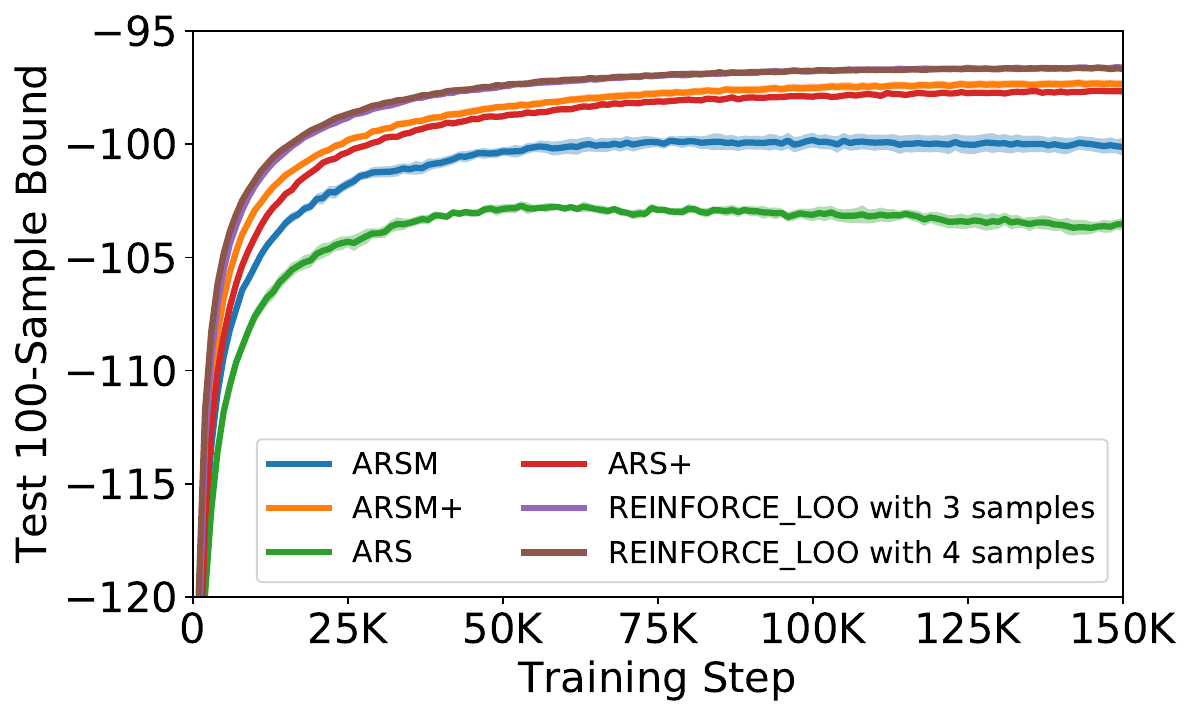}
    \;
    \includegraphics[width=0.29\linewidth]{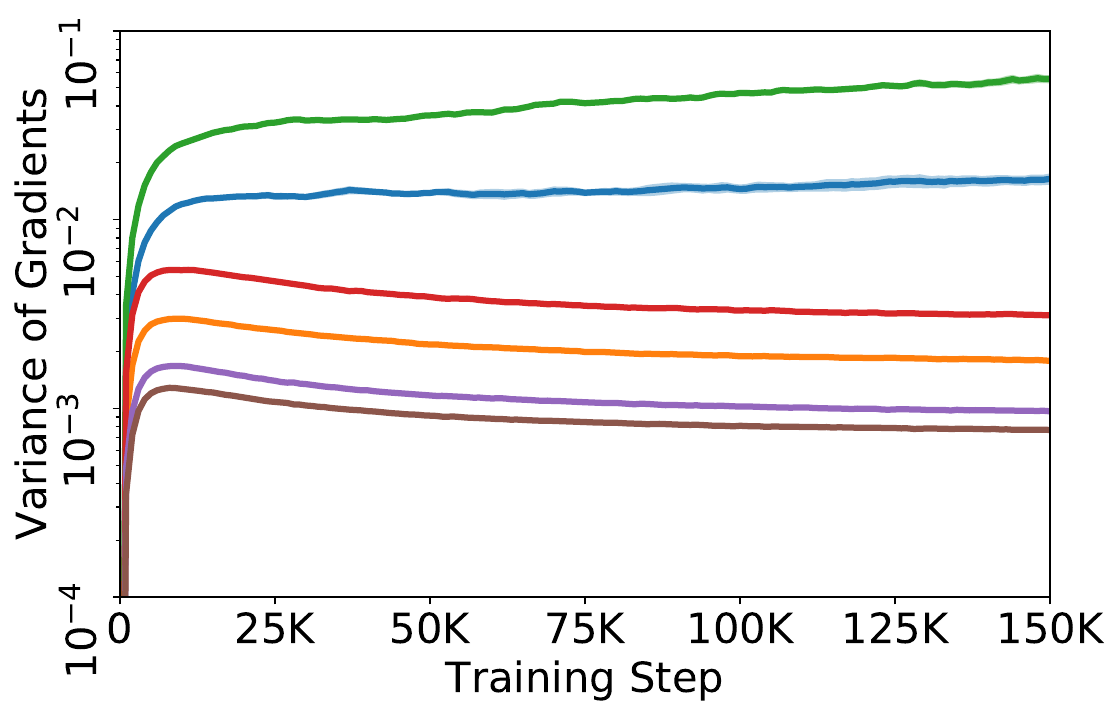}
    \\
    \rotatebox[origin=l]{90}{{\scriptsize {\quad \qquad $2$-category}}}
    \;
    \includegraphics[width=0.31\linewidth]{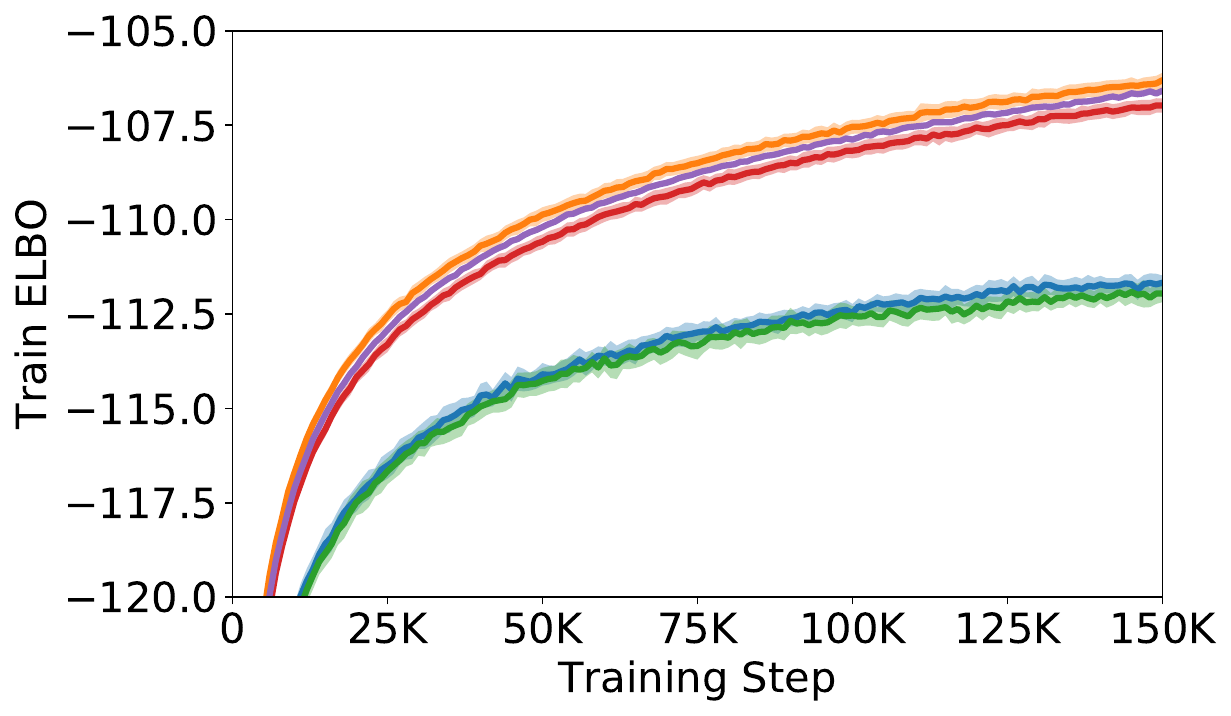}
    \;
    \includegraphics[width=0.31\linewidth]{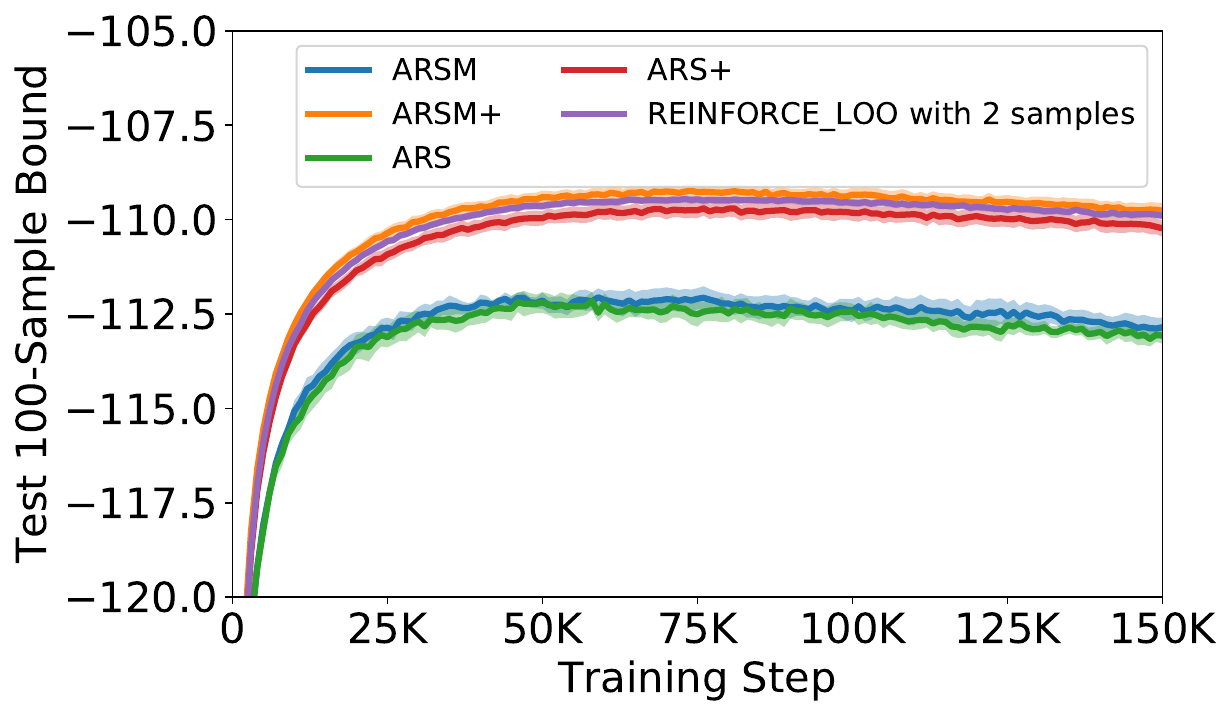}
    \;
    \includegraphics[width=0.29\linewidth]{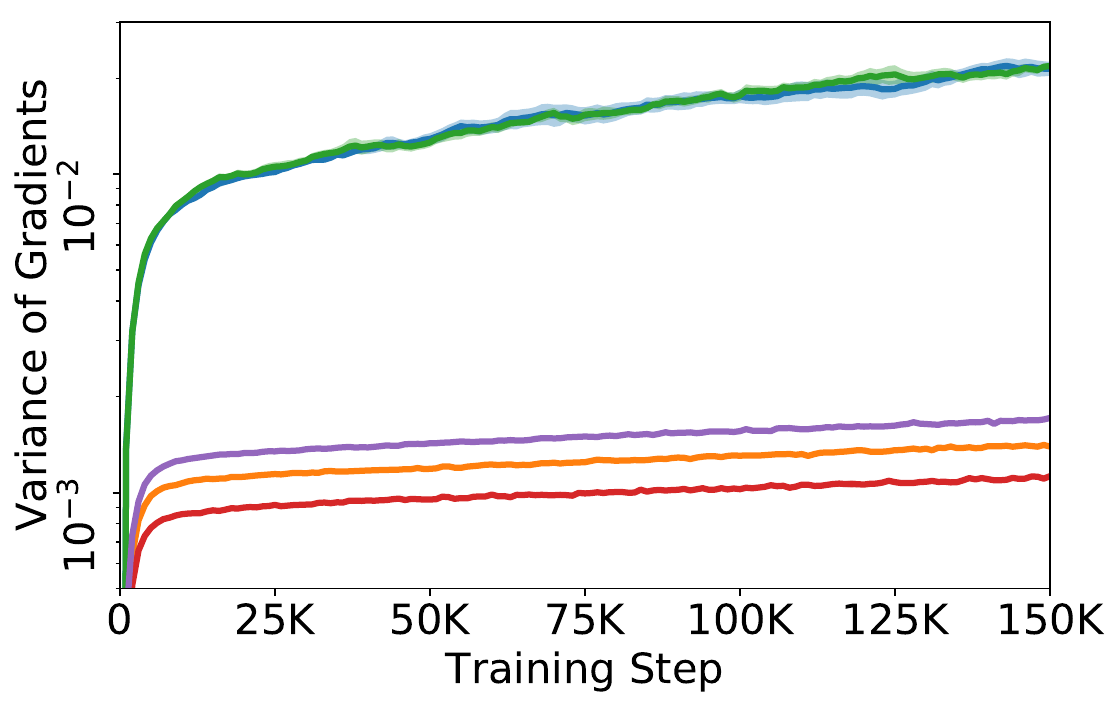}
    \\
    \caption{Training a non-linear categorical VAE with latent variables with $10$/$5$/$3$/$2$ categories on dynamically binarized MNIST dataset by maximizing the ELBO. We plot the train ELBO (left column), the test 100-sample bound (middle column), and the variance of gradient estimator (right column). We plot the mean and one standard error based on $5$ runs from different random initializations.}
    \label{fig-appendix:experiment-arsm}
\end{figure}

As shown in \Cref{fig:experiment-arsm} and Appendix Figure~\ref{fig-appendix:experiment-arsm}, the proposed estimators, ARS+/ARSM+, significantly outperform ARS/ARSM. Surprisingly, we find that both ARS and ARSM underperform the simpler RLOO baseline in all cases. For $C=2$, ARS+ and ARSM+ reduce to DisARM/U2G and as expected, outperform REINFORCE LOO; however, for $C > 2$, REINFORCE LOO is superior and the gap increases as $C$ does. This suggests that partially integrating out the randomness is insufficient to account for the variance introduced by the continuous augmentation.


\end{document}